%% file: saliency_manuscript_v2.tex
\RequirePackage{snapshot}
\documentclass[11pt, a4paper, twocolumn]{article}
\pdfoutput=1

\usepackage[utf8]{inputenc}
\usepackage{amsmath}
\usepackage{amsfonts}
\usepackage{dsfont}
\usepackage{amssymb}
\usepackage[per-mode = symbol]{siunitx}
\usepackage{hyperref}
\usepackage{graphicx}
\usepackage{longtable}
\usepackage{todonotes}

\usepackage[font=small]{caption}

\newcommand*{\noktikz}{}
\usepackage{tikz}
\usetikzlibrary{calc}
\usetikzlibrary{shapes}

\graphicspath{{figures/}}

\newcommand\EE{\mathds{E}}
\newcommand\RR{\mathds{R}}
\DeclareMathOperator{\DKL}{D_{KL}}

\newcommand\xfix{x_\text{fix}}
\newcommand\xnonfix{x_\text{nonfix}}

\DeclareSIUnit\fix{fix}
\DeclareSIUnit\bits{bits}

\include{figures/data}
\include{figures/data_kienzle}

\author{Matthias Kümmerer, Thomas Wallis, Matthias Bethge}
\title{How close are we to understanding image-based saliency?}

\begin{document}
\maketitle

\abstract{
Within the set of the many complex factors driving gaze placement, the properities of an image that are associated with fixations under ``free viewing'' conditions have been studied extensively.
There is a general impression that the field is close to understanding this particular association. 
Here we frame saliency models probabalistically as point processes, allowing the calculation of log-likelihoods and bringing saliency evaluation into the domain of information.
We compared the information gain of state-of-the-art models to a gold standard and find that only one third of the explainable spatial information is captured.
We additionally provide a principled method to show where and how models fail to capture information in the fixations.
Thus, contrary to previous assertions, purely spatial saliency remains a significant challenge.

}

\section{Introduction}

\begin{figure}
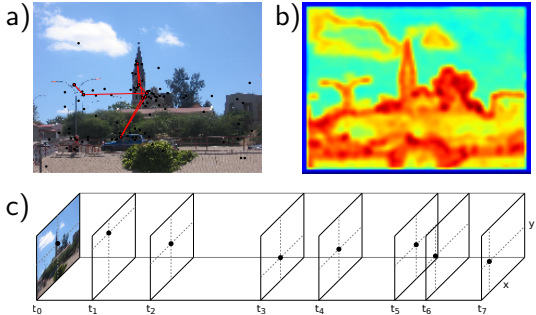

  \center
  \include{figures/intro_figure.pgf}
  \caption{
  \small
Fixations, saliency and point processes.
\textbf{(a)} Sample image with fixations from several subjects (black points) and the fixation train of one specific subject (red lines).
\textbf{(b)} Saliency map produced by the AIM model for the same image.
Red and blue denote higher and lower saliency respectively.
\textbf{(c)} The fixation train from (a) represented as a sample of a spatiotemporal point process.
Note that spatiotemporal point processes take into account both the spatial position and inter-fixation durations.
}
\label{fig:fixations_saliency_point_processes}
\end{figure}

The properties of an image that attract fixations during free viewing are said to be \textit{salient}, and models that attempt to capture these image features are saliency models (Figure \ref{fig:fixations_saliency_point_processes}a and b). 
Predicting where people fixate is relevant to both understanding visual information processing in biological systems and also to applications in computer vision and engineering. 
Beginning with the influential model of \cite{Itti1998}, there are now over 50 models of saliency as well as around 10 models that seek to incorporate top-down effects (see \cite{Borji2013, Borji2013b, Borji2013d} for recent reviews and analyses of this extensive literature).
There exists a general impression that these models have captured most of the association between image properties and fixation structure under free viewing conditions \cite{Einhauser2010, Borji2013b}.
For example, \cite{Einhauser2010} write that ``Recent elaborations of such stimulus-driven models are now approaching the limits imposed by intersubject variability''.
But how can we assess this question in a quantitatively precise way?

To judge our level of understanding, we must measure the distance between our model predictions and some upper bound of possible prediction performance using an interpretable scale.
The field is currently drowning in model comparison metrics, indicating uncertainty about what the right measure is.
For example, a recent paper \cite{Riche2013} included a comparison of 12 metrics with the recommendation that researchers use three of them to avoid the pitfalls of any one.
Following this recommendation would mean that measuring saliency and comparing saliency models is inherently ambiguous, since it is impossible to define a unique ranking if any two of the considered rankings are inconsistent.
Thus, one could never answer the question we pose.
Since no comparison of existing metrics can resolve this issue, we instead advocate a return to first principles.

Saliency is operationalised by measuring fixation densities. 
How well a model predicts these densities can be directly assessed using information theory.
As originally shown by Shannon \cite{Shannon1949}, information theory provides a unique and universally accepted measure of information gain from three basic axioms.
\textit{Information gain} is defined as the entropic difference between the prior and the posterior distribution.
In the context of image-based saliency, this quantifies the reduction in uncertainty (intuitively, the scatter of predicted fixations) about where people look.

To reiterate, understanding saliency is the \textit{distance} between model predictions and a gold standard.
By one of Shannon's axioms, information is a linear scale.
This means we can simply take the difference between a model and the gold standard to judge our level of understanding, making information gain distinct from any other metric.

To evaluate saliency models in terms of information we must use a probabalistic model.
It is natural to consider fixations and saliency as a probabalistic process, because humans do not make the same fixations as each other, or as themselves if viewing an image twice.
Probabalistic models can distinguish the inherent stochasticity of the fixation paths from uncertainty in the predictions. 
As recognised and applied by \cite{Barthelme2013}, an ideal framework to consider fixation distributions probabalistically is that of spatial point processes: the class of probability distributions of sets of points in space and time.

For probabalistic models, information gain is computed as the difference in average log-likelihood\footnote{Strictly, information gain refers to the information difference between prior and posterior; here we include the information difference between baseline and image-based saliency-model}.
While average log-likelihood can be rewritten in terms of Kullback-Leibler divergence (KL-divergence), our approach is fundamentally different from how KL-divergence has previously been used to compare saliency models.
Most importantly, we point out that average log-likelihoods are not invariant under reparameterization of the scale used to measure saliency.
Therefore, we emphasize that it is important to find the optimal scale for measuring saliency rather than just relying on its rank.
The theoretical underpinnings of our approach and its relationship to the KL-divergence measures used previously are discussed in detail in the Appendix (\ref{sec:kl_divergence}).


Here we evaluate saliency models using log-likelihood within a point process framework and compare each model's information gain relative to a baseline model to a gold standard.
Contrary to the impression that image-based saliency is essentially solved, our results show that current state-of-the-art models capture only one third of the explainable image-based spatial information in fixation locations.

\section{Methods}

\subsection{Point processes and log-likelihoods}
\label{sec:point_processes_methods}

We define a \textit{fixation train} as a three-dimensional point process.
A fixation train consists of $N$ fixations with positions $x_i$, $y_i$, $t_i$, where $x_i$ and $y_i$ denote the spatial position of the fixation in the image and $t_i$ denotes the time of the fixation (see Figure \ref{fig:fixations_saliency_point_processes}c).
Conceiving of fixation trains as 3D point processes allows us to model the joint probability distribution of all fixations of a subject on an image.
In general, a model's log-likelihood is $\frac1N\sum_k \log p(x_k)$, where $p$ is the probability function of the model and $x_k$, $k=1, \dots, N$ are samples from the probabilistic process that we would like to model.
Our likelihoods are therefore of the form
\[
  p(x_1, y_1, t_1, \dots, x_N, y_N, t_N, N),
\]
where $N$ is part of the data distribution, not a fixed parameter.
By chain rule, this is decomposed into conditional likelihoods

\begin{align*}
  &p(x_1, y_1, t_1, \dots, x_N, y_N, t_N, N)\\
  &\quad =p(N)\prod_{i=1}^N p(x_i, y_i, t_i
  \mid N, x_1, y_1, t_1,
 \\ &\qquad \qquad \qquad \qquad
   \dots, x_{i-1}, y_{i-1}, t_{i-1})
\end{align*}

The above holds true for any three-dimensional point process.
We make the additional assumption that the conditional likelihoods do not depend on $N$\footnote{
To be precise, we condition on $N \geq i$, which the notation $p(x_i, y_i, t_i \mid x_1, y_1, t_1, \dots, x_{i-1}, y_{i-1}, t_{i-1})$ already implies.
Therefore the condition $N \geq i$ is omitted by abuse of notation.
}.
Furthermore, we assume that all models account for the factor $p(N)$ in the same way.
In the calculation of the log-likelihood, this gives rise to an additive constant that is the same for each model.
Since we are only interested in differences of log-likelihoods, we can omit it completely.

Here we use the logarithm to base two, meaning that the log-likelihoods we report are in bits.
If one model's log-likelihood exceeds another model's log-likelihood by 1~bit, the first model assigns on average double the likelihood to the data: it predicts the data twice as well.
Model comparison within the framework of likelihoods is well-defined and the standard of any statistical model comparison enterprise.

\subsection{A definition of ``image-based saliency''}
\label{sec:saliency_definition}

Image-based saliency is usually considered as the properties of the image that are associated with fixation selection, independent of image-independent factors like centre bias.
However, a precise definition of this term is lacking.
Now that we have defined a probabalistic framework for saliency, we can provide this precise definition because the influences of image-independent factors can be divided out.
This is not possible as long as saliency maps use only a range scale (i.e. are invariant under monotonic transformations).

We define the image-based saliency $s(x,y \mid I)$ for an image $I$ to be

\[
  s(x, y \mid I) = \frac{p(x, y \mid I)}{p(x,y)},
\]

where $p(x,y \mid I)$ is the fixation distribution of this image while $p(x,y)$ is the image independent prior distribution.
Consequently, the prediction of image-based saliency for a model $\hat p(x, y \mid I)$ is $\hat s(x,y \mid I) = \frac{\hat p(x,y \mid I)}{p(x,y)}$.
This definition of image-based saliency has the property of having the best possible shuffled AUC (area under the receiver operating characteristic curve; \cite{Barthelme2013}).

\begin{figure*}[!ht]
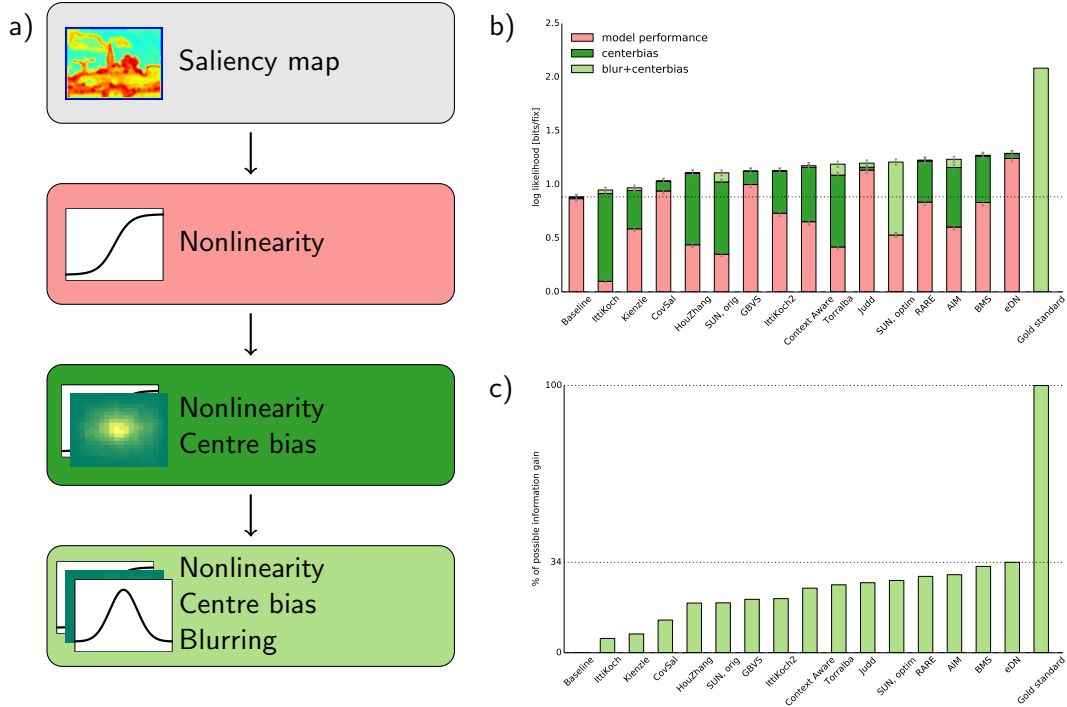

  \center
  \include{figures/modeling_stationary.pgf}

  \caption{
  \small
Modelling procedure and results. 
\textbf{(a)} For each saliency model we optimised three spatial, image-independent factors that increased the complexity of each model: a pointwise nonlinearity, the centre bias and the blur kernel.
\textbf{(b)} Log likelihood (bits/fixation) for each model considered relative to a null model which assumes uniform fixation density over the image. 
The total bar height represents the optimised performance of each model; the coloured sections show the relative contributions of each model factor from (a).
The baseline model is a nonparametric model of all fixations from other images (thus capturing purely image-independent spatial biases).
The gold standard model is the (Gaussian blurred) empirical fixation distributions for all other observers for each image: it represents all the spatial structure of fixations that can be captured (image-dependent information).
\textbf{(c)} Fully-optimised model performance, expressed as a percentage of the possible information gain from baseline (0) to gold standard (100).
The best performing model explains only \SI{\BestModelPercentOfGoldRelBaseline}{\percent}.
}
\label{fig:modelling_and_results}
\end{figure*}

\subsection{Upper and lower bounds}

In evaluating the performance of a model we would like to know the best performance a model \textit{could possibly} have given some constraints, so that we can better judge how well a model does.
We created three baseline models to facilitate comparison of the saliency models.

The \textbf{maximum entropy model} assumes that fixations are uniformly distributed over the image (i.\,e. the gaze density is constant).
All log-likelihoods are calculated relative to this model.
Likelihoods greater than zero therefore represent an improvement (in bits/fixation) over the prediction performance of constant gaze density.

The \textbf{lower-bound model} is a 2D histogram model crossvalidated \textit{between} images (trained on all fixations for all observers on all other images).
Thus, it captures the image-independent spatial information in the fixations, and in this paper is used as an estimate of $p_{prior}$. 
Bincount and regularization parameters were optimised by gridsearch.
If a saliency model captured all the behavioural fixation biases but nothing about what causes parts of an image to attract fixations, it would do as well as the lower-bound model.

Fixation preferences that are inconsistent between observers are by definition unpredictable from fixations alone.
If we have no additional knowledge about inter-observer differences, the best predictor of an observer's fixation pattern on a given image is therefore to average the fixation patterns from all other observers and add regularization. 
This is our \textbf{gold standard model}. 
It was created by blurring the fixations with a Gaussian kernel, learned by 10-fold crossvalidation between subjects.
It accounts for the amount of information in the spatial structure of fixations to a given image that can be explained while averaging over the biases of individual observers.
This model is the upper bound on prediction in the dataset (see \cite{Wilming2011} for a thorough comparison of this gold standard and other upper bounds capturing different constraints).

\subsection{Saliency models}

We converted a range of influencial saliency map models into point processes and calculated their likelihoods, relative to the maximum entropy model. 
We included all models from the saliency benchmark in \cite{Judd2012} and the top performing models that have been added later on the web page belonging to this benchmark\footnote{\url{http://saliency.mit.edu/}} up to July 2014. 
For all models, the original source code and default parameters have been used unless stated otherwise. 
The included models are \textbf{Itti \& Koch}
\cite{Itti1998} (here, two implementations have been used: The one from
the Saliency Toolbox\footnote{\url{http://www.saliencytoolbox.net}} and the
variant specified in the GBVS paper),
\textbf{Torralba}\footnote{\url{http://people.csail.mit.edu/tjudd/SaliencyBenchmark/Code/torralbaSaliency.m}}
\cite{Torralba2006},
\textbf{GBVS}\footnote{\url{http://www.klab.caltech.edu/~harel/share/gbvs.php}} \cite{Harel2006},
\textbf{SUN} \cite{Zhang2008} (we used a scale parameter of $0.64$, corresponding to the pixel size of $2.3^\prime$ of visual angle of the dataset used to learn the filters. This model will be called ``SUN, orig'' in the following. Additionally, we did a grid search over the scale parameter. This resulted in a scale parameter of $0.15$ for the model called ``SUN, optim''),
\textbf{Kienzle}\footnote{Code provided by Simon Barthelm\'{e}}\cite{Kienzle2007a, Kienzle2009} (with a patch size of $195$px corresponding to the optimal patchsize of $5.4^\circ$ reported by them).
\textbf{Hou \&
Zhang}\footnote{\url{http://www.klab.caltech.edu/~xhou/projects/spectralResidual/spectralresidual.html}}
\cite{Hou2007},
\textbf{AIM}\footnote{\url{http://www-sop.inria.fr/members/Neil.Bruce/}}
\cite{Bruce2009},
\textbf{Judd}\footnote{\url{http://people.csail.mit.edu/tjudd/WherePeopleLook/index.html}}
\cite{Judd2009},
\textbf{Context-Aware
saliency}\footnote{\url{http://webee.technion.ac.il/labs/cgm/Computer-Graphics-Multimedia/Software/Saliency/Saliency.html}}
\cite{Goferman2010, Goferman2012},
\textbf{CovSal}\footnote{\url{http://web.cs.hacettepe.edu.tr/~erkut/projects/CovSal/}} \cite{Erdem2013},
%
\textbf{RARE2012}\footnote{\url{http://www.tcts.fpms.ac.be/attention/?categorie17/rare2012}}\cite{Riche2013a},
Boolean Map-based Saliency \textbf{(BMS)}\footnote{\url{http://cs-people.bu.edu/jmzhang/BMS/BMS.html}},
\cite{Zhang2013, Zhang2014}
 and finally
the Ensemble of Deep Networks \textbf{(eDN)}\footnote{\url{http://github.com/coxlab/edn-cvpr2014}}
\cite{Vig2014}.

\subsubsection{Conversion into point processes}

The models we consider above are all explicitly spatial: they do not include any temporal dependencies.
Therefore, here the general point process formula simplifies via marginalisation over time.
Each model was converted into a point process by treating the normalized saliency map as conditional gaze density for the next fixation (notation from Section \ref{sec:point_processes_methods}; $s(x,y)$ denotes the saliency at point $(x, y)$):

\begin{align*}
&p(x_j, y_j, t_j \mid (x_0, y_0, t_0) \dots (x_{j-1}, y_{j-1}, t_{j-1}))\\
  &\qquad= p(x_j, y_j, t_j)\\
  &\qquad = p(x_j, y_j)p(t_j)\\
  &\qquad \propto s(x_j, y_j)
\end{align*}

The first equality expresses the fact that the saliency models assume independence of the previous fixation history; the second equality states that they also do not take the actual time of the fixations into account.

Since many of these models were optimised for AUC, and since AUC is invariant to monotonic transformations whereas log-likelihood is not, simply comparing the models' raw saliency map would not be fair.
The saliency map for each model was therefore transformed by a \textit{pointwise monotonic nonlinearity} that was optimised to give the best log-likelihood for that model (see Figure \ref{fig:modelling_and_results}a).
This corresponds to picking the model with the best log-likelihood from all models that are equivalent (under AUC) to the original model.

Every saliency map was jointly rescaled to range from 0 to 1 (i.e. over all images at once, not per image, keeping contrast changes from image to image intact).
The pointwise monontonic nonlinearity was then applied.
This nonlinearity was modeled by a pointwise linear and continous function that was parametrized as a continuous piecewise linear function supported in 20 equidistant points $x_i$ between $0$ and $1$ with values $y_i$ with $0 \leq x_0 \leq \dots \leq x_{19}$:
$p_{\text{nonlin}}(x, y) \propto f_{\text{nonlin}}(s(x, y))$ with

\[
  f_{\text{nonlin}}(x) = \frac{y_{i+1}-y_{i}}{x_{i+1}-x_{i}} (x-x_i)+y_i
\]
for $x_i \leq x \leq x_{i+1}$.

\subsubsection{Including additional factors}

We iteratively increased the complexity of the basic saliency-as-point-process model (described above) by adding components that included additional factors (see Figure \ref{fig:modelling_and_results}a).
First, we optimised a \textit{centre bias} term that accounts for the fact that human observers tend to look towards the centre of the screen.
Second, we optimised a \textit{blur} term that compensates for models that make overly-precise, confident predictions of fixation locations \cite{Judd2012}.

The center bias was modeled as 

\begin{align*}
&p_{\text{centerbias}}(x,y) \\
&\qquad \propto f_{\text{centerbias}}(d(x,y))p_{\text{nonlin}}(x,y)
\end{align*}

Here, $d(x,y) = \sqrt{(x-x_{\text{c}})^2+\alpha(y-y_{\text{c}})^2}/d_{\text{max}}$ is the normalized distance of $(x,y)$ to the center of the image $(x_\text{c}, y_\text{c})$ with eccentricity $\alpha$, $f_{\text{centerbias}}(d)$ is again a continuous piecewise linear function that was fitted in 12 points.

The blurring was modeled by blurring $s(x, y)$ with a Gaussian kernel with radius $\sigma$ before the nonlinearity and the center bias were applied.

\subsubsection{Optimization}

For each case of spatial optimization described above (nonlinearity, nonlinearity+centerbias, nonlinearity+centerbias+blur), all parameters were optimized jointly using the L-BFGS SLSQP algorithm from \texttt{scipy.optimize} \cite{Scipy}. 

\subsubsection{Dataset}

We evaluated model log-likelihoods using a subset of the MIT-1003 fixation dataset \cite{Judd2009}. 
To better estimate the nonparametric baseline model we used only the most common image size (1024 $\times$ 768), resulting in 463 images included in the evaluation. 
We also compared model performance in the full dataset of Kienzle \cite{Kienzle2009} (see Discussion).

\section{Results}

\subsection{Whole database performance}

Figure \ref{fig:modelling_and_results}b shows the results of the model fitting procedure for spatial point processes using the Judd dataset \cite{Judd2009}.
The total bar heights are the optimised performance of each model (in bits/fixation) relative to the mutual entropy null model.

The gold standard model shows that the total mutual information between the image and the spatial structure of the fixations amounts to \SI{\MutualInformationImageSpatial}{\bits\per\fix}.
To give another intuition for this number, a model that would for every fixation always correctly predict the quadrant of the image in which it falls would also have a likelihood of \SI{2}{\bits\per\fix}. 

The lower-bound model is able to explain \SI{\BaselinePerformance}{\bits\per\fix} of this mutual information.
That is, \SI{\BaselinePercentOfGold}{\percent} of the information in spatial fixation distributions can be accounted for by behavioural biases (e.g. the bias of human observers to look at the centre of the image).

The \BestModel\ model performs best of all the saliency models compared, with \SI{\BestModelPerformance}{\bits\per\fix}, capturing \SI{\BestModelPercentOfGold}{\percent} of the total mutual information.
It accounts for \SI{\BaselineToBestPercentOfGold}{\percent} more than the lower-bound model, or \SI{\BestModelPercentOfGoldRelBaseline}{\percent} of the possible information gain (\SI{\PossibleInformationGain}{\bits\per\fix}) between baseline and gold standard.
Of the approximately \SI{\BestModelPerformance}{\bits\per\fix} captured by the \BestModel\ model, \SI{\BestModelNonlinearityPercent}{\percent} of this is due to the optimised nonlinearity, \SI{\BestModelNonlinearityToCenterbiasPercent}{\percent} is added by the centre bias, and \SI{\BestModelCenterbiasToBlurPercent}{\percent} by spatial blurring.

Considering only model performance (i.e. without also including centre bias and blur factors; the pink sections in Figure \ref{fig:modelling_and_results}b) shows that many of the models perform worse than the lower-bound model.
This means that the centre bias is more important than the portion of image-based saliency that these models do capture \cite{Tatler2005}.

Readers will also note that the centre bias and blurring factors account for very little of the performance of the Judd model relative to most other models.
This is because the Judd model already includes a centre bias that is optimised for the Judd dataset.

\subsection{Image-level comparisons}

\begin{figure*}[!ht]
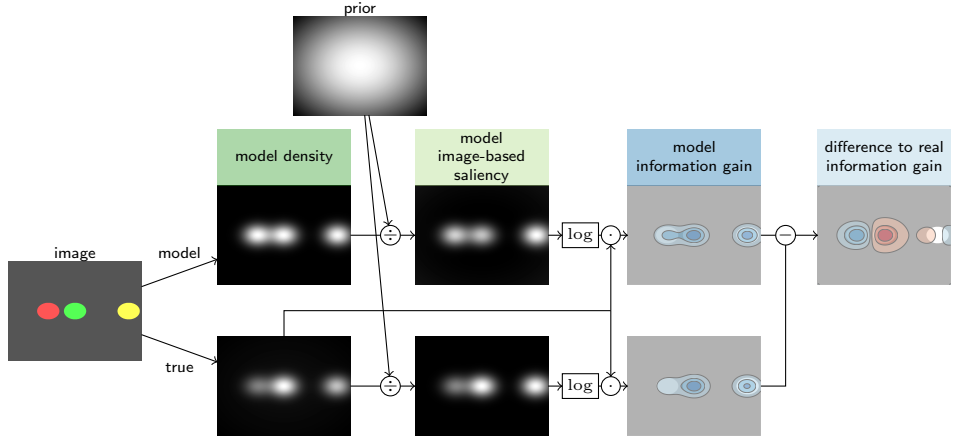

  \center
  \include{figures/information_gains_schema.pgf}
  \caption{Calculation of information gain in the image space.
  The first column shows an example image.
  Hypothetical fixation densities of the gold standard (``true'') and model predictions are shown in the second column.
  These are divided by the baseline model (prior) to get the image-based saliency map.
  Both saliency maps are then log-transformed and multiplied by the gold standard density to calculate the information gain in the image space.
  Subtracting the gold standard information gain from the model's information gain yields a difference map of the possible information gain: that is, 
  where and by how much the model's predictions fail.
  In this case, the model overestimates (blue contours) the fixation density in the left (red) spot in the image, underestimates (red contours) the centre bias and predicts the right-most (yellow) spot almost perfectly.
  }
\end{figure*}

\begin{figure*}
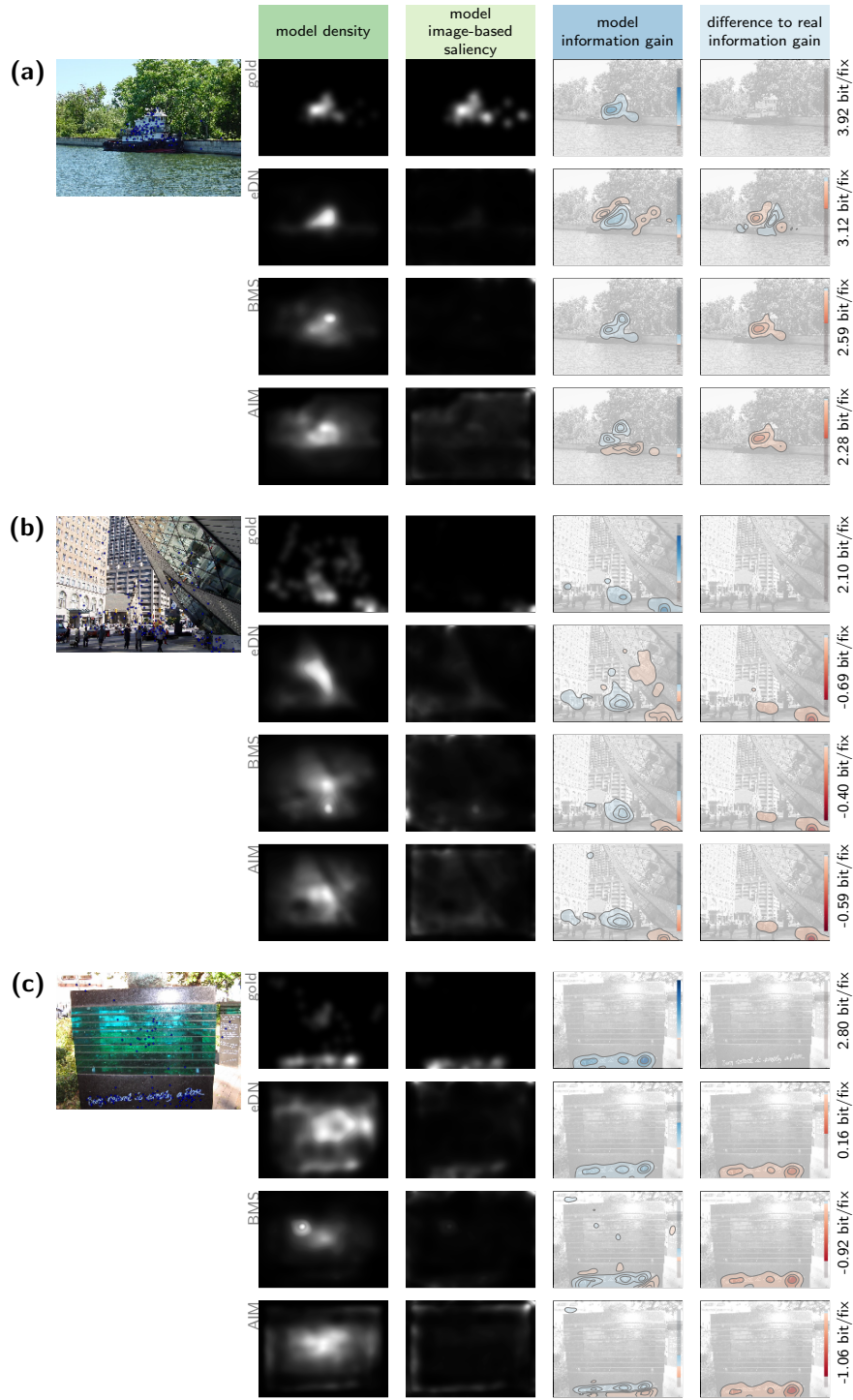

   \center
   \include{figures/information_gains.pgf}
  
\caption{
Information gain in the image space.
\textbf{(a)} The left column displays the image for which the eDN model explains most of the information gain.
The second column shows the model density (where the model expects fixations to be), for the gold standard, eDN, BMS and AIM in each row.
The third column shows the ratio of the model density to the baseline density (where the model believes the image-based saliency to be).
The fourth column shows the information gain of the model relative to the baseline.
Contours are equally spaced and consistent over the column (more dense contour lines indicate a peak in information gain).
The fifth column shows how this differs from the possible information gain estimated by the gold standard.
The log-likelihood of the model for each image is shown to the right of the fifth column.
\textbf{(b)} Same as (a) for the image on which the AIM and BMS models perform best.
\textbf{(c)} Same as (a) for the image on which the AIM and BMS models perform worst.
}
\label{fig:information_gain_images}
\end{figure*}

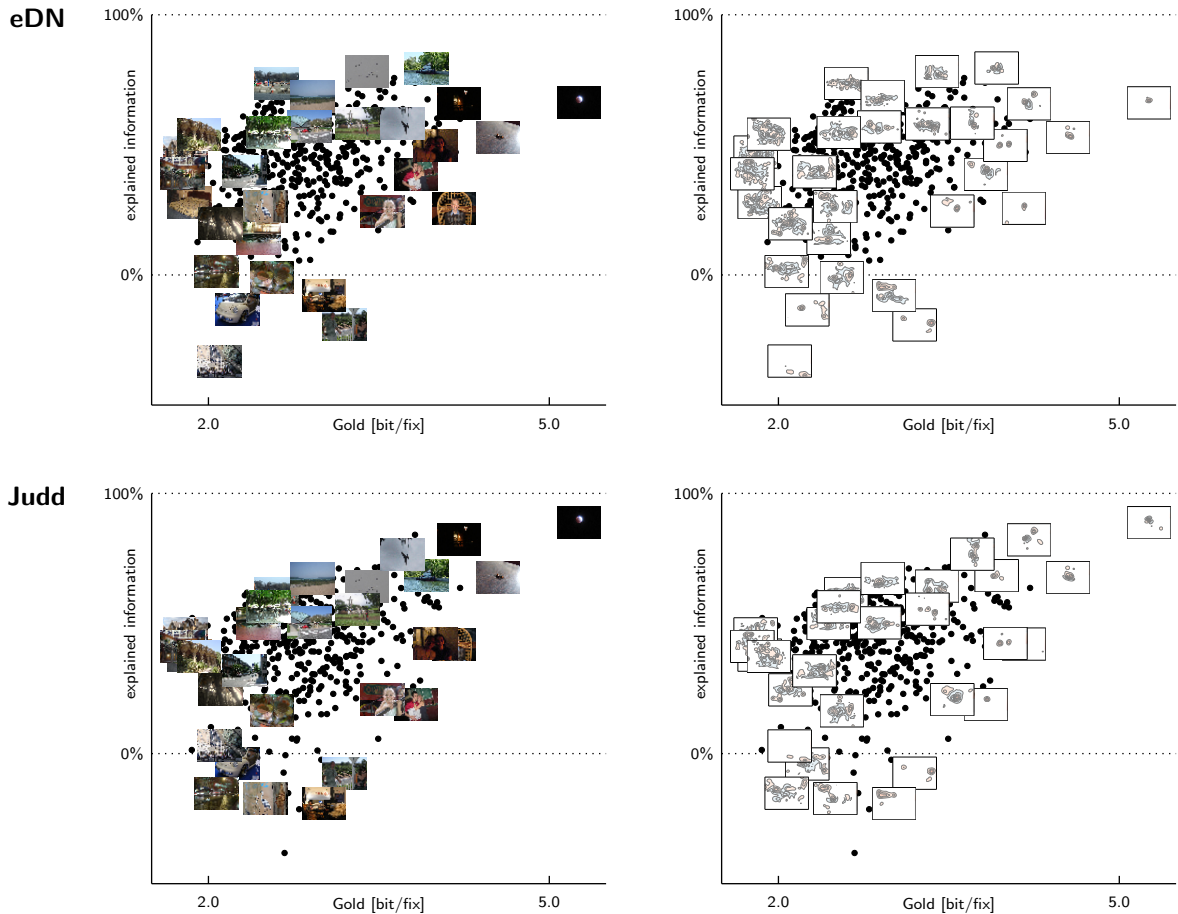
\begin{figure*}[!ht]

  \input{figures/information_gains_scatter_combined.pgf}

  \caption{
    Distribution of information gains and explained information over
    all images in the dataset. 
    Each black dot represents an image from the dataset. 
    For some example cases, in the left column
    the images themselves are shown. In the right column for these cases
    the information gain difference to the gold standard is shown.
    These plots allow a modeller to assess the performance of a model on all images in the dataset.
    In the lower right of the plot are images where a lot of information could be explained but is not; these are where the model could be best improved.
  }
  \label{fig:information_gain_scatter}
\end{figure*}

One major advantage of using a probabalistic framework rather than AUC is that the information gain of a model can be compared with the gold standard in the image space.
Here we visualise where the information that saliency models fail to capture is located in several images.

To choose example images in a principled way, we computed the images for which the best model (\BestModel) explained most and least of the possible information gain. 
We additionally included the image that the two second and third ranking models (BMS and AIM) performed worst on (this happened to be the same image).

First, we plot the model density for each model (second column in Figure \ref{fig:information_gain_images}). 
This is $\hat p(x,y \mid I)$, and shows where the model expects fixations to occur in the given image.
Then we plot the model's prediction of image-based saliency (see above) in the third column of Figure \ref{fig:information_gain_images}.
It tells us where and how much the model believes the fixation distribution in a given image is different from the prior $p(x,y)$ (centre bias).
If the ratio is greater than one, the model predicts there should be more fixations than the centre bias expects.

Now we wish to understand, for each point in the image, how much more the model knows about the fixations than the centre bias.
This can be achieved by separating the information gain (an integral over space) into its constituent pixels (shown in Figure \ref{fig:information_gain_images} column four), as $p(x,y \mid I) \log(\hat p(x,y \mid I) / p(x,y))$.
The information gain quantifies how much more efficient the model is in describing the fixations than the centre bias, by quantifying how much channel capacity it could save (in \si{\bits\per\fix}).
These plots quantify \textit{where} in the image the information is saved.
Note also that weighting by the gold standard $p(x,y \mid I)$ results in a weaker penalty for incorrect predictions in areas where there are fewer fixations.

Finally, the fifth column in Figure \ref{fig:information_gain_images} shows the difference between the model's information gain and the possible information gain, estimated by the gold standard, resulting in $p(x,y \mid I) \log(\hat p(x,y \mid I) / p(x,y \mid I))$.
It tells us where and by how much the model's belief is wrong: where and how much information (bits) is being wasted that could be used to describe the fixations more efficiently. 
If the contour plots in the last column of Figure \ref{fig:information_gain_images} are integrated over the image, we get exactly the negative image-based Kullback-Leibler divergence. 
The advantage of our approach is obvious: we can see not only how much a model fails, but exactly where it fails.
We believe this metric can be used to make informed decisions about how to improve a saliency model.

For example, in Figure \ref{fig:information_gain_images}c, it is important to capture the text in the lower part of the image.
The BMS and AIM models fail to capture this difference, while eDN does noticably better.
In addition, all models in these images appear to underestimate the spread of fixations.
Take for example the first image: the eDN model's prediction could be improved by placing more density around the ship rather than on it.

To extend this image-based analysis to the level of the entire dataset, we display each image in the dataset according to its possible information gain and the percentage of that information gain explained by the model (Figure \ref{fig:information_gain_scatter}).
In this space, points to the bottom right represent images that contain a lot of explainable information in the fixations that the model fails to capture.
Points show all images in the dataset, and for a subset of these we have displayed the image itself (left column) and the information gain difference to the gold standard (right column).
For the eDN model (top row), the images in the bottom-right of the plot tend to contain human faces.
The Judd model contains an explicit face detection module, and as can be seen in the bottom row of Figure \ref{fig:information_gain_scatter}, it tends to perform better on these images.
In terms of the whole dataset however, the eDN model generally performs better on images with a moderate level of explainable information (around 3 bit/fixation).

\section{Discussion}

\subsection{How close are we to understanding image-based saliency?}

We have evaluated a number of saliency models within the framework of point processes using log-likelihoods.
For the dataset examined here, the total amount of mutual information that can be extracted from an image about the spatial structure of fixations was 2.1 bits/fixation.
This represents all variation in fixation patterns that is consistent between observers (the gold standard model); a better prediction is not possible without additional knowledge about inter-observer differences in fixation behaviour.
The best saliency model (\BestModel) accounts for \SI{\BestModelPercentOfGold}{\percent} of this total information, while a model that ignores image content and captures only observers' centre bias accounts for \SI{\BaselinePercentOfGold}{\percent} of this information.
Partialling out the mutual information explained by spatial behavioural biases shows that the best existing saliency models explain only \SI{\BestModelPercentOfGoldRelBaseline}{\percent} of the mutual information. 
To examine the generality of these findings across eye movement datasets, we also ran our analysis on the Kienzle dataset \cite{Kienzle2009}, which has been designed to remove photographer bias (placing salient objects in the centre of the photograph) such that salient points are roughly equally distributed across the image.
In this dataset (see see Appendix \ref{sec:app_kienzle}), even less \SI{\KienzleBestModelPercentOfGoldRelBaseline}{\percent} of the possible information gain is covered by the best model (here, \KienzleBestModel).
Our results show that there remains a significant amount of information that image-based saliency models could explain but do not.

In order to improve models it is useful to know where this unexplained information is located.
We developed methods to assess not only model performance on a database level, but also to show where and by how much model predictions fail in individual images (Figures \ref{fig:information_gain_images} and \ref{fig:information_gain_scatter}).
We expect these tools will be useful for the model development community, and provide them in our free software package.



We compared a number of models on their performance over two datasets.
While the primary goal of this paper was to show the distance between state-of-the-art models and the gold standard, benchmarking and model comparison is important to gauge progress \cite{Borji2012,Judd2012,Borji2013d,Borji2013b}.
Happily, the ranking of models by their log-likelihoods does not differ substantially from rankings according to AUC (see Appendix, Figure \ref{fig:metric_comparison}), indicating that the results of previous model comparison efforts would be unlikely to change qualitatively under log-likelihoods.
Furthermore, metrics like AUC and KL-divergence are highly correlated with log-likelihoods if correctly computed (though often they are not; see Appendix \ref{sec:kl_divergence} and Figure \ref{fig:metric_comparison}).
Nevertheless, log-likelihood is the only metric that is linear in information, and therefore the only one we could use to answer the question ``how close are we'':
having a linear information scale allows us to judge not only the ranking between models, but quantitatively how much one model is better than another.

\subsection{Point processes and probabalistic modelling}

It is appealing to evaluate saliency models using log-likelihoods because it is mathematically principled: it is the standard of all statistical model comparison.
When modeling fixations, point processes are the appropriate probability distributions to use \cite{Barthelme2013}.
While in the traditional framework the saliency map output by a model gives values on some unknown scale, considering fixation selection as a point process allows a precise definition of a saliency map: $\frac{p(x,y\mid I)}{p(x,y)}$ (see Section \ref{sec:saliency_definition}).
Previously saliency was loosely defined as a spatial combination of image features that can predict eye fixation patterns.
We make this connection precise.

We suggest that the biggest practical advantage in using a probabalistic framework is its generality.
Once a model is formulated in a probabalistic way many kinds of ``task performance'' can be calculated, depending on problems of applied interest.
The AUC is one such task. 
It is the performance of a model in a 2AFC task ``which of these two points was fixated?''.
There are other cases where different tasks may be a better metric for model evaluation than log-likelihood, depending on the application.
For example, we might be interested in whether humans will look at an advertisement on a website, or whether the top half of an image is more likely to be fixated than the bottom half.
These predictions are a simple matter of integrating over the probability distribution.
Without a definition of the scale of saliency values this type of evaluation is not well defined.
In addition, a probabalistic model allows the examination of any statistical moments of the probability distribution that might be of practical interest.
For example, Engbert et al \cite{Engbert2014} examine the properties of second-order correlations between fixations in one scanpath.
Using a probabalistic framework does not restrict us from using any type of task-based evaluation metric; on the contrary, it enables the evaluation of multiple tasks for the same model.

Using information to evaluate saliency models has several other desirable properties.
First, it allows the contribution of different factors in explaining data variance to be quantified. 
For example, it is possible to show how much the centre bias contributes to explaining fixation data independent of image-based saliency contributions, as we have done here.
Second, it is differentiable in the point process density, allowing models to be numerically-optimised using off-the-shelf techniques like gradient descent or quasi-newton methods.
Third, it is readily extended to include other factors such as temporal effects or task dependencies, the first of which we do below.

\subsection{Extension to temporal effects} 

\begin{figure*}
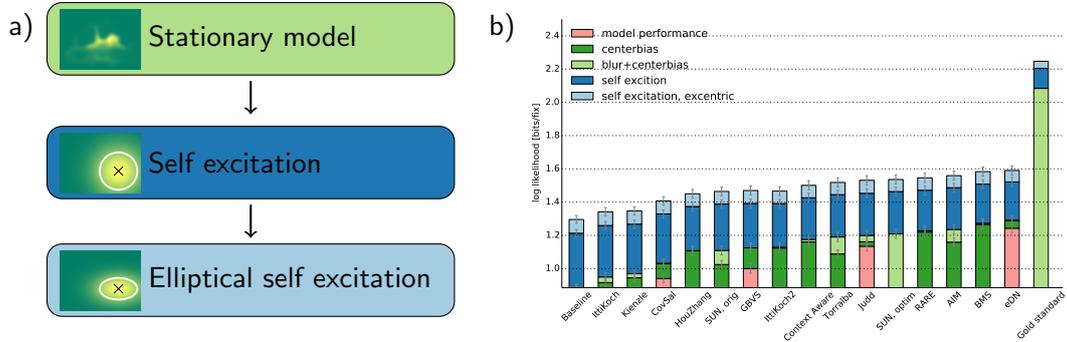

  \center
  \include{figures/modeling_temporal.pgf}

  \caption{
  \small
Temporal modelling and results.   
\textbf{(a)} For each optimised spatial saliency model (from Figure \ref{fig:modelling_and_results}) we optimised two temporal image-independent factors: self-excitation captures the tendency of observers to fixate in a circular region nearby their previous fixation point, and aspect-optimised self-excitation is where we additionally optimised the shape of the self-excitation process.
\textbf{(b)} Log likelihood (bits/fixation) for each model considered relative to a null model which assumes uniform fixation density over the image. 
The total bar height represents the optimised performance of each model; the coloured sections show the relative contributions of each model factor from (a).
Note the y-axis has been cropped to the raw baseline model performance.
All green bars show the fully-optimised spatial models from Figure \ref{fig:modelling_and_results}, blue bars show the additional influence of temporal factors.
}
\label{fig:temporal_modelling_and_results}
\end{figure*}

The saliency field has recognised that free viewing static images is not the most representative condition to study naturalistic eye movement behaviour \cite{Tatler2008, Tatler2009, Tatler2011, Ehinger2009, Dorr2010}.
Understanding image based saliency is not only a question of ``where?'', but of ``when?'' and ``in what order?''.
It is the spatiotemporal pattern of fixation selection that is increasingly of interest to the field, rather than purely spatial predictions of fixation locations.

The point process framework we outline in this paper is easily extended to study spatiotemporal effects.
We demonstrate this by modelling a simple temporal correlation between fixations.
Figure \ref{fig:temporal_modelling_and_results} shows the model fitting with the additional inclusion of temporal effects (see Appendix \ref{sec:temporal_effects_methods} for model fitting details).
With \SI{\BaselineSelfExcitationPerformance}{\bits\per\fix}, the baseline model including this optimised temporal component predicted fixations better than the best saliency model (\BestModel) without temporal effects (\SI{\BestModelPerformance}{\bits\per\fix}).
While the gold standard shows that there is still a lot of spatial structure in the fixations left to explain, currently a simple image agnostic model of eye movement dynamics performs better than the best saliency maps.

Optimising the magnitude, scale and shape of the temporal component produced a weighting that assigned higher likelihoods to points near to the last fixation.
That is, human observers were more likely to look near to where they had previously fixated (see also \cite{Hooge2005, Tatler2008, Smith2009, Engbert2014}).
Furthermore, the temporal weighting had an elliptical shape with a bias towards horizontal saccades improved performance (see also recent work by \cite{Clarke2014}).
This is consistent with previous literature showing that the spatial distribution of saccades when viewing natural scenes can be horizontally biased \cite{Tatler2009, Dorr2009, Dorr2010}.

``Inhibition of return'' has been posited as a biologically-plausible way to select new fixations \cite{Klein2000, Itti2000, Itti2001, Borji2013d}.  
In part this was necessitated by the decision stages of these models (winner-takes-all): since the saliency map for a given static image does not change, these models could only predict a single fixation without some mechanism to avoid the point of maximum saliency.
Framing saliency maps probabalistically does not require this mechanism because the probability of fixating a given point of the image is usually much less than one, meaning that the spatial distribution of fixations in an image will tend to spread over the image rather than remaining at the location of maximum saliency.
Our results examining temporal correlations show that a local self-reinforcement effect provides a more complete description of human eye movement behaviour for this dataset.

Finally, this self-reinforcement effect reveals that fixation selection is an example of a \textit{self-exciting point process}.
Self-exciting point processes have already proven a powerful modelling tool for a variety of other spatial phenomena, including earthquakes \cite{Ogata1998}, civilian deaths in Iraq \cite{Lewis2010}, and burglaries in Los Angeles \cite{Mohler2011}.
An awareness of this connected literature may lead to new ideas for approaches to saliency models in the future.

\subsection{Conclusions}

Our results show that almost two-thirds of the information (once behavioural biases are partialled out) in fixations that \textit{could} be explained purely from the image still remains to be explained by image-based saliency models.
The use of log-likelihoods to evaluate and optimise saliency models holds a number of advantages over existing metrics. 
To facilitate the use of this metric, we will make a free and open source software framework available (\url{www.bethgelab.org}).

Of course, accounting for the entirety of human eye movement behaviour in naturalistic settings will require incorporating information about the task, high-level scene properties, and mechanistic constraints on the eye movement system \cite{Tatler2011, Tatler2009, Ehinger2009, Engbert2014}.
Our gold standard contains the influence of high-level (but still purely image-dependent) factors to the extent that they are consistent across observers.
Successful image-based saliency models will therefore need to use such higher-level features, combined with task relevant biases, to explain how image features are associated with the spatial distribution of fixations over scenes.

\begin{figure}
  \begin{center}
    \includegraphics[width=7cm]{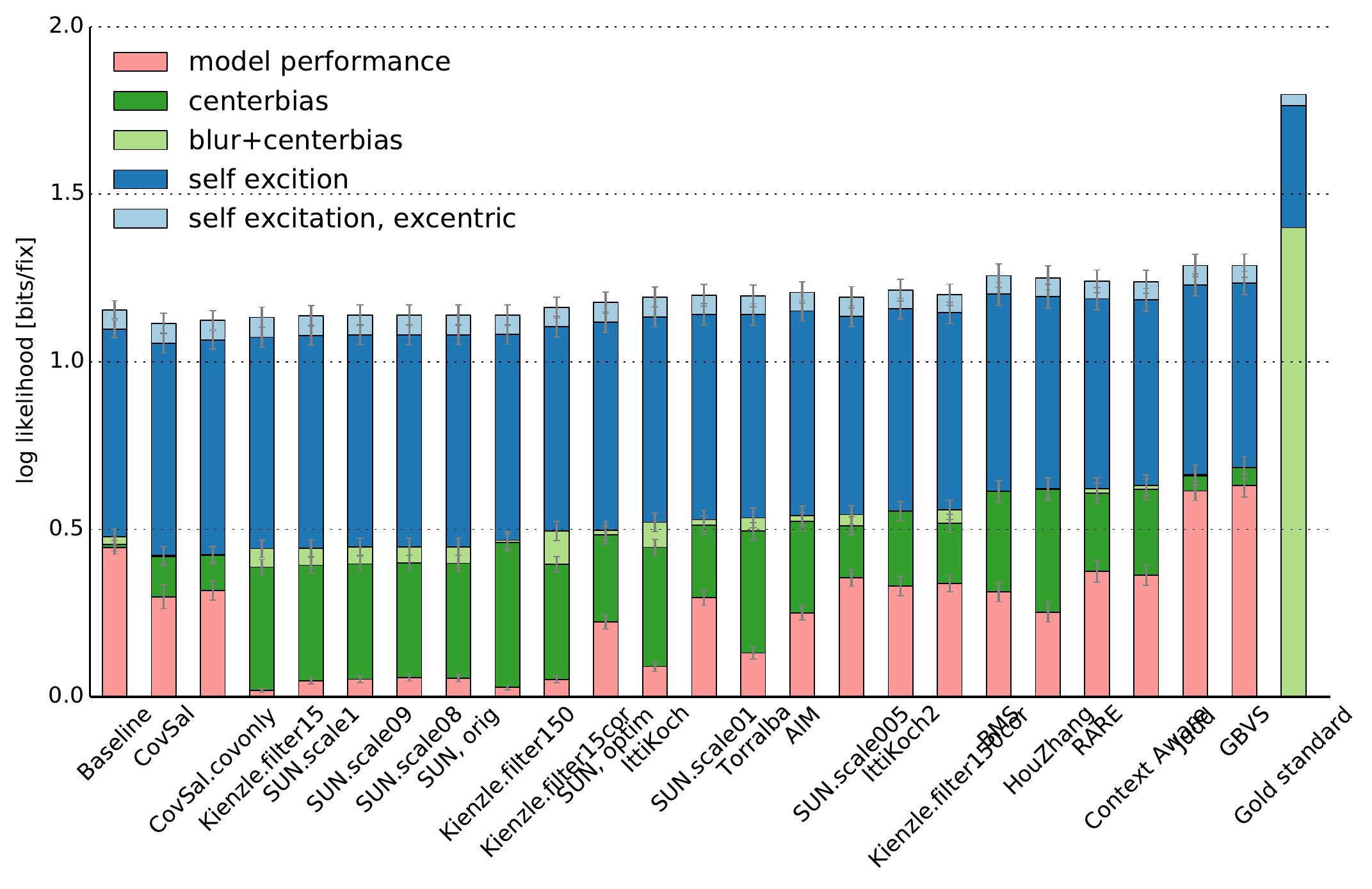}
  \end{center}
  \caption{
	Spatial and temporal results (as in Figure \ref{fig:temporal_modelling_and_results}) assessed on the Kienzle dataset.  
  }
  \label{fig:kienzle_results}
\end{figure}

\newpage

\section{Appendices}

\subsection{Kienzle data set}
\label{sec:app_kienzle}

We repeated the full evaluation on the dataset of Kienzle et al in \cite{Kienzle2009}. It consists of 200 grayscale images of size $1024\times678$ and $15$ subjects. This dataset is of special interest, as the authors removed the photographer bias by using random crops from larger images. The results are shown in Figure \ref{fig:kienzle_results}.

In this dataset, with \SI{\KienzleBestModelPercentOfGoldRelBaseline}{\percent} even less of the possible information gain is covered by the best model (here, \KienzleBestModel\footnote{Note that we were not yet able to include eDN into this comparision, as the source code was not yet released at the time of the analysis.}).
Removing the photographer bias leads to a smaller contribution (\SI{\KienzleBaselinePercentOfGold}{\percent}) of the nonparametric model compared to the increase in log-likelihood by saliency map based models. The possible information gain is with \SI{\KienzlePossibleInformationGain}{\bits\per\fix} smaller than for the Judd dataset (\SI{\PossibleInformationGain}{\bits\per\fix}) There are multiple possible reasons for this. Primarily, this dataset contains no pictures of people, but a lot of natural images. In addition, the images are in grayscale.
The increase by considering simple temporal correlations is still higher than the increase of the best spatial saliency map model, the effect is even stronger than in the Judd dataset.

\subsection{Other evaluation metrics}

Here we consider the relationship between log-likelihoods and prominent existing saliency metrics: AUC and KL-divergence.

\label{sec:app_other_metrics}

\subsubsection{AUC}

\begin{figure*}[!ht]
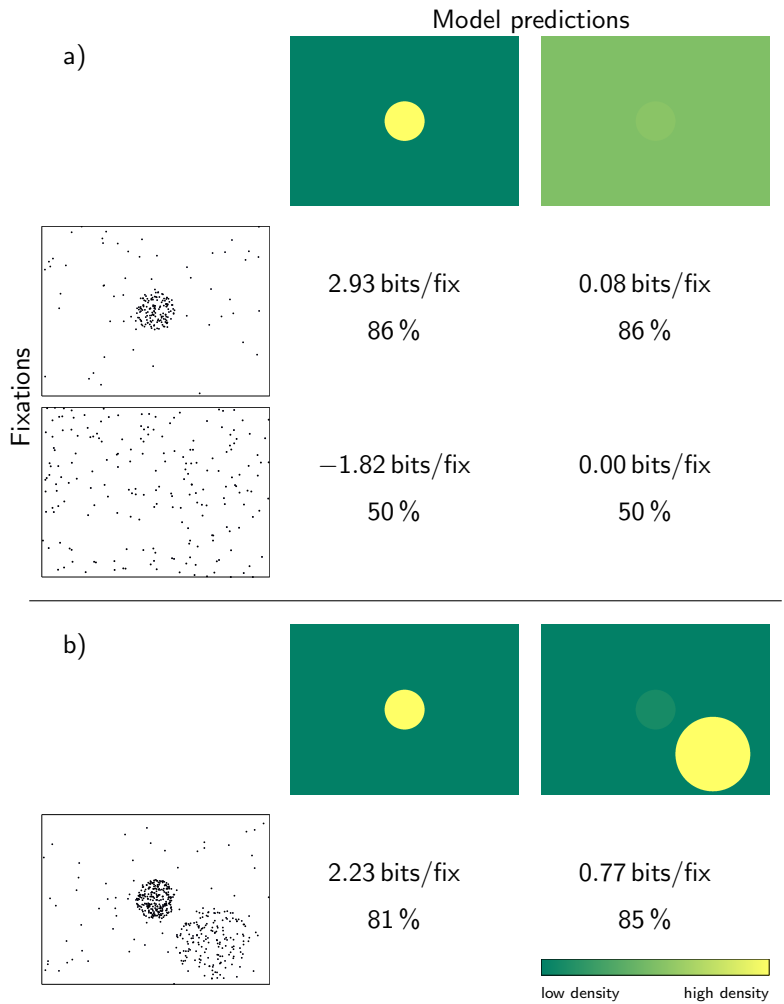

  \center
  \include{figures/auc_problems.pgf}
  \caption{
  \small
The AUC is invariant to monotonic transformations of the saliency maps, and this can produce counterintuitive results. 
Several hypothetical saliency maps (coloured images in columns) and fixation distributions (point scatters in rows) are shown in a grid; 
each cell shows the log-likelihood (bits/fixation) and AUC score (\%) for that model/fixation combination.
\textbf{(a)} Two saliency maps with the same spatial pattern of predictions but different prediction strengths.
The two fixation distributions are sampled from the first and second models respectively.
The first model makes a strong prediction that fixations should cluster in the centre; when they do (first row), this model receives a much higher likelihood than the alternative model.
When the prediction is wrong and fixations are nearly uniformly distributed across the image (second row), the first model is penalised with a much lower likelihood.
AUC does not differentiate these models because it is insensitive to prediction strength, only rank order matters for the performance.
\textbf{(b)} The property of AUC demonstrated in (a) can lead to counterintuitive results.
If higher saliency values are intuitively associated with more fixations then the left model is the better model.
Counterintuitively, the left model has a lower AUC score than the right model. 
Likelihoods are more in line with our intuitions in this case.
}
\label{fig:auc_problems}
\end{figure*}

The most prominent metric used in the saliency literature is the area under the receiver operating characteristic curve (AUC).
The AUC is the area under a curve of model hit rate against false positive rate for each threshold.
It is equivalent to the performance in a two-alternative forced-choice (2AFC) task where the model is ``presented'' with two image locations: one at which an observer fixated and another from a \textit{nonfixation distribution}.
The thresholded saliency value is the model's decision, and the percent correct of the model in this task across all possible thresholds is the AUC score.
The different versions of AUC used in saliency research differ primarily in the nonfixation distribution used. 
This is usually either a uniformly-selected distribution of not-fixated points across the image (e.g. in \cite{Judd2012}), or the distribution of fixations for other images in the database (the shuffled AUC, \cite{Tatler2005, Borji2013b, Borji2013d}).
The latter provides an effective control against \textit{centre bias} (a tendency for humans to look in the centre of the screen, irrespective of the image content), assuming both fixation and nonfixation distributions have the same image-independent bias\footnote{It is important to bear in mind that this measure will penalise models that explicitly try to model the centre bias.}.
The AUC therefore depends critically on the definition of the nonfixation distribution.
In the case of the uniform nonfixation distribution, AUC is tightly related to \textit{area counts}: optimizing for AUC with uniform nonfixation distribution is equivalent to finding for each percentage $0\leq r \leq 100$ the area consisting of $r\%$ of the image which includes most fixations \cite{Barthelme2013}.

One characteristic of the AUC that is often considered an advantage is that it is sensitive only to the rank-order of saliency values, not their scale (i.e. it is invariant under monotonic pointwise transformations) \cite{Tatler2005}.
This allows the modelling process to focus on the shape (i.e. the geometry of iso-saliency points) of the distribution of saliency without worrying about the scale, which is argued to be less important for understanding saliency than the contour lines \cite{Tatler2005}.
However, in certain circumstances the insensitivity of AUC to differences in saliency can lead to counterintuitive behaviour, if we accept that higher saliency values are intuitively associated with more fixations (Figure \ref{fig:auc_problems}).

By using the likelihood of points as a classifier score, one can compute the AUC for a probabilistic model just as for saliency maps. 
This has a principled connection with the probabilistic model itself: 
if the model performed the 2AFC task outlined above using maximum likelihood classification, then the model's performance is exactly the AUC.
Given the real fixation distribution, it can also be shown that the best saliency
map in terms of AUC with uniform nonfixation distribution is exactly the gaze
density of the real fixation. 
However, this does not imply that a better AUC score will yield a better log-likelihood or vice versa.

For more details and a precise derivation of these claims, see \cite{Barthelme2013}. In Figure \ref{fig:metric_comparison} we compare our log-likelihood evaluation to AUC performances.

\subsubsection{Kullback-Leibler divergence}
\label{sec:kl_divergence}

\begin{figure}[!ht]
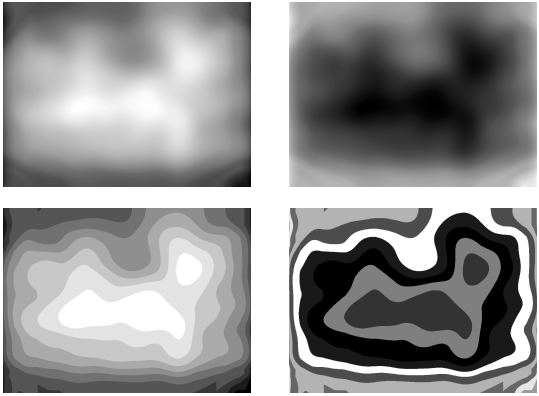

  \center
  \include{figures/kl_divergence.pgf}
  \caption{Fixation-based Kullback-Leibler divergence for saliency maps.
  The top-left image shows a real saliency map (from eDN); the top-right is inverted,
  bottom-left is the same map with binned saliency values, and in the bottom-right map, the saliency assigned to 
  each bin is shuffled.
  These maps have identical fixation-based KL-divergence (and very different log-likelihoods).
  }
  \label{fig:kl_divergence}
\end{figure}

\begin{table*}[!ht]
  \center
\include{kl_divergence_table}
  \caption{
	Papers using KL-divergence to evaluate saliency models. We describe what was used as the estimate of the true distribution for image-based KL-divergence. 
  }
  \label{tab:kl_divergence}
\end{table*}

Kullback-Leibler divergence (KL-divergence) is tightly related to
log-likelihoods.
However, KL-divergence as used in practice in the saliency literature
is not the same as the approach we advocate.

In general, the KL-divergence between two probability distributions
$p$ and $q$ is given by
\[
 D_{KL}[p \| q] = \int \log\left(\frac{p(x)}{q(x)}\right)p(x)dx,
\]

and is a popular measure of the difference between two probability
distributions.
In the saliency literature, there are at least two different model
comparison metrics that have been called ``Kullback-Leibler divergence''.
Thus, when a study reports a KL metric, one needs to check how this
was computed.
The first variant treats the saliency map as a two dimensional
probability distribution and computes the KL-divergence between this
predicted distribution and the empirical density map of fixations
\cite{Tatler2005, Wilming2011}; we will call this \textit{image-based
KL-divergence}.
The second metric referred to as ``Kullback-Leibler divergence'' is
the KL-divergence between the distribution of saliency values at
fixations and the distribution of saliency values at nonfixation
locations; we will call this \textit{fixation-based KL-divergence}
\cite{itti2005bayesian}.
This is calculated by binning 
the saliency values at 
fixations and nonfixations into a
histogram  and then computing the KL-divergence of these histograms.
Like AUC, it depends critically on the definition of the nonfixation
distribution, and additionally on the histogram binning.
In Table \ref{tab:kl_divergence} we list a number of papers using one
of these two definitions of KL-divergence.

We now precisely show the relationship between these measures and our
information theoretic approach. Very generally, Information theory can be derived from the task of assigning code words to different events that occur with different probabilities such that their average code word length becomes minimal. It turns out that the negative log-probability is a good approximation to the optimal code word length possible which gives rise to the definition of the log-loss:
$$
l(x) = - \log p(x)
$$
In case of a discrete uniform distribution $p(x)=\frac{1}{n}$ the log-loss for any possible $x$ is simply $\log n$, i.e. the log of the number of possible values of $x$. Accordingly, the more ambiguous the possible values of a variable are, the larger its average log-loss which is also known as its entropy 
$$
H[X]= \EE[-\log p(x)]
$$
If $p(x)$ denotes the true distribution which accurately describes the variable behavior of $x$ and we have a model $q(x)$ of that distribution then we can think of assigning code words to different values of $x$ that are of length $-\log q(x)$ and compute the average log-loss for the model distribution
\begin{align*}
  &\EE[-\log q(x)] \\
  &\quad = - \int p(x) \log q(x) dx\\
  &\quad = H[X] + D_{KL}[p(x)||q(x)]
\end{align*}
That is the KL-divergence measures how much the average log-loss of a model distribution $q(x)$ exceeds the average log-loss of the true distribution. The KL-divergence is also used to measure the information gain of an observation if p(x) denotes a posterior distribution that correctly describes the variability of $x$ after the observation has been made while q(x) denotes the prior distribution. In a completely analog fashion we can measure how much more or less information one model distribution $q_1(x)$ provides about $x$ than an alternative model $q_2(x)$ does by computing how much the average log-loss of model 1 is reduced (or increased) relative to the average log-loss of model 2. This can also be phrased as an expected log-likelihood ratio \footnote{The concept of log-likelihood ratios is familiar to readers with knowledge of model comparison using e.g. $\chi^2$ tests.} (ELLR):
\begin{align*}
ELLR & = \EE[- \log q_2(x)] - \EE[- \log q_1(x)]\\
& =  \EE[\log q_1(x)] - \EE[ \log q_2(x)]\\
& = \int p(x) \log \frac{q_1(x)}{q_2(x)} dx.
\end{align*}
In other words, very generally, the amount of information model 2 provides about a variable relative to model 1 can be measured by asking how much more efficiently the variable can be encoded when assuming the corresponding model distribution $q_2(x)$ instead of $q_1(x)$ for the encoding. Note, that this reasoning does not require any of the two model distributions to be correct. For example, in the context of saliency maps we can ask what the best possible model distribution is which does not require any knowledge of the actual image content. This baseline model can capture general biases of the subjects such as the center bias. In order to evaluate the information provided by a saliency map that can be assigned to the specific content of an image we thus have to ask how much more the model distribution of that saliency model provides relative to the baseline model.

Our information gain metric reported in the paper is exactly the ELLR,
where $q_1$ is the model, $q_2$ is the baseline, and we estimated the
expectation value using the sampling estimator.
The ELLR can be rewritten as a difference between KL-divergences:

\begin{align*}
\text{ELLR} :=& \EE[ \log (q_1(x)/q_2(x))]\\
            =& \EE[ \log q_1(x)] - \EE[\log q_2(x)]\\
            =& \DKL[p(x) \| q_2(x)] - \DKL[p(x) \| q_1(x)]
\end{align*}

This naturally raises the question: is our measure equivalent to the
KL-divergence that has been used in the saliency literature?
The answer is no.


It is crucial to note that in the past the scale used for saliency
maps was only a rank scale.
This was the case because AUC was the predominant performance measure,
and is invariant under such transformations.
That is, two saliency maps $S_1(x)$ and $S_2(x)$ were considered
equivalent if a strictly monotonic increasing function $g: \RR \to
\RR$ exists such that $S_1(x)$ = $g(S_2(x))$.
In contrast, in the equation for ELLR, the two distributions $q_1$ and
$q_2$ are directly proportional to the saliency map times the center
bias distribution, and well-defined only if the scale used for
saliency maps is meaningful.
In other words, if one applies a nonlinear invertible function to a
saliency map the ELLR changes.

\textit{Fixation-based KL-divergence} is the more common variant in the literature: researchers wanted to apply information theoretic measures to saliency evaluation while remaining consistent with the rank-based scale of AUC \cite{itti2005bayesian}.
Therefore they did not interpret saliency maps themselves as probability distributions, but applied the KL-divergence to the distribution of saliency values obtained when using the fixations to that obtained when using non-fixations.
We emphasize that this measure has an important conceptual caveat:
rather than being invariant under only monotonic increasing transformations, KL-divergence is invariant under \textit{any} reparameterisation.
This implies that the measure only cares about which areas are of equal saliency, but does not care which of any two areas is actually the more salient one.
For illustration, for any saliency map $S(x, y)$, its negative counterpart $\bar S(x,y) := \sup(S) - S(x,y)$ is completely equivalent with respect to the fixation-based KL metric, even though for any two image regions $\bar S$ would always make the opposite prediction about their salience (see Figure \ref{fig:kl_divergence}a).
Furthermore, the measure is sensitive to the histogram binning used, and in the limit of small bin width all models have the same KL-divergence: the model-independent KL-divergence between $p(\xfix)$ and $p(\xnonfix)$.

\textit{Image-based KL-divergence} requires that the saliency maps are interpreted as probability distributions.
Previous studies employing this method (Table \ref{tab:kl_divergence}) simply divided the saliency values by their sum to obtain such probability distributions. 
However, they did not consider that this measure is sensitive to the scale used for the saliency maps.
Optimization of the pointwise nonlinearity (i.e. the scale) has a huge effect on the performance of the different models (see below in \ref{sec:app_metric_comparison}).
More generally, realising that image-based KL-divergence treats saliency maps as probability distributions means that other aspects of density estimation, like centre bias and regularisation strategies (blurring), must also be taken into account.

The only conceptual difference between image-based KL-divergence and log-likelihoods is that for estimating expected log-likelihood ratios, it is not necessary to have a gold standard.
One can simply use the unbiased sample mean estimator (see \ref{sec:app_estimation_considerations}).
Furthermore, by conceptualising saliency in an information-theoretic way, we can not only assign meaning to expected values (such as ELLR or DKL) but we also know how to measure the information content of an individual event (here, a single fixation) using the notion of its log-loss (see our application on the individual image level in Figure \ref{fig:information_gain_images}).
Thus, while on a theoretical level log-likelihoods and image-based KL-divergence are tightly linked, on a practical level a fundamental reinterpretation of saliency maps as probability distributions is necessary \ref{sec:app_metric_comparison}.


\subsubsection{Metric comparison}
\label{sec:app_metric_comparison}

\begin{figure*}[!ht]
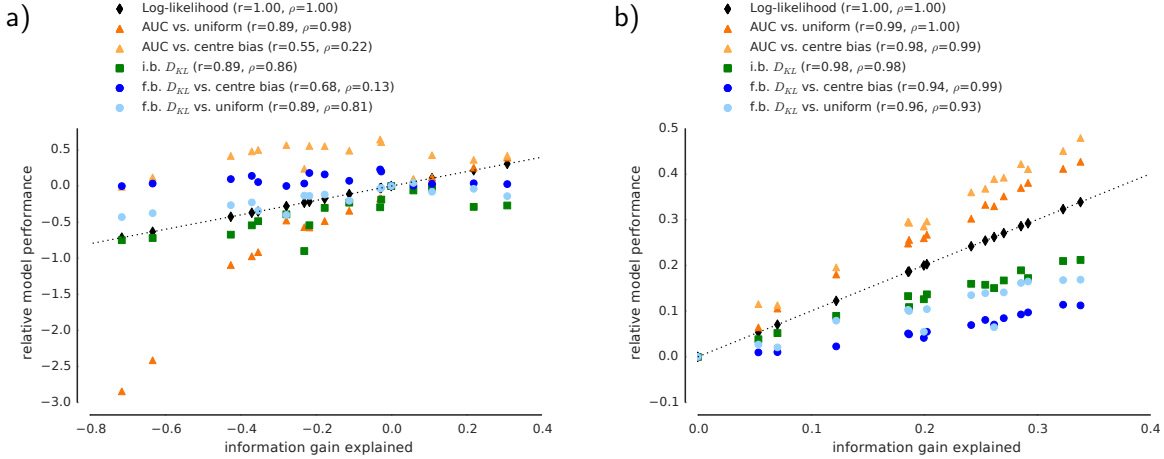

  \center
  \include{figures/other_measures.pgf}
  \caption{Possible information gain explained (x-axis) compared to evaluation metrics (y-axis). 
  All metrics have been rescaled to yield zero for the baseline and one for the gold standard. 
  Each column of points corresponds to one model (e.g. the right-most column is the eDN).
  The dashed diagonal line shows model log-likelihoods, which are linear in information.
  \textbf{(a)}: 
  Metrics (other than log-likelihoods) calculated on raw model output.
  The legend shows the Pearson correlation coefficient $r$ and Spearman rank order correlation $\rho$ between each metric and information gain explained.
  Many metrics are inconsistent with information, in particular showing rank order changes.
  \textbf{(b)}: 
  Included models are all optimised for nonlinearity, blur and centre bias.
  All metrics are highly correlated with information gain explained.
  Even though log-likelihoods should be preferred for the reasons outlined in our paper, results using other metrics are unlikely to change qualitatively so long as they are appropriately converted to probabalistic models.
  However, note also that since all metrics must converge to the gold-standard model at (1, 1), all metrics apart from log-likelihoods are nonlinear in information.
   }
  \label{fig:metric_comparison}
\end{figure*}

While many saliency metrics have been shown to provide inconsistent model rankings in the past \cite{Riche2013}, we now show how treating saliency maps in a stringent probabalistic fashion can resolve these discrepancies, resulting in a fully principled and unambiguous evaluation framework (see Figure \ref{fig:metric_comparison}).

In Figure \ref{fig:metric_comparison} we show a comparison between a number of metrics and information gain.
Crucially, we do this both for the raw model output and for models that have been appropriately converted to a probabalistic framework (as we do in the paper, by jointly optimising for nonlinearity, centre bias and blurring).
To allow comparison between the metrics, we have rescaled the performance metrics to yield zero for the baseline (centre bias) model and one for the gold standard.
This corresponds to the ratio of the explainable information gain (gold standard) to that explained by the models.
That is, for each metric, the baseline model is always at point (0, 0) and the gold standard at point (1, 1; not shown in the plot).

Figure \ref{fig:metric_comparison}a shows this space for all metrics computed on the raw model output (except for log-likelihoods, which must be optimised for nonlinearities to be meaningful).
Log-likelihood is a straight line running through the space, because its differences are exactly information gain.
The metrics strongly disagree, showing inconsistent rankings --- this is similar to what has been previously shown when comparing saliency metrics \cite{Riche2013}.
In addition, they are inconsistently correlated with information gain, from almost perfectly rank-order correlated (e.g. AUC with a uniform nonfixation distribution) to not at all (fixation-based KL-divergence with a centre bias nonfixation distribution).

In contrast, when the saliency maps are appropriately translated into probabalistic models (Figure \ref{fig:metric_comparison}b), all metrics become highly correlated with log-likelihood (and thus information gain) in both value and ranking.
This shows that model comparisons that have used other metrics would be unlikely to change qualitatively if log-likelihoods were compared instead --- provided that the models are properly converted into a probabalistic framework.



Additionally, this visualisation serves to demonstrate two desirable features of using log-likelihoods.
First, note that for four of the metrics the model ranks with information do change slightly, whereas the model ranking according to log-likelihood remains consistent with information.
Second, recall that all metrics must converge to the gold-standard model at (1, 1).
This means that all metrics aside from log-likelihoods are nonlinear in information, since they diverge from the line linking (0, 0) and (1, 1).
These metrics will answer the question of ``how close are we to understanding image-based saliency'' in a distorted way.

However, we can see that all of these metrics will nevertheless provide an answer of ``not very'' for the dataset here (for example, AUC would say ``40\%!''). 
To reach this conclusion, it is essential to use the correct baseline model.
If you are interested in image-based saliency, the appropriate baseline is the centre bias model, not the uniform distribution.

\subsection{Estimation considerations}
\label{sec:app_estimation_considerations}


One principle advantage of using log-likelhoods instead of image-based KL-divergence is that for all model comparisons but comparing against the gold standard we do not have to rely on the assumptions made for the gold standard but can simply use the unbiased sample mean estimator:
$$
\hat \EE[\log q_1(x)/q_2(x)] = \frac{1}{N} \sum_{k=1}^N \log q_1(x_k)/q_2(x_k)
$$
This is why we used the sample mean estimator for all model comparisons rather than the gold standard to estimate the ELLR. 

However, estimating the upper limit on information gain still requires a gold standard (an estimate of the true distribution $p(x)$). 
Image-based KL-divergence requires this not only for estimating the upper bound, but for calculating the performance of any model.
There, it has usually been done using a 2D histogram or Gaussian kernel density estimate (see Table \ref{tab:kl_divergence}), and
the hyper parameters (e.g. bin size, kernel size) have commonly been chosen based on fovea size or eye tracker precision.
In our framework of interpreting saliency maps as probability distributions, a principled way of choosing these hyper parameters is to cross-validate over them to get the best possible estimate of the true distribution.

For our dataset, the optimal cross-validated kernel size was $27$ pixels which is relatively close to the commonly used kernel size of $1^\circ$ (37 pixels).
However, with more fixations in the dataset the optimal cross-validated kernel sizes will shrink, because the local density can be estimated more precisely.
Therefore, choosing these hyperparameters on criteria other than cross-validation will produce inaccurate estimates of the ELLR in the large data limit.


Since we conclude that our understanding of image-based saliency is surprisingly limited, we have been using a conservative strategy 
for estimating the information gain of the gold standard that is downward biased such that we obtain a conservative upper bound on the fraction of how much we understand about image-based saliency.
To this end, we not only used the unbiased sample estimator for averaging over the true distribution but we resorted to a cross-validation strategy for estimating the gold standard that takes into account how well the distributions generalize across subjects:
$$
\hat E[p_{gold}] = \sum_{j=1}^M \frac{1}{N_j} \sum_{k=1}^{N_j} \log p_{gold}(x_{jk}|j)
$$
where the first sum runs over all subjects j and $p_{gold}(x_{jk}|j)$ denotes a kernel density estimator which uses all fixations but the one of subject $j$.
For comparison, if one would simply use the plain sample mean estimator for the gold standard the fraction explained would drop to an even smaller value of only 22\%.
Our approach guarantees that it is very likely that the true vale falls into the range between 22 and 34\%.

\subsection{Modelling temporal effects}
\label{sec:temporal_effects_methods}

We quantified the extent to which humans use inhibition of return when viewing static images by including these temporal effects into the model comparison.
If humans use inhibition of return then adding this mechanism to the model should lead to higher log-likelihoods than models without IoR.

The temporal effects were optimized independently of the other factors detailed above.
We modeled inhibition of return by multiplying the likelihood with a factor depending on $\Delta = \sqrt{(x_i-x_{i-1})^2+(y_i-y_{i-1})^2}$:

\begin{align*}
&p(x_i, y_i | x_1, y_1, \dots, x_{i-1}, y_{i-1}) \\
&\qquad  \propto (1+f(\Delta))p(x_i, y_i)
\end{align*}

If $f(\Delta)=0$ then the saliency map prediction is unchanged by temporal information.
If $f(\Delta)<0$ then saliency values at this distance ($\Delta$) to the previous fixation are decreased: they become less likely, reflecting inihibition of return.
Conversely, if $f(\Delta)>0$ this reflects an attraction or self-excitation effect.

Notice however that in order to make $(1+f(\Delta))p(x_i, y_i)$ a density once more, it must be renormalized to sum to $1$. 
This results in a shift of the threshold $\theta$ that separates excitation ($f(\Delta)>\theta$) from inhibition ($f(\Delta)<\theta$).

\begin{figure}[!ht]
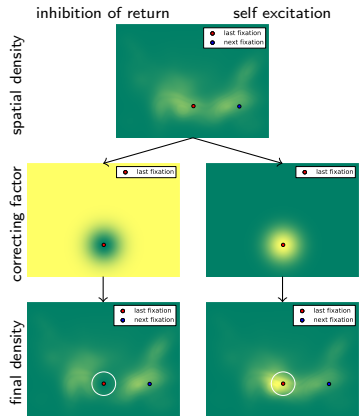

  \center
  \include{figures/temporal_effects.pgf}

  \caption{
	\small  
  Modelling temporal effects.
  In all images, yellow depicts higher density and green lower density.  
  We weight the spatial point process density (top image) by a correction factor centred over the previous fixation (red point) that either decreases (inhibition of return, left column) or increases local saliency (self-excitation, right column).
  The final density is then used to predict the next fixation (blue point).
  }
\end{figure}

We modeled $f$ using a Gaussian, i.e. $f(\Delta) = -\delta\exp(-\tfrac12 \Delta^2/\sigma^2)$. 
Inspired by results of \cite{Tatler2008, Tatler2009} we extended the temporal effect to include eccentricity effects (i.e. allow the region of temporal inhibition / excitation to be elliptical) by setting $\Delta = \sqrt{(x_i-x_{i-1})^2+\alpha (y_i-y_{i-1})^2}$.
We also tested a difference of Gaussians model with separable inhibitory and self-reinforcing components, but omit these results here since they produced the same results as the simple Gaussian (i.e. no evidence for an inhibitory component).

To optimise these temporal parameters the fully optimized spatial density (nonlinearity, centre bias and blur) was used as a base. 
All temporal parameters where again optimized jointly with L-BFGS SLSQP, but independent of the already optimized spatial parameters.


%
%

\section*{Acknowledgements}

MK, TSAW and MB conceived of the experiments.
MK analysed the data.
MK, TSAW and MB wrote the paper.
We thank Lucas Theis for his suggestions, and Eleonora Vig for helpful comments on an earlier draft of this manuscript.
TSAW was supported by a Humboldt Postdoctoral Fellowship from the Alexander von Humboldt Foundation.

\bibliography{library.bib,own_literature.bib}

\end{document}

%% file: figures/data.tex
\newcommand{\MutualInformationImageSpatial}{2.1}
\newcommand{\BaselineToBestPercentOfGold}{19}
\newcommand{\BestModelPerformance}{1.29}
\newcommand{\BestModelPercentOfGoldRelBaseline}{34}
\newcommand{\BestModelCenterbiasToBlurPercent}{0}
\newcommand{\BaselineSelfExcitationPerformance}{1.30}
\newcommand{\BestModel}{eDN}
\newcommand{\PossibleInformationGain}{1.20}
\newcommand{\BestModelNonlinearityToCenterbiasPercent}{4}
\newcommand{\BaselinePerformance}{0.89}
\newcommand{\BestModelPercentOfGold}{62}
\newcommand{\BaselinePercentOfGold}{42}
\newcommand{\BestModelNonlinearityPercent}{96}

%% file: figures/data_kienzle.tex
\newcommand{\KienzleBestModel}{GBVS}

\newcommand{\KienzleBestModelPercentOfGoldRelBaseline}{22}

\newcommand{\KienzleBaselinePercentOfGold}{34}

\newcommand{\KienzlePossibleInformationGain}{0.92}

%% file: figures/intro_figure.pgf.tex
\ifdefined\noktikz
\else
  \usetikzlibrary{calc}
  \usetikzlibrary{shapes}
  \graphicspath{{/kyb/agmb/mkuemmerer/Documents/Uni/Bethge/Saliency/TPAMI/figures/}}

\fi

\ifdefined\abstikzunit
\else
    \newlength\abstikzunit
    \newlength\tikzunit
\fi

\setlength\abstikzunit{1cm}
\newcommand\scalingfactor{0.6}
\newcommand\imgwidth{5.0}
\newcommand\fullimgwidth{11}
\newcommand\labelalpha{0.4}
\newcommand\labelheight{1.5}
\setlength\tikzunit{\scalingfactor\abstikzunit}

\newcommand\loglik[1]{\textsf{$\mathsf{#1}$\,bit/fix}}
\newcommand\auc[1]{\textsf{$\mathsf{#1}$\,\%}}

\begin{tikzpicture}[scale=\scalingfactor]

  \coordinate (col2) at (5.9, 0);
  \coordinate (row2) at (0, -4.2);
  \coordinate (label_rel_subfig) at (-0.6, 0);


  \node[anchor=north west] (fix_high)  at (0,0)
    {\includegraphics[width=\imgwidth\tikzunit]{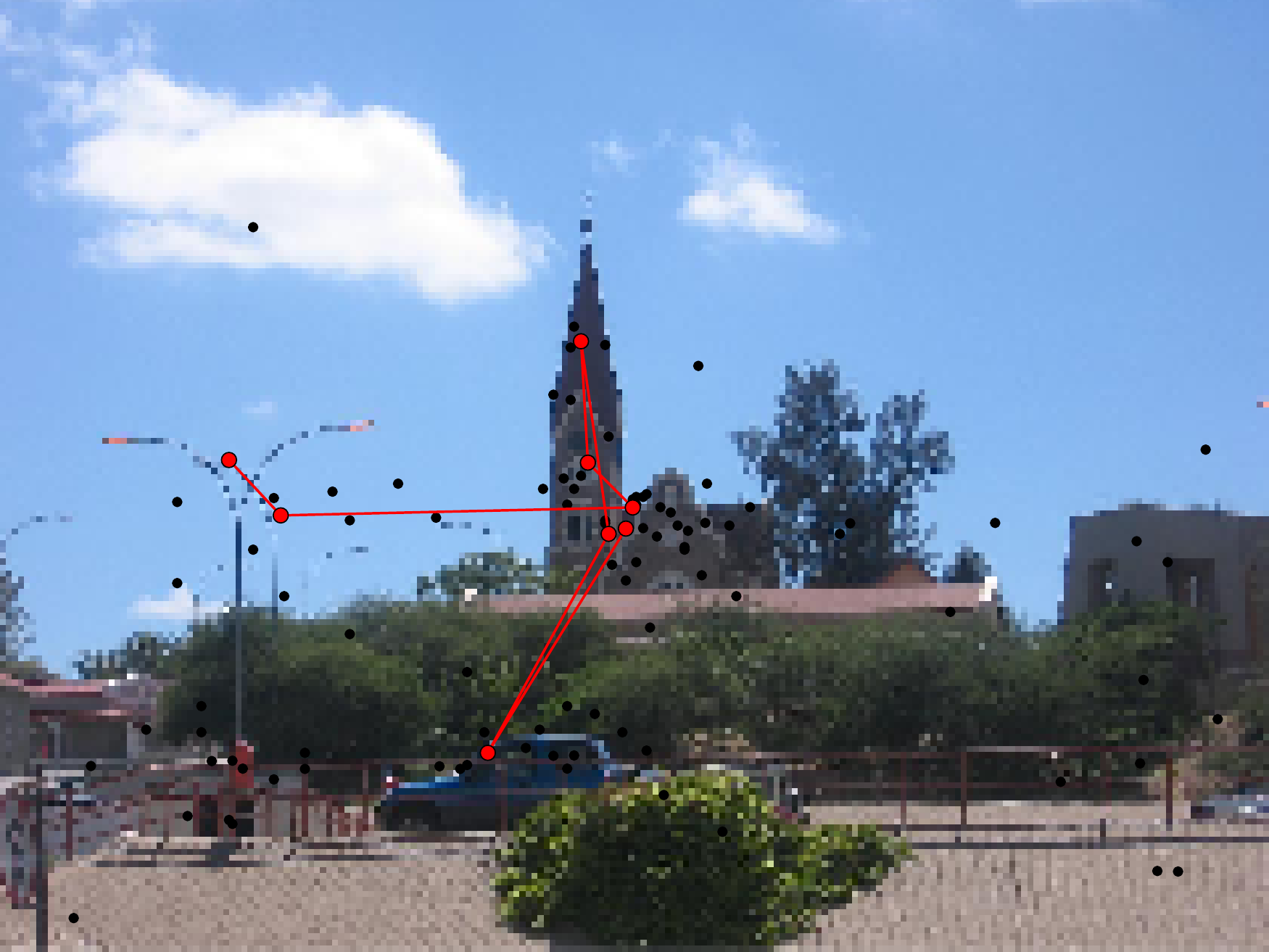}};

  \node[anchor=north west] at (label_rel_subfig) {\textsf{a)}};

  \node[anchor=north west] (fix_high)  at (col2)
    {\includegraphics[width=\imgwidth\tikzunit]{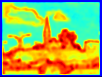}};

  \node[anchor=north west] at ($ (col2) + (label_rel_subfig) $) {\textsf{b)}};

  \node[anchor=north west] (fix_high)  at (row2)
    {\includegraphics[width=\fullimgwidth\tikzunit]{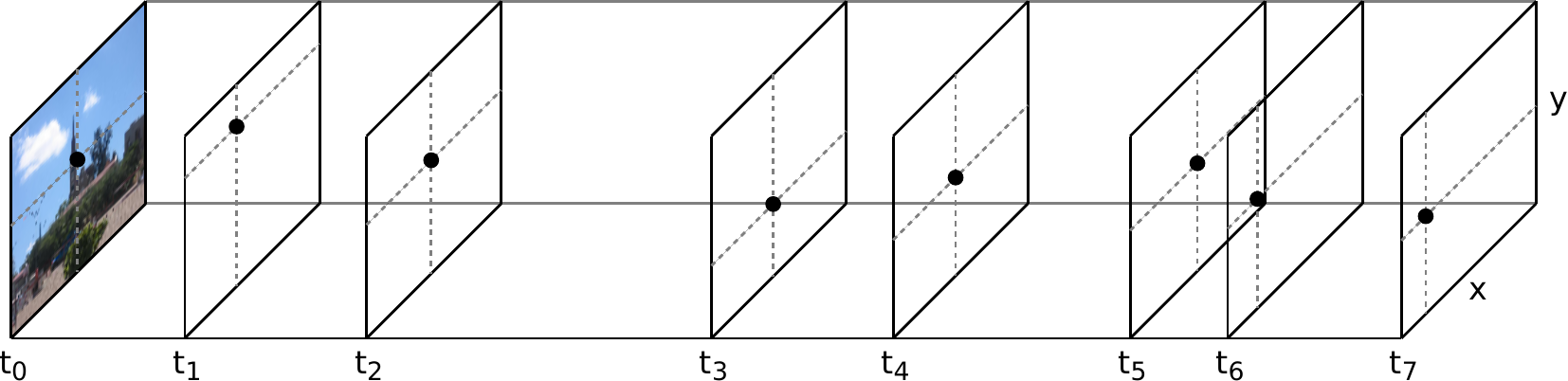}};

  \node[anchor=north west] at ($ (row2) + (label_rel_subfig) $) {\textsf{c)}};

\end{tikzpicture}

\let\abstikzunit\undef
\let\scalingfactor\undef
\let\imgwidth\undef
\let\tikzunit\undef

%% file: figures/modeling_stationary.pgf.tex
\ifdefined\noktikz
  \input{figures/model_colors}
\else
  \usetikzlibrary{calc}
  \usetikzlibrary{shapes}
  \graphicspath{{/kyb/agmb/mkuemmerer/Documents/Uni/Bethge/Saliency/TPAMI/figures/}}

  \input{/kyb/agmb/mkuemmerer/Documents/Uni/Bethge/Saliency/TPAMI/figures/model_colors}
\fi

\newcommand\loglik[1]{\textsf{$\mathsf{#1}$\,bit/fix}}
\newcommand\auc[1]{\textsf{$\mathsf{#1}$\,\%}}


\ifdefined\abstikzunit
	\setlength\abstikzunit{1cm}
	\newcommand\scalingfactor{0.8}
	\newcommand\imgwidth{5}
	\setlength\tikzunit{\scalingfactor\abstikzunit}
\else
	\newlength\abstikzunit
	\setlength\abstikzunit{1cm}
	\newlength\tikzunit
	\newcommand\scalingfactor{0.8}
	\newcommand\imgwidth{5}
	\setlength\tikzunit{\scalingfactor\abstikzunit}
\fi

\begin{tikzpicture}[scale=\scalingfactor,font=\sffamily]
  \definecolor{fillcolor}{rgb}{0.9, 0.9, 0.9};
  \tikzset{boxstyle/.style={fill=fillcolor, rounded corners=2mm}};


  \coordinate (leftcolwidth) at (6.7,0);
  \coordinate (leftcolheight) at (0,-2);
  \newcommand{\leftcolimgheight}{1.2}

  \coordinate (leftcolsize) at ($ (leftcolwidth) + (leftcolheight) $);

  \coordinate (leftcolsep) at (0, -1);
  \coordinate (leftcolop) at ($ (leftcolheight) + 0.5*(leftcolsep) + 0.5*(leftcolwidth) $);

  \coordinate (lefttotalcolheight) at ($ (leftcolheight) + (leftcolsep) $);

  \coordinate (leftcolimage) at (0.3,-0.4);
  \coordinate (leftcoltext) at ($ 0.5*(leftcolheight) + (2,0) $);
  \coordinate (leftcolimagestack) at (+0.15,-0.15);

  \coordinate (leftcoltemporalsep) at (0,-0.2);

  \coordinate (col2) at (5.9, 0);
  \coordinate (row2) at (0, -4.2);
  \coordinate (label_rel_subfig) at (-0.6, 0);


  \coordinate (rightcol) at ($ (leftcolwidth) + (1,0) $);
  \newcommand\rightcolwidth{9}

  \coordinate (rightcolheight) at (0,-6);

  \node[anchor=north west] at ($ (label_rel_subfig) + (-0.2, 0) $) {a)};


  \path[draw, boxstyle] (0,0) rectangle  ($ (leftcolwidth) + (leftcolheight) $);
  \node[anchor=north west, inner sep=0] at (leftcolimage)
    {\includegraphics[height=\leftcolimgheight\tikzunit]{saliencymap.png}};
  \node[anchor=west] at (leftcoltext) {Saliency map};

  \draw[->,thick] ($ (leftcolheight) + 0.5*(leftcolwidth) + (0, -0.15) $) -- +($ (leftcolsep) + (0, 0.3) $);


  \path[draw, boxstyle,fill=colornonlinearity] (lefttotalcolheight) rectangle  ($  (lefttotalcolheight) + (leftcolsize) $);
  \node[anchor=north west, inner sep=0] at ($ (lefttotalcolheight) + (leftcolimage) $)
    {\includegraphics[height=\leftcolimgheight\tikzunit]{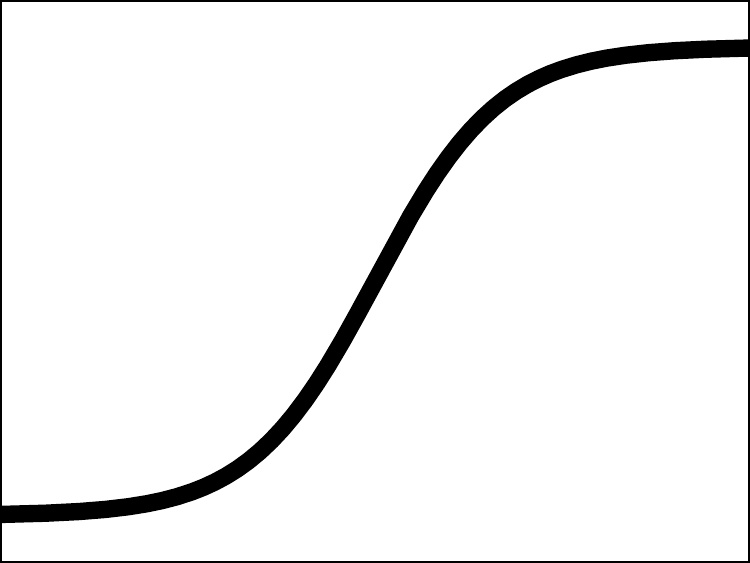}};
  \node[anchor=west] at ($ (lefttotalcolheight) + (leftcoltext) $) {Nonlinearity};

  \draw[->,thick] ($ 1*(lefttotalcolheight) + (leftcolheight) + 0.5*(leftcolwidth) + (0, -0.15) $) -- +($ (leftcolsep) + (0, 0.3) $);


  \path[draw, boxstyle, fill=colorcenterbias]
    ($ 2*(lefttotalcolheight) $) rectangle ($  2*(lefttotalcolheight) + (leftcolsize) $);
  \coordinate (this_symbol_pos) at ($ 2*(lefttotalcolheight) + (leftcolimage) $);
  \node[anchor=north west, inner sep=0] at ($ (this_symbol_pos) -0.5*(leftcolimagestack) $)
    {\includegraphics[height=\leftcolimgheight\tikzunit]{nonlinearity_symbol.pdf}};
  \node[anchor=north west, inner sep=0] at ($ (this_symbol_pos) +0.5*(leftcolimagestack) $)
    {\includegraphics[height=\leftcolimgheight\tikzunit]{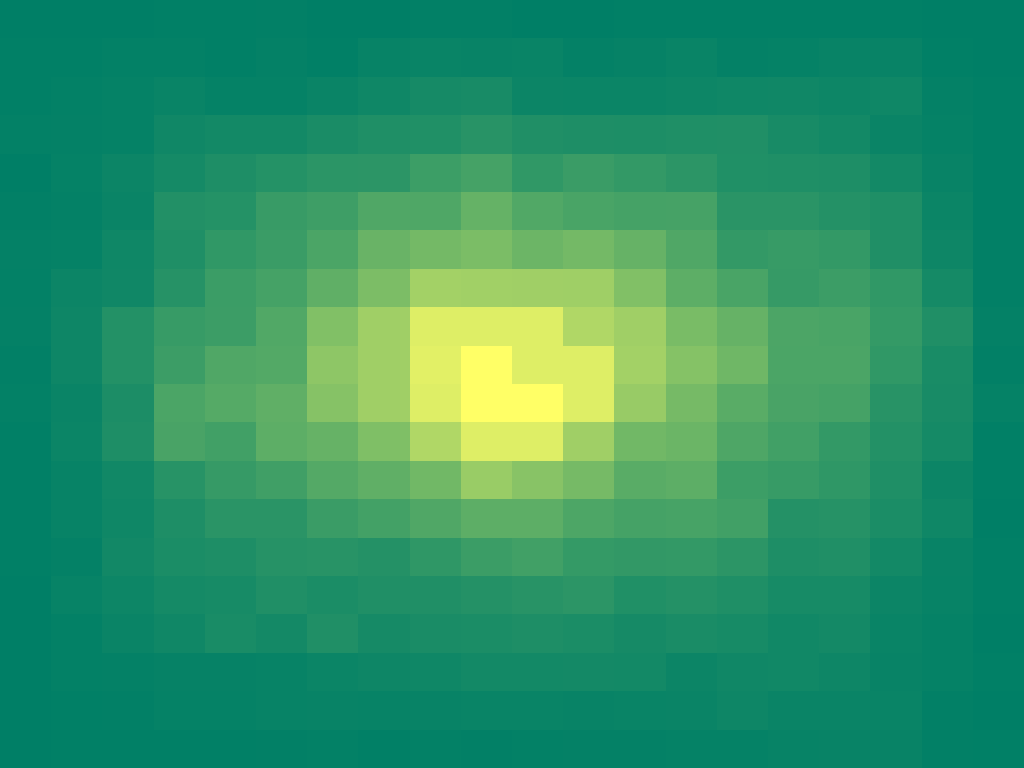}};
  \node[anchor=west,align=left] at ($ 2*(lefttotalcolheight) + (leftcoltext) $) {Nonlinearity\\Centre bias};

  \draw[->,thick] ($ 2*(lefttotalcolheight) + (leftcolheight) + 0.5*(leftcolwidth) + (0, -0.15) $) -- +($ (leftcolsep) + (0, 0.3) $);


  \path[draw, boxstyle, fill=colorblur]
    ($ 3*(lefttotalcolheight) $) rectangle ($  3*(lefttotalcolheight) + (leftcolsize) $);
  \coordinate (this_symbol_pos) at ($ 3*(lefttotalcolheight) + (leftcolimage) $);
  \node[anchor=north west, inner sep=0] at ($ (this_symbol_pos) -1*(leftcolimagestack) $)
    {\includegraphics[height=\leftcolimgheight\tikzunit]{nonlinearity_symbol.pdf}};
  \node[anchor=north west, inner sep=0] at ($ (this_symbol_pos) +0.0*(leftcolimagestack) $)
    {\includegraphics[height=\leftcolimgheight\tikzunit]{saliencymap_nonparametric.png}};
  \node[anchor=north west, inner sep=0] at ($ (this_symbol_pos) +1.0*(leftcolimagestack) $)
    {\includegraphics[height=\leftcolimgheight\tikzunit]{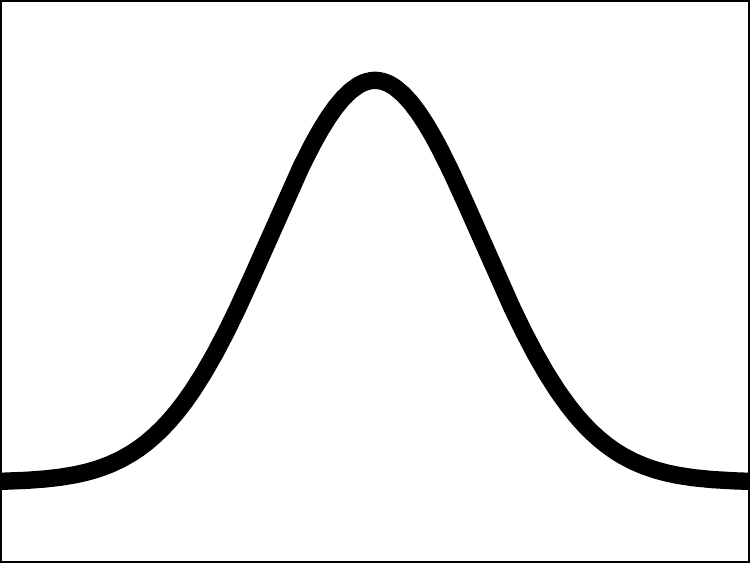}};
  \node[anchor=west,align=left] at ($ 3*(lefttotalcolheight) + (leftcoltext) $) {Nonlinearity\\Centre bias\\Blurring};


  \node[anchor=north west] at ($ (rightcol) + (label_rel_subfig) $) {b)};

  \node[anchor=north west] at (rightcol) {\includegraphics[width=\rightcolwidth\tikzunit]
    {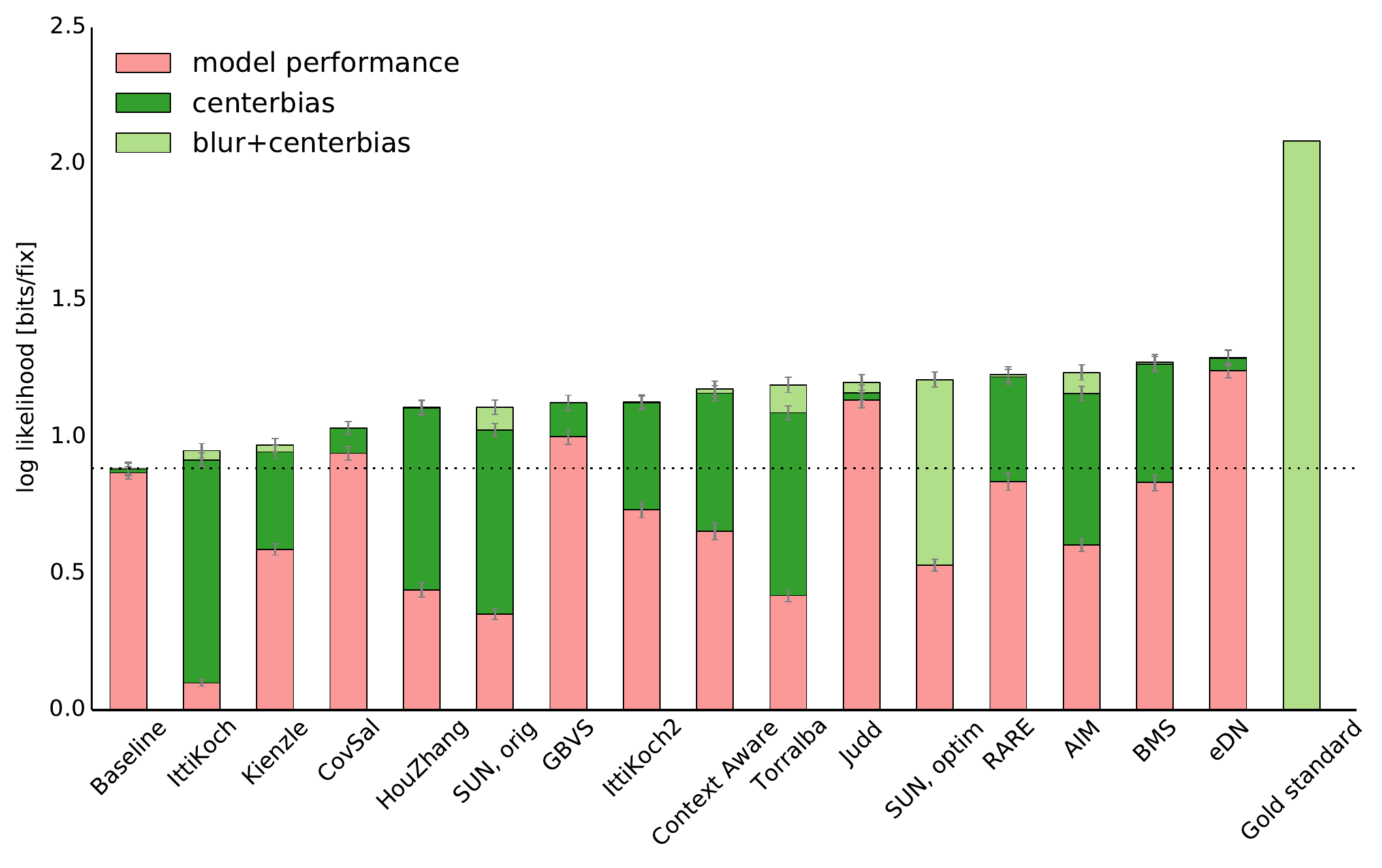}};

  \node[anchor=north west] at ($ (rightcol) + (rightcolheight) + (label_rel_subfig) $) {c)};

  \node[anchor=north west] at ($ (rightcol) + (rightcolheight) $)
   {\includegraphics[width=\rightcolwidth\tikzunit]%
     {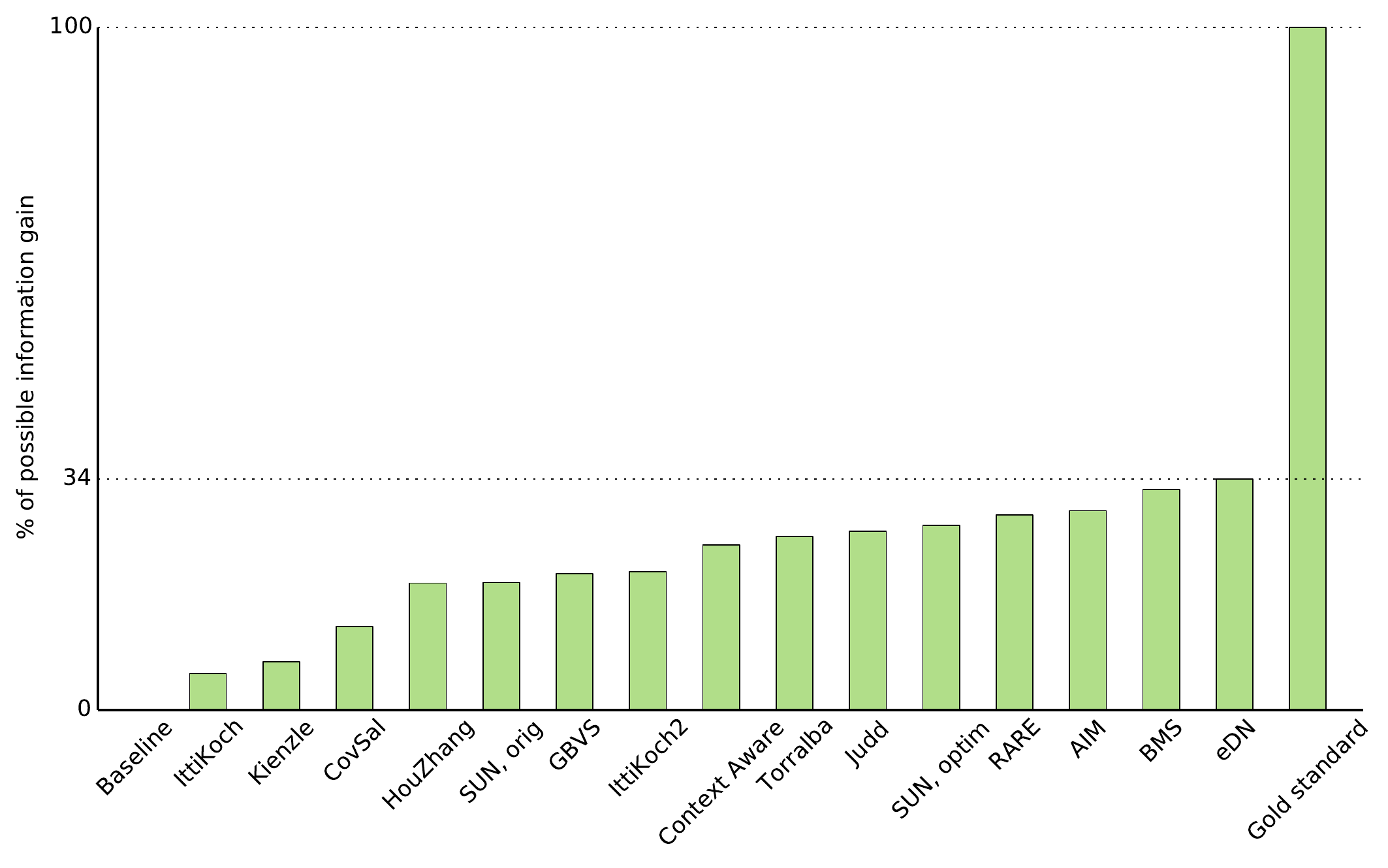}};


\end{tikzpicture}

\let\abstikzunit\undef
\let\scalingfactor\undef
\let\imgwidth\undef

%% file: figures/information_gains_schema.pgf.tex
\ifdefined\noktikz
  \input{figures/informationgainsdata}
  \input{figures/ig_colors}
\else
  \usetikzlibrary{calc}
  \usetikzlibrary{shapes}
  \graphicspath{{/kyb/agmb/mkuemmerer/Documents/Uni/Bethge/Saliency/TPAMI/figures/}}

  \input{/kyb/agmb/mkuemmerer/Documents/Uni/Bethge/Saliency/TPAMI/figures/informationgainsdata}
  \input{/kyb/agmb/mkuemmerer/Documents/Uni/Bethge/Saliency/TPAMI/figures/ig_colors}
\fi

\newcommand\loglik[1]{\textsf{$\mathsf{#1}$\,bits/fix}}
\newcommand\auc[1]{\textsf{$\mathsf{#1}$\,\%}}


\ifdefined\abstikzunit
\else
	\newlength\abstikzunit
	\newlength\tikzunit
\fi

\setlength\abstikzunit{1cm}
\newcommand\scalingfactor{0.5}
\newcommand\imgwidth{3.5}
\newcommand\labelalpha{0.4}
\newcommand\labelheight{1.5}
\newcommand\symbolwidth{1}
\setlength\tikzunit{\scalingfactor\abstikzunit}

\setlength\abstikzunit{1cm}
\setlength\tikzunit{\scalingfactor\abstikzunit}



\begin{tikzpicture}[scale=\scalingfactor,font=\tiny\sffamily,every node/.style={inner sep=0,outer sep=0}]
  \tikzset{labelnode/.style={anchor=south, inner ysep=1,align=center,font=\tiny\sffamily,
    minimum height=\labelheight\tikzunit,text width=\imgwidth\tikzunit,
    fill opacity=\labelalpha,text opacity=1.0}}

  \coordinate (colwidth) at (4.0, 0);
  \coordinate (row2) at (0, -4);

  \coordinate (col2) at (5.5, 0);
  \coordinate (col3) at ($ (col2) + 1.3*(colwidth) $);
  \coordinate (col4) at ($ (col3) + 1.4*(colwidth) $);
  \coordinate (col5) at ($ (col4) + 1.25*(colwidth) $);

  \coordinate (row_center) at (0, 0);

  \coordinate (centerbias) at ($ (col2) + 0.5*(colwidth) + (0,4.5) $);
  \coordinate (centerbias_div) at ($ (col2) + 0.7*(colwidth)  + (row_center) $);

  \coordinate (nonlinearity) at ($ (col3) + 0.65*(colwidth)  + (row_center) $);
  \coordinate (gold_mul) at ($ (col3) + 0.85*(colwidth)  + (row_center) $);

  \coordinate (gold_mul_central) at ($ (gold_mul) + 0.5*(row2) $);

  \coordinate (ig_diff) at ($ (col4) + 0.6*(colwidth) $);

  \coordinate (colwidth) at (4.0, 0);
  \coordinate (colheight) at (0, -3);

  \coordinate (label_rel_fig) at (-4.5,3.5);
  \coordinate (ll_rel) at (2.5,-2);

  \coordinate (imagesep) at (0.0, -0.5);

  \coordinate (input_image) at ($ 0.5*(row2) $);


  \node (input_image_node) at (input_image)
      {\includegraphics[width=\imgwidth\tikzunit]{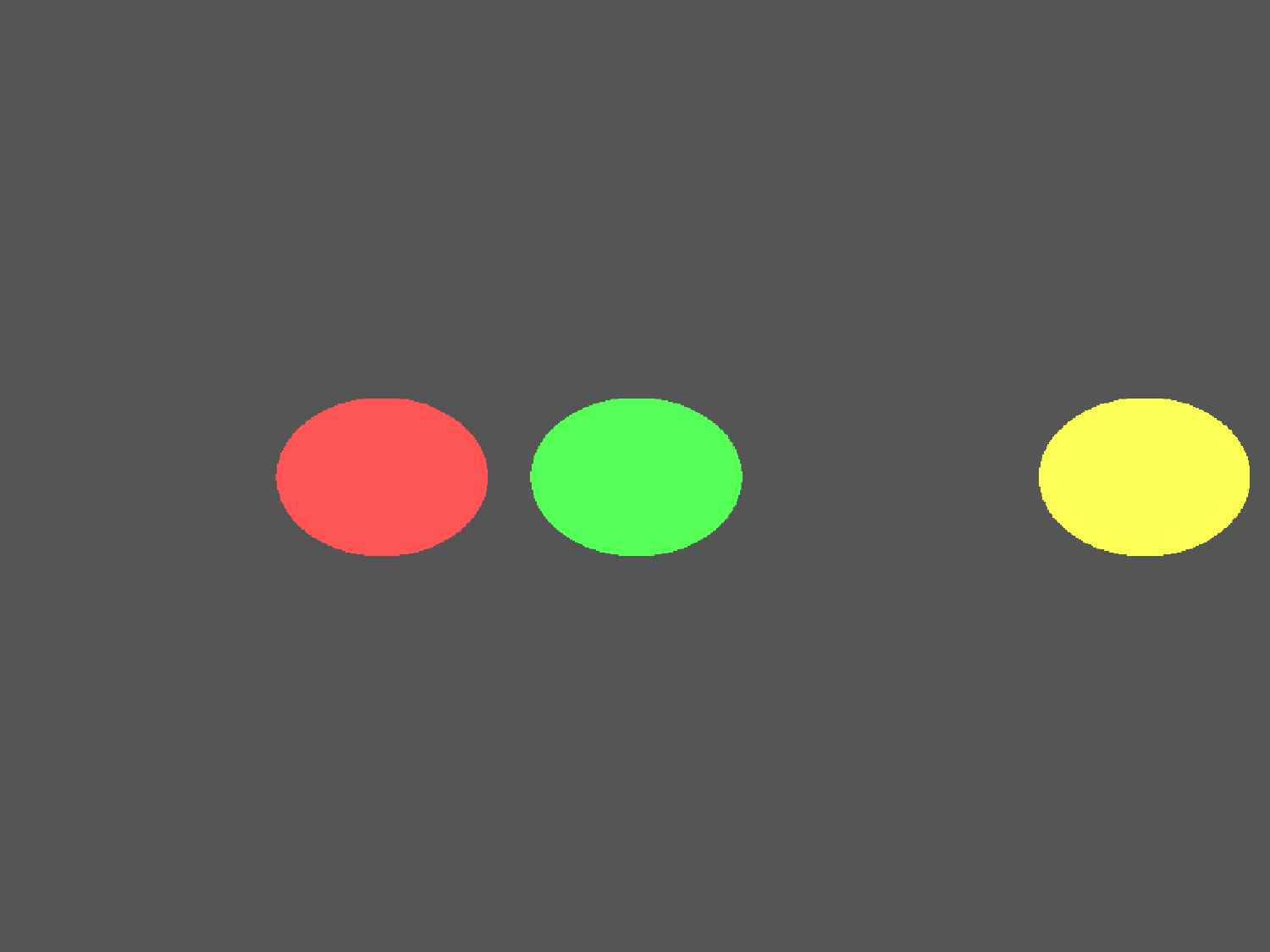}};


  \node (model_density_node) at ($ (col2) $)
      {\includegraphics[width=\imgwidth\tikzunit]{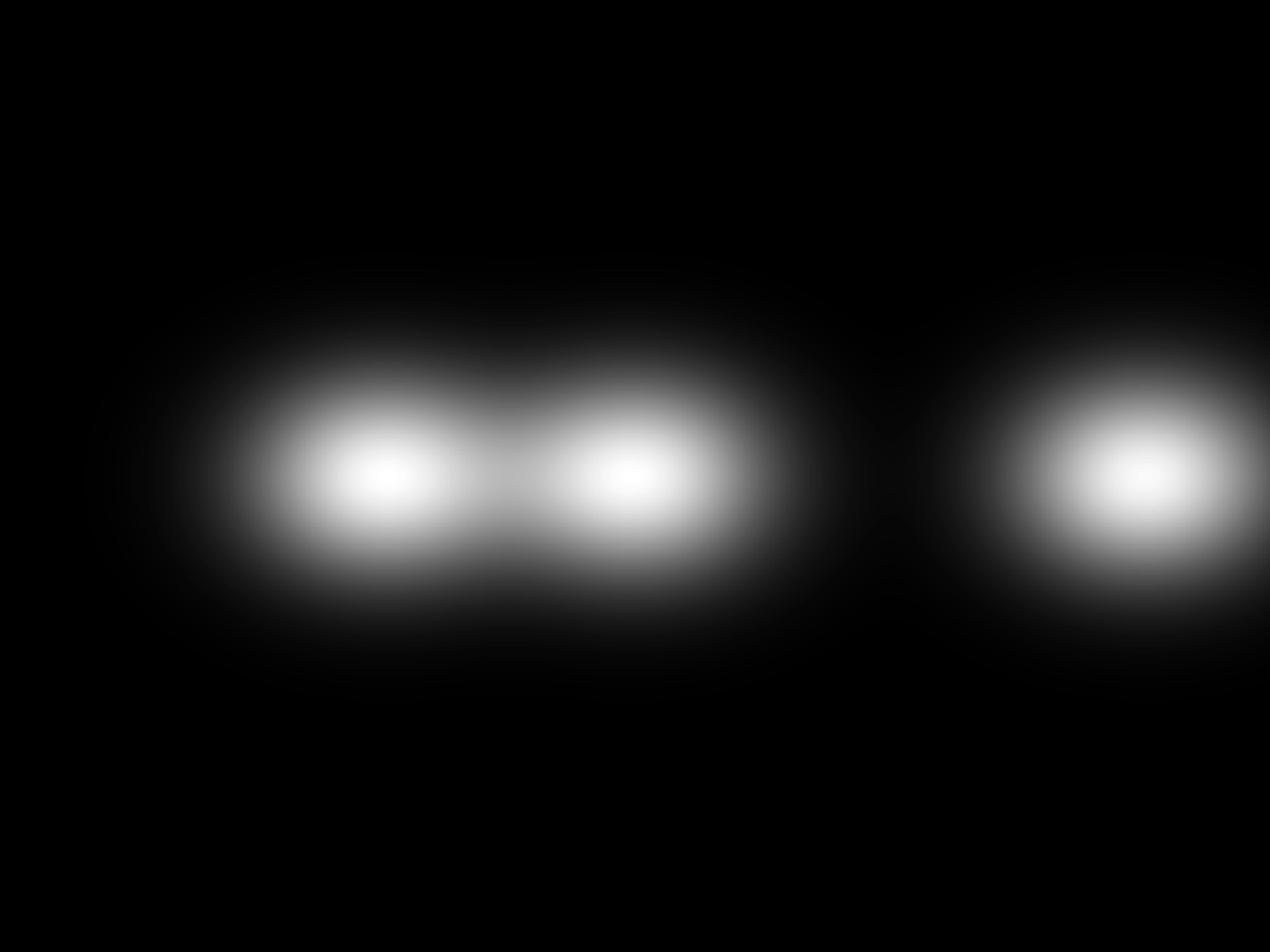}};

  \node (gold_density_node) at ($ (col2) + (row2) $)
      {\includegraphics[width=\imgwidth\tikzunit]{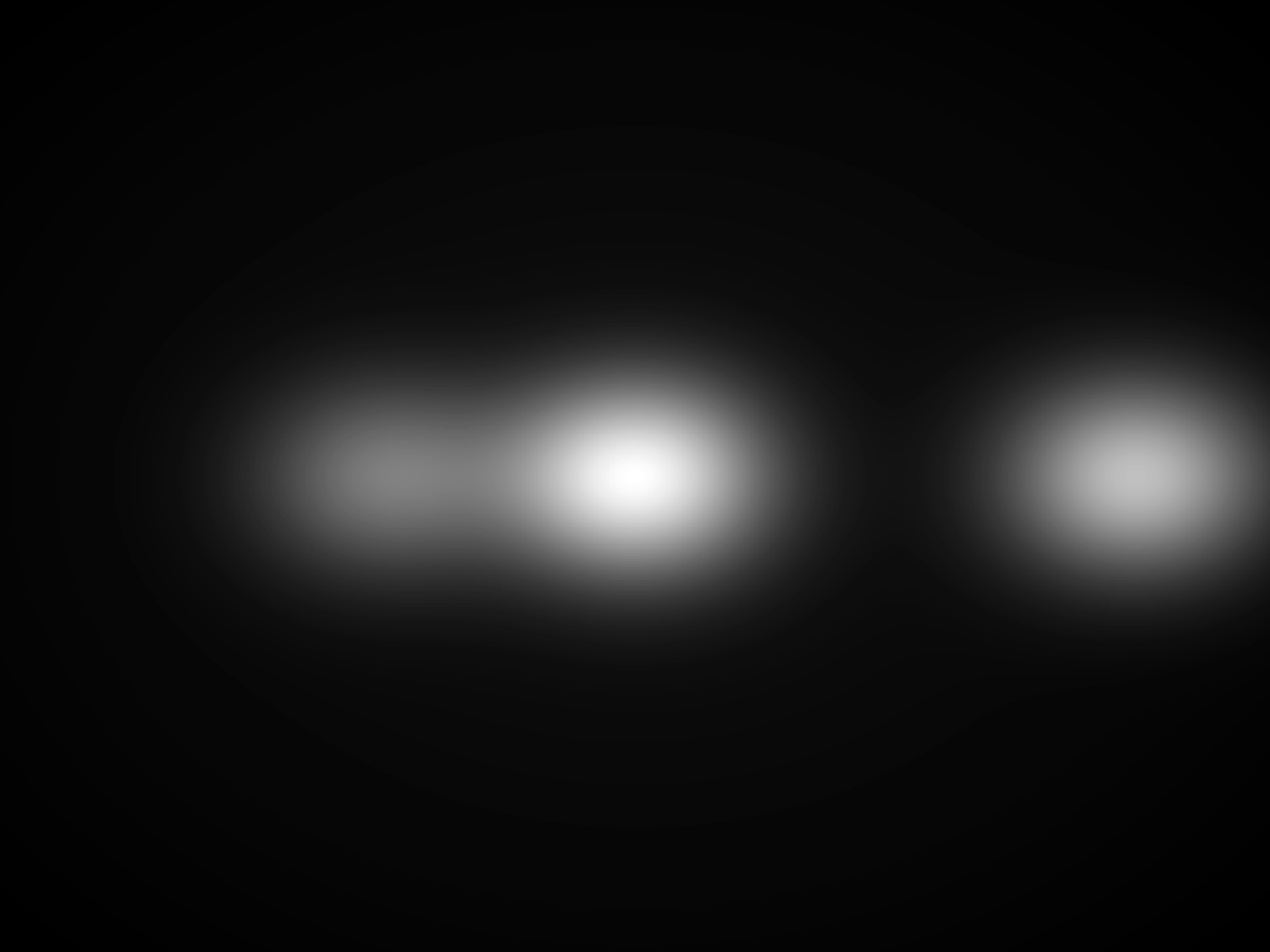}};

  \draw[->] (input_image_node) -- (model_density_node) node[pos=0.5,label={[label distance=5pt]above:model}] {};
  \draw[->] (input_image_node) -- (gold_density_node) node[pos=0.5,label={[label distance=5pt]below:true}] {};


  \node (centerbias_node) at ($ (centerbias) $)
      {\includegraphics[width=\imgwidth\tikzunit]{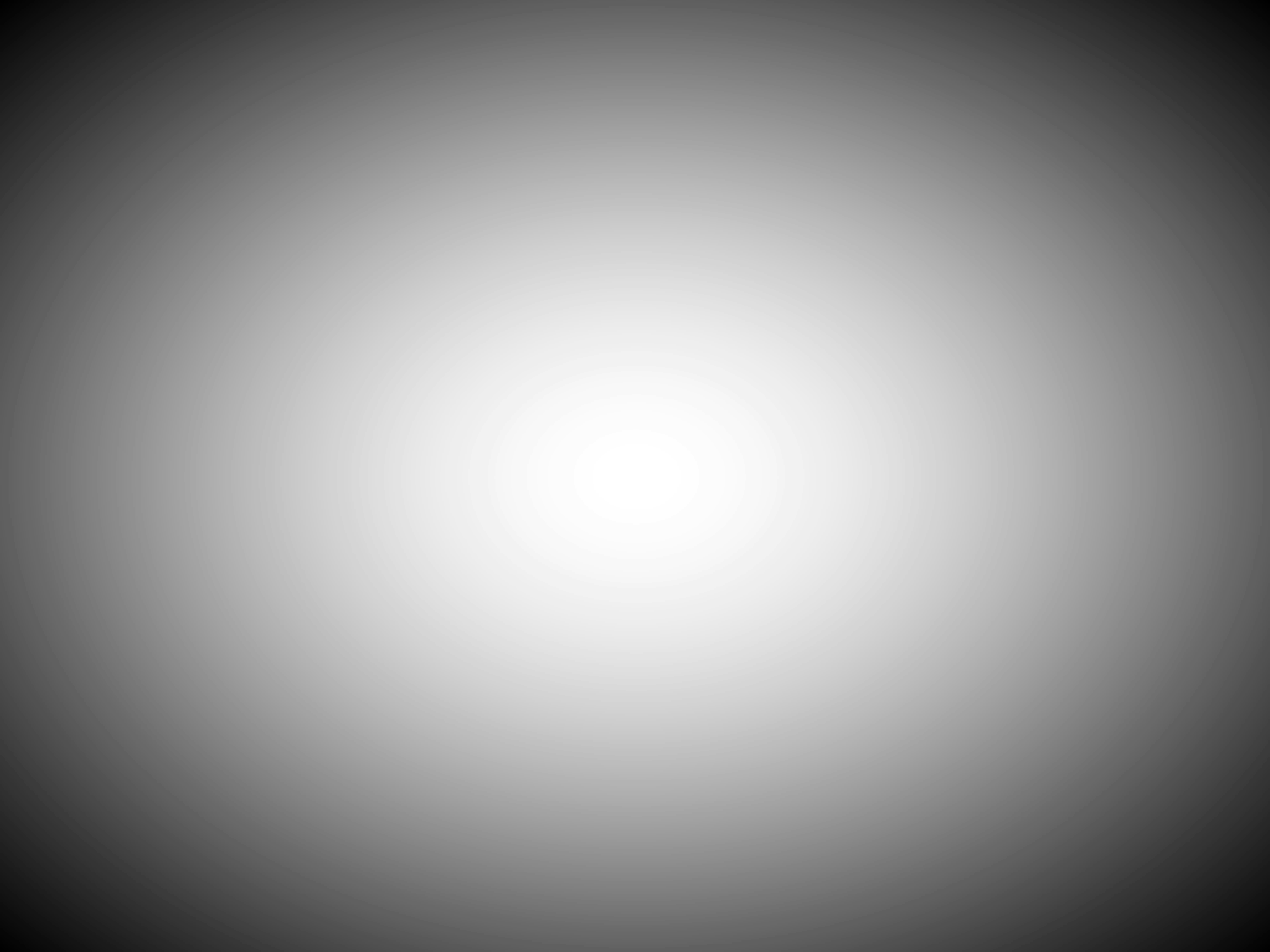}};

  \node[circle,draw] (centerbias_model_div_node) at ($ (centerbias_div) $) {$\div$};
  \node[circle,draw] (centerbias_gold_div_node) at ($ (centerbias_div) + (row2) $) {$\div$};

  \draw[->] (centerbias_node) -- (centerbias_model_div_node);
  \draw[->] (centerbias_node) -- (centerbias_gold_div_node);

  \draw (model_density_node) -- (centerbias_model_div_node);
  \draw (gold_density_node) -- (centerbias_gold_div_node);


  \node (model_saliency_node) at ($ (col3) $)
      {\includegraphics[width=\imgwidth\tikzunit]{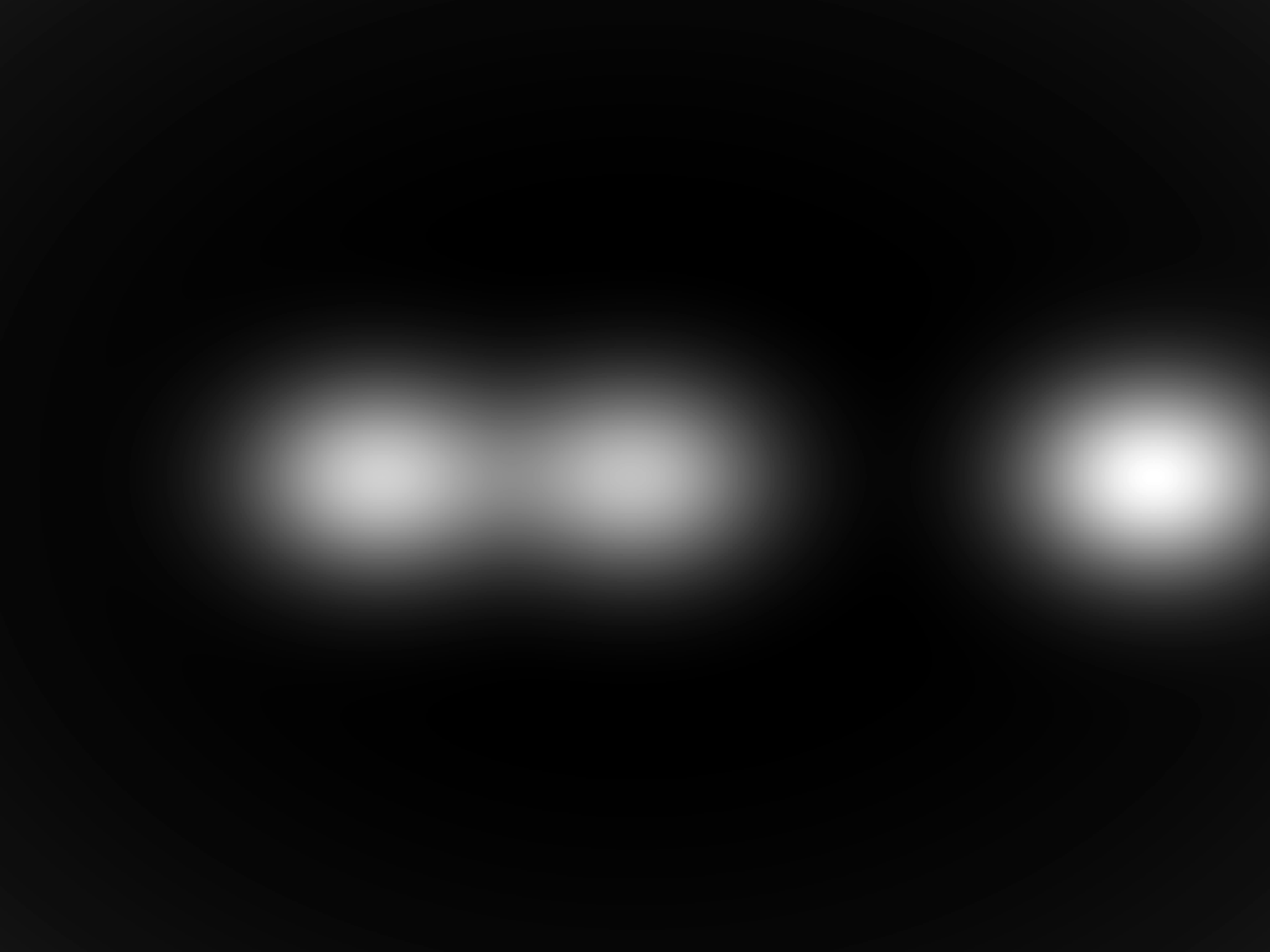}};

  \node (gold_saliency_node) at ($ (col3) + (row2) $)
      {\includegraphics[width=\imgwidth\tikzunit]{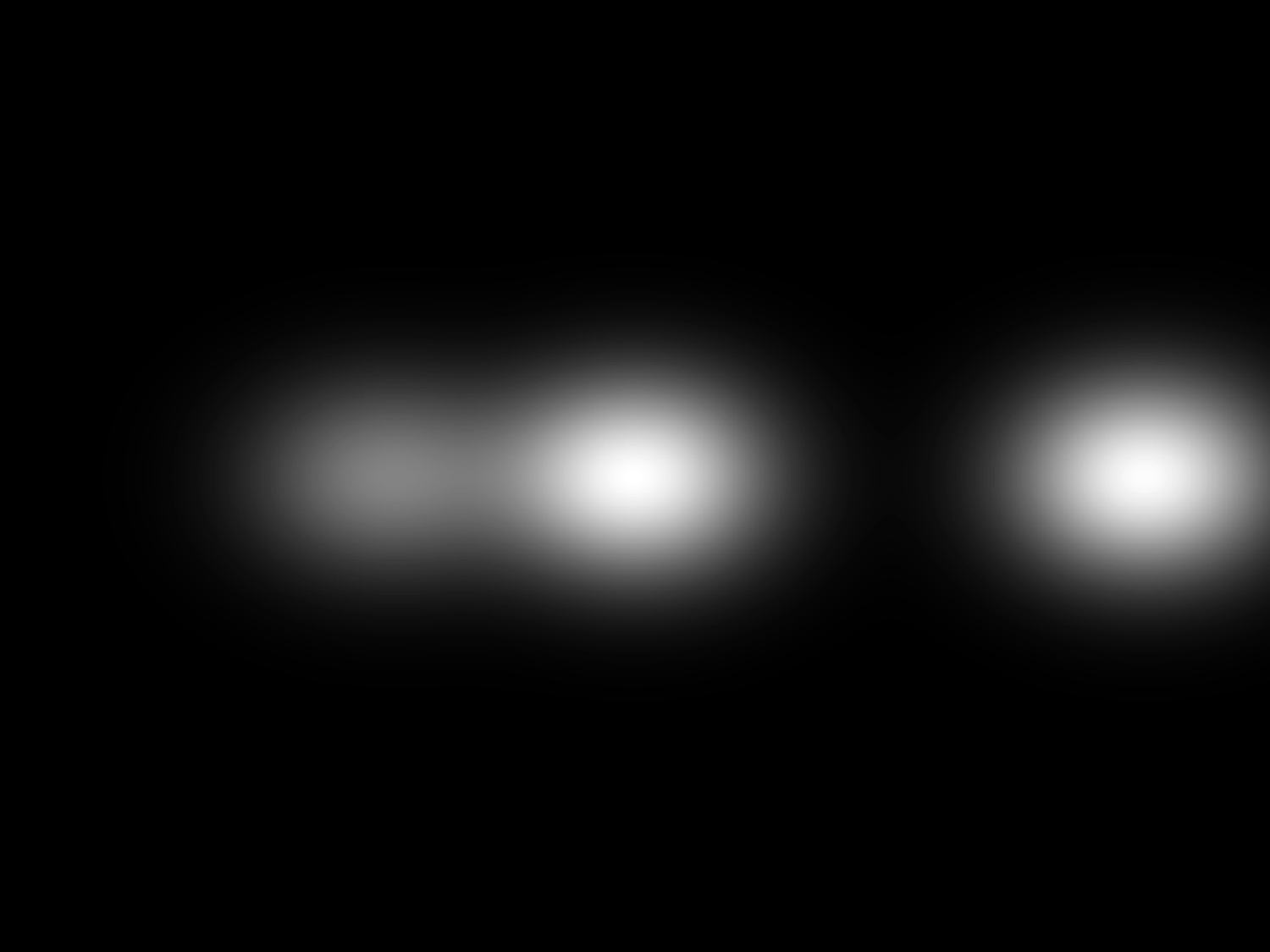}};

  \draw[->] (centerbias_model_div_node) -- (model_saliency_node);
  \draw[->] (centerbias_gold_div_node) -- (gold_saliency_node);


  \node[rectangle,draw,inner sep=2pt] (model_nonlinearity_node) at (nonlinearity)
    {$\log$};
  \node[rectangle,draw,inner sep=2pt] (gold_nonlinearity_node) at ($ (nonlinearity) + (row2) $)
    {$\log$};

  \draw[->] (model_saliency_node) -- (model_nonlinearity_node);
  \draw[->] (gold_saliency_node) -- (gold_nonlinearity_node);


  \node[circle,draw] (gold_mul_model_node) at ($ (gold_mul) $) {$\ \cdot\ $};
  \node[circle,draw] (gold_mul_gold_node) at ($ (gold_mul) + (row2) $) {$\ \cdot\ $};

  \draw (gold_density_node) |- (gold_mul_central);
  \draw[->] (gold_mul_central) -- (gold_mul_model_node);
  \draw[->] (gold_mul_central) -- (gold_mul_gold_node);

  \draw (model_nonlinearity_node) -- (gold_mul_model_node);
  \draw (gold_nonlinearity_node) -- (gold_mul_gold_node);


  \node (model_ig_node) at ($ (col4) $)
      {\includegraphics[width=\imgwidth\tikzunit]{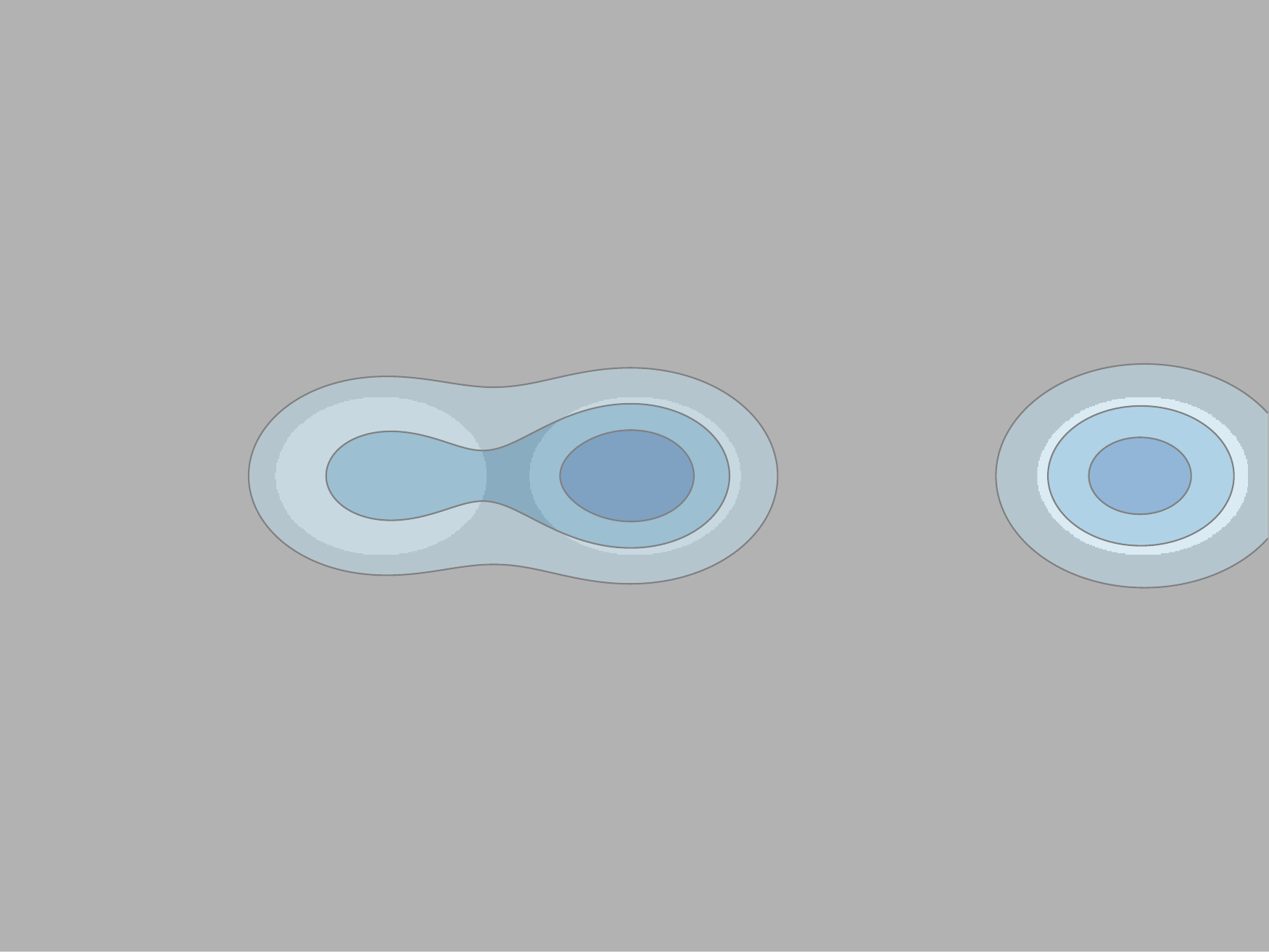}};
  \node (gold_ig_node) at ($ (col4) + (row2) $)
      {\includegraphics[width=\imgwidth\tikzunit]{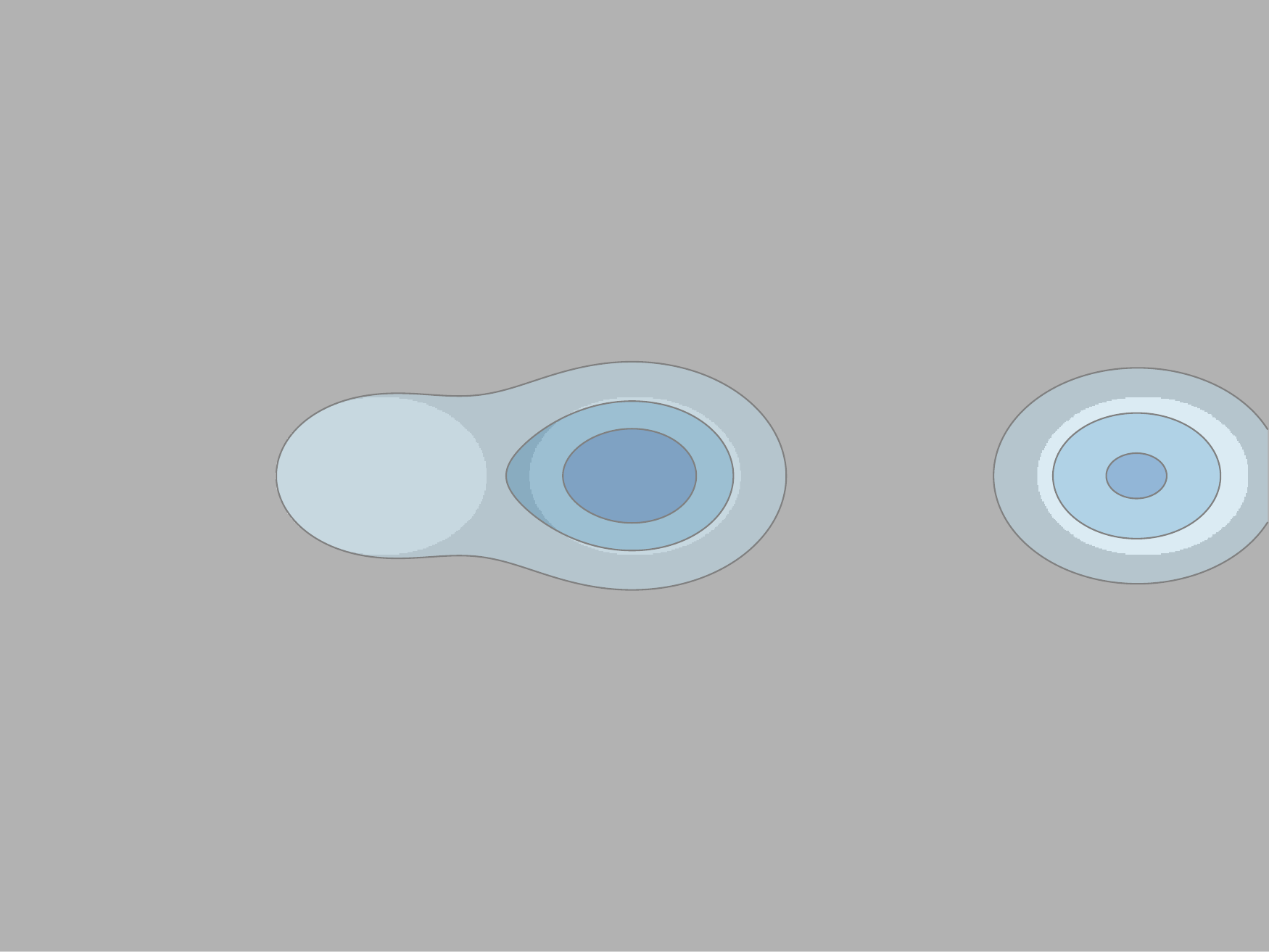}};

  \draw[->] (gold_mul_model_node) -- (model_ig_node);
  \draw[->] (gold_mul_gold_node) -- (gold_ig_node);


  \node[circle,draw] (model_diff_node) at ($ (ig_diff) $) {$-$};

  \draw (model_ig_node) -- (model_diff_node);
  \draw (gold_ig_node) -| (model_diff_node);

  \node (model_igdiff_node) at ($ (col5) $)
      {\includegraphics[width=\imgwidth\tikzunit]{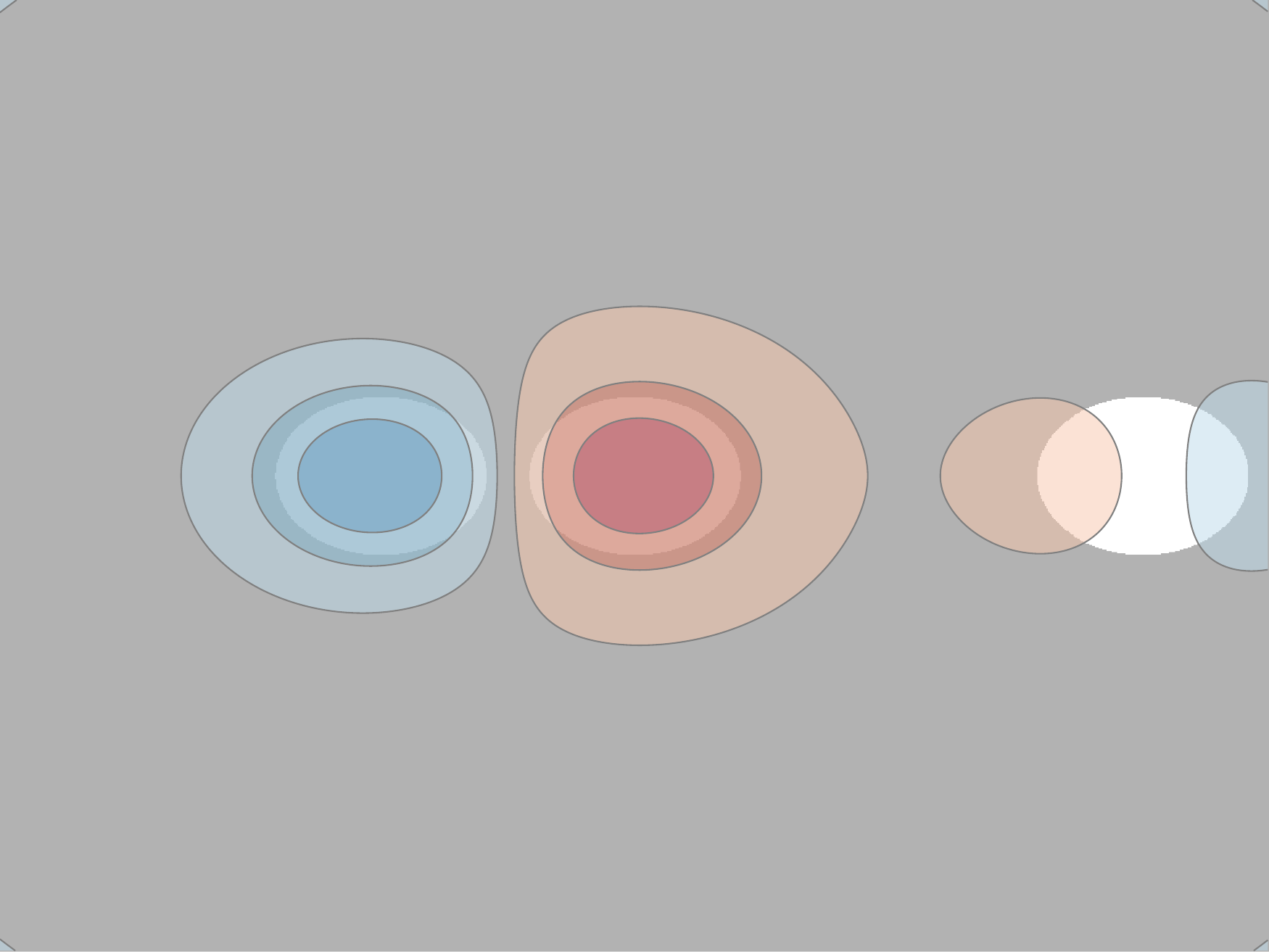}};

  \draw[->] (model_diff_node) -- (model_igdiff_node);


  \node[anchor=south] at (input_image_node.north) {image};
  \node[labelnode,fill=colordensity] at (model_density_node.north) {model density};
  \node[anchor=south] at (centerbias_node.north) {prior};
  \node[labelnode,fill=colorsaliency] at (model_saliency_node.north) {model\\ image-based\\ saliency};
  \node[labelnode,fill=colorig] at (model_ig_node.north) {model\\information gain};
  \node[labelnode,fill=colorigdiff] at (model_igdiff_node.north) {difference to real\\ information gain};

\end{tikzpicture}

\let\abstikzunit\undef
\let\scalingfactor\undef
\let\imgwidth\undef
\let\pwidth\undef
\let\labelheight\undef
\let\labelalpha\undef
\let\tikzunit\undef

%% file: figures/information_gains.pgf.tex
\ifdefined\noktikz
  \input{figures/informationgainsdata}
  \input{figures/ig_colors}
\else
  \usetikzlibrary{calc}
  \usetikzlibrary{shapes}
  \graphicspath{{/kyb/agmb/mkuemmerer/Documents/Uni/Bethge/Saliency/TPAMI/figures/}}
  \input{/kyb/agmb/mkuemmerer/Documents/Uni/Bethge/Saliency/TPAMI/figures/ig_colors}
  \input{/kyb/agmb/mkuemmerer/Documents/Uni/Bethge/Saliency/TPAMI/figures/informationgainsdata}
\fi

\newcommand\loglik[1]{\textsf{$\mathsf{#1}$\,bits/fix}}
\newcommand\auc[1]{\textsf{$\mathsf{#1}$\,\%}}


\ifdefined\abstikzunit
\else
	\newlength\abstikzunit
	\newlength\tikzunit
\fi

\setlength\abstikzunit{1cm}
\newcommand\scalingfactor{0.5}
\newcommand\imgwidth{5.0}
\newcommand\pwidth{3.5}
\newcommand\labelalpha{0.4}
\newcommand\labelheight{1.5}
\setlength\tikzunit{\scalingfactor\abstikzunit}

\setlength\abstikzunit{1cm}
\setlength\tikzunit{\scalingfactor\abstikzunit}



\begin{tikzpicture}[scale=\scalingfactor,font=\small\sffamily,every node/.style={inner sep=0,outer sep=0}]
  \tikzset{labelnode/.style={anchor=south, inner ysep=1,align=center,font=\tiny\sffamily,
    minimum height=\labelheight\tikzunit,text width=\pwidth\tikzunit,
    fill opacity=\labelalpha,text opacity=1.0}}
  \coordinate (col2) at (5.5, 0);
  \coordinate (colwidth) at (4.0, 0);
  \coordinate (colheight) at (0, -3);

  \coordinate (row2) at (0, -4);
  \coordinate (label_rel_fig) at (-4.5,3.5);
  \coordinate (ll_rel) at (2.5,-2);

  \coordinate (imagesep) at (0.0, -0.5);

  \node[labelnode,fill=colordensity]
    at ($ (col2) + 0*(colwidth) + 0.5*(\pwidth, 0) $)
    {\tiny \textsf{model density}};
  \node[labelnode,fill=colorsaliency]
    at ($ (col2) + 1*(colwidth) + 0.5*(\pwidth, 0) $)
    {model\\ image-based\\ saliency};
  \node[labelnode,fill=colorig]
     at ($ (col2) + 2*(colwidth) + 0.5*(\pwidth, 0) $)
    {model\\information gain};
  \node[labelnode,fill=colorigdiff]
     at ($ (col2) + 3*(colwidth) + 0.5*(\pwidth, 0) $)
    {difference to real\\ information gain};

  \foreach[count=\imgpos] \imgindex/\labelname in {\imgindexa/(a),\imgindexb/(b),\imgindexc/(c)}
  {
    \coordinate (imgcoord) at ($ \imgpos*4*(colheight) - 4*(colheight) + \imgpos*(imagesep) - (imagesep) $);

    \node[anchor=north east] at ($ (imgcoord) - (0.3, 0) $) {\textbf{\labelname}};

    \node[anchor=north west] at (imgcoord)
      {\includegraphics[width=\imgwidth\tikzunit]{informationgains-\imgindex-image.png}};

    \foreach[count=\rowpos] \modelname in {gold,eDN,BMS,AIM}
    {
	   \node[anchor=south east, rotate=90, color=gray] at ($ (imgcoord) + (col2) + \rowpos*(colheight) - (colheight)  $)
            {\tiny \modelname };

      \foreach \pos/\name in {0/p,1/ratio, 2/ig, 3/igdiff} {
          \node[draw,anchor=north west] at ($ (imgcoord) + (col2) + \pos*(colwidth) + \rowpos*(colheight) - (colheight) $)
            {\includegraphics[width=\pwidth\tikzunit]{informationgains-\imgindex-\modelname-\name.pdf}};
      }


        \node[anchor=north east, rotate=90] at ($ (imgcoord) + (col2) + 4*(colwidth) + \rowpos*(colheight) - (colheight) + (-0.3, -0.0) $)
          {\tiny \csname informationgain\imgindex\modelname ig\endcsname\ bit/fix};

    }
  }

%

\end{tikzpicture}

\let\abstikzunit\undef
\let\scalingfactor\undef
\let\imgwidth\undef
\let\pwidth\undef
\let\labelheight\undef
\let\labelalpha\undef
\let\tikzunit\undef

%% file: figures/information_gains_scatter_combined.pgf.tex
\ifdefined\noktikz
  \input{figures/informationgainsdata}
  \newcommand\figurepath{figures}
\else
  \usetikzlibrary{calc}
  \usetikzlibrary{shapes}
  \graphicspath{{/kyb/agmb/mkuemmerer/Documents/Uni/Bethge/Saliency/TPAMI/figures/}}
  \newcommand\figurepath{/kyb/agmb/mkuemmerer/Documents/Uni/Bethge/Saliency/TPAMI/figures}

\fi




\begin{tikzpicture}[font=\small\sffamily,every node/.style={inner sep=0,outer sep=0}]
  \coordinate (col2) at (7.5,0);
  \coordinate (rowsep) at (0,-0.7);
  \coordinate (labelsep) at (-0.5, 0);

  \node[anchor=north west] (eDN1) at (0,0) {
      
     \begin{minipage}[t]{8.5cm}
        {
             \newcommand\imageprefix{general}
             \newcommand\imagesuffix{image}
             \newcommand\imageext{png}
			 \newcommand\modelname{eDN}
             \input{\figurepath/information_gains_scatter_all.pgf}
       }
     \end{minipage}
     };

  \node[anchor=north west] (eDN2) at ($ (eDN1.north west) + (col2) $) {
       \begin{minipage}[t]{8.5cm}
        {
             \newcommand\imageprefix{eDN}
             \newcommand\imagesuffix{igdiff}
             \newcommand\imageext{pdf}
			 \newcommand\modelname{eDN}
             \input{\figurepath/information_gains_scatter_all.pgf}
       }
     \end{minipage}
     };

  \node[anchor=north east] at ($ (eDN1.north west) + (labelsep) $) {\textbf{eDN}};

  \node[anchor=north west] (Judd1) at ($ (eDN1.south west) + (rowsep) $) {
      
     \begin{minipage}[t]{8.5cm}
        {
             \newcommand\imageprefix{general}
             \newcommand\imagesuffix{image}
             \newcommand\imageext{png}
			 \newcommand\modelname{Judd}
             \input{\figurepath/information_gains_scatter_all.pgf}
       }
     \end{minipage}
     };

  \node[anchor=north west] (Judd2) at ($ (Judd1.north west) + (col2) $) {
       \begin{minipage}[t]{8.5cm}
        {
             \newcommand\imageprefix{Judd}
             \newcommand\imagesuffix{igdiff}
             \newcommand\imageext{pdf}
			 \newcommand\modelname{Judd}
             \input{\figurepath/information_gains_scatter_all.pgf}
       }
     \end{minipage}
     };

  \node[anchor=north east] at ($ (Judd1.north west) + (labelsep) $) {\textbf{Judd}};

\end{tikzpicture}

%% file: figures/modeling_temporal.pgf.tex
\ifdefined\noktikz
  \input{figures/model_colors}
\else
  \usetikzlibrary{calc}
  \usetikzlibrary{shapes}
  \graphicspath{{/kyb/agmb/mkuemmerer/Documents/Uni/Bethge/Saliency/TPAMI/figures/}}

  \input{/kyb/agmb/mkuemmerer/Documents/Uni/Bethge/Saliency/TPAMI/figures/model_colors}
\fi

\newcommand\loglik[1]{\textsf{$\mathsf{#1}$\,bit/fix}}
\newcommand\auc[1]{\textsf{$\mathsf{#1}$\,\%}}


\ifdefined\abstikzunit
	\setlength\abstikzunit{1cm}
	\newcommand\scalingfactor{0.8}
	\newcommand\imgwidth{5}
	\setlength\tikzunit{\scalingfactor\abstikzunit}
\else
	\newlength\abstikzunit
	\setlength\abstikzunit{1cm}
	\newlength\tikzunit
	\newcommand\scalingfactor{0.8}
	\newcommand\imgwidth{5}
	\setlength\tikzunit{\scalingfactor\abstikzunit}
\fi

\begin{tikzpicture}[scale=\scalingfactor,font=\sffamily]
  \definecolor{fillcolor}{rgb}{0.9, 0.9, 0.9};
  \tikzset{boxstyle/.style={fill=fillcolor, rounded corners=2mm}};


  \coordinate (leftcolwidth) at (6.7,0);
  \coordinate (leftcolheight) at (0,-1.2);
  \newcommand{\leftcolimgheight}{1.0}

  \coordinate (leftcolsize) at ($ (leftcolwidth) + (leftcolheight) $);

  \coordinate (leftcolsep) at (0, -0.75);
  \coordinate (leftcolop) at ($ (leftcolheight) + 0.5*(leftcolsep) + 0.5*(leftcolwidth) $);

  \coordinate (lefttotalcolheight) at ($ (leftcolheight) + (leftcolsep) $);

  \coordinate (leftcolimage) at (0.2,-0.1);
  \coordinate (leftcoltext) at ($ 0.5*(leftcolheight) + (1.5,0) $);

  \coordinate (leftcoltemporalsep) at (0,-0.1);

  \coordinate (col2) at (5.9, 0);
  \coordinate (row2) at (0, -4.2);
  \coordinate (label_rel_subfig) at (-0.6, 0);


  \coordinate (rightcol) at ($ (leftcolwidth) + (1,0) $);
  \newcommand\rightcolwidth{9}

  \coordinate (rightcolheight) at (0,-6);

  \node[anchor=north west] at ($ (label_rel_subfig) + (-0.2, 0) $) {a)};


  \path[draw, boxstyle, fill=colorblur]
    ($ 0*(lefttotalcolheight) $) rectangle ($  0*(lefttotalcolheight) + (leftcolsize) $);
  \node[anchor=north west, inner sep=0] at ($ 0*(lefttotalcolheight) + (leftcolimage) $)
    {\includegraphics[height=\leftcolimgheight\tikzunit]{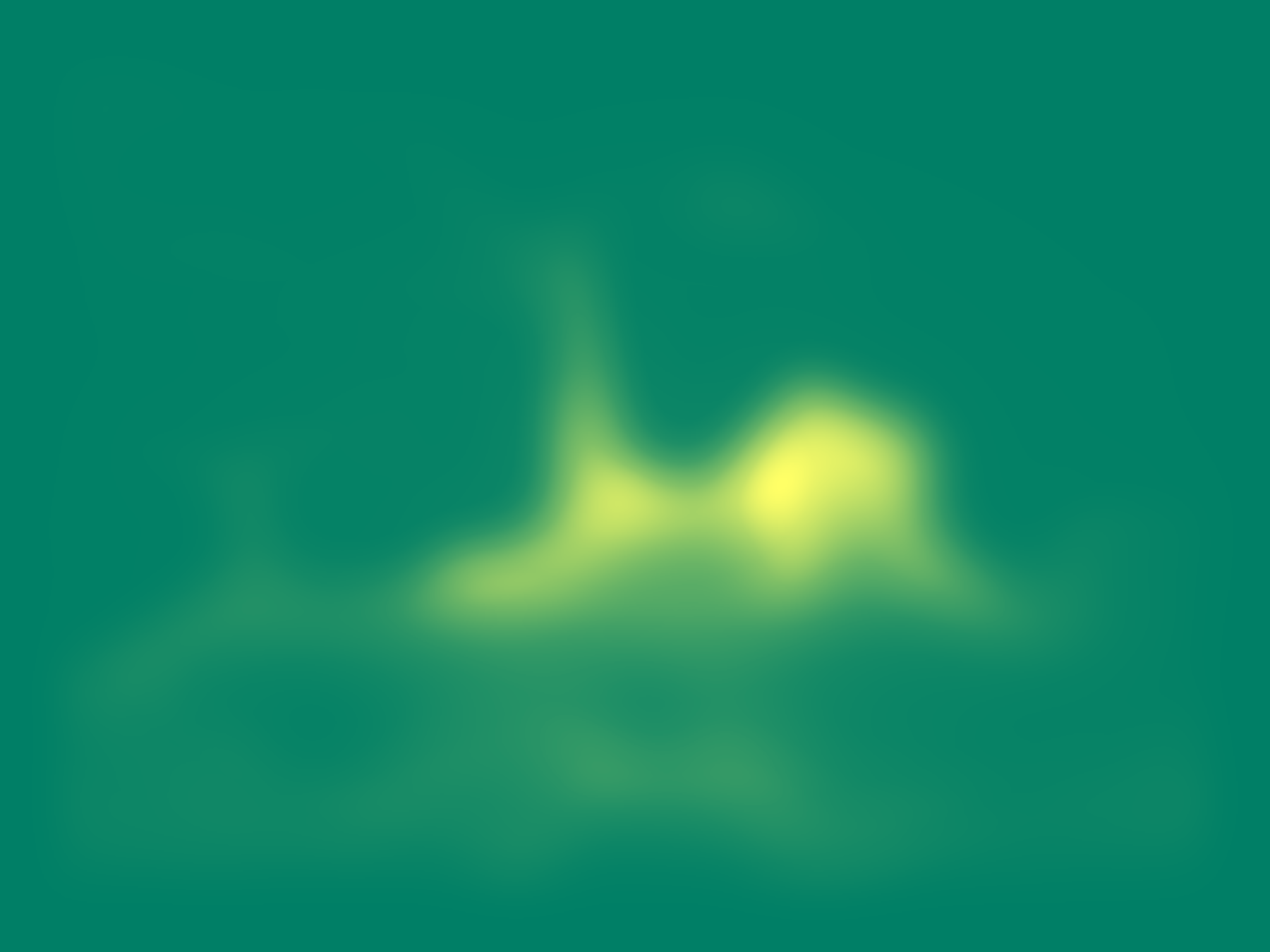}};
  \node[anchor=west] at ($ 0*(lefttotalcolheight) + (leftcoltext) $) {Stationary model};

  \draw[->,thick] ($ (leftcolheight) + 0.5*(leftcolwidth) + (0, -0.1) $) -- +($ (leftcolsep) + (0, 0.2) $);




  \path[draw, boxstyle, fill=colorexcitation]
    ($ 1*(lefttotalcolheight) + (leftcoltemporalsep) $) rectangle +($ (leftcolsize) $);
  \node[anchor=north west, inner sep=0] at ($ 1*(lefttotalcolheight) + (leftcoltemporalsep) + (leftcolimage) $)
    {\includegraphics[height=\leftcolimgheight\tikzunit]{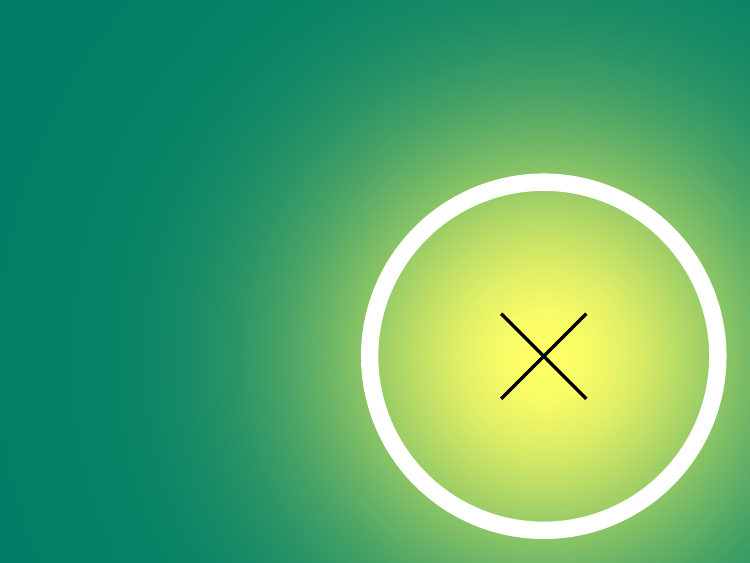}};
  \node[anchor=west] at ($ 1*(lefttotalcolheight) + (leftcoltemporalsep) + (leftcoltext) $) {Self excitation};


  \draw[->,thick] ($ 1*(lefttotalcolheight) + (leftcolheight) + (leftcoltemporalsep) + 0.5*(leftcolwidth) + (0, -0.1) $) -- +($ (leftcolsep) + (0, 0.2) $);


  \path[draw, boxstyle, fill=colorexcentric]
    ($ 2*(lefttotalcolheight) + (leftcoltemporalsep) $) rectangle +(leftcolsize);
  \node[anchor=north west, inner sep=0] at ($ 2*(lefttotalcolheight) + (leftcoltemporalsep) + (leftcolimage) $)
    {\includegraphics[height=\leftcolimgheight\tikzunit]{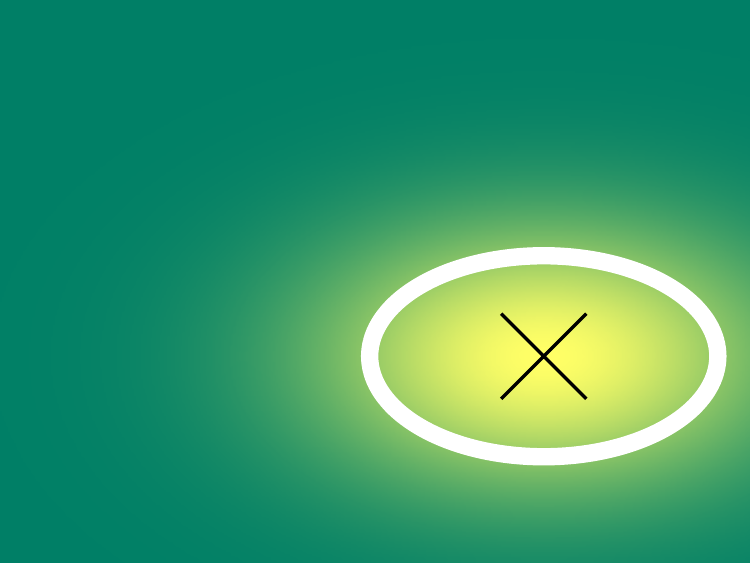}};
  \node[anchor=west] at ($ 2*(lefttotalcolheight) + (leftcoltemporalsep) + (leftcoltext) $) {Elliptical self excitation};


  \node[anchor=north west] at ($ (rightcol) + (label_rel_subfig) $) {b)};

  \node[anchor=north west] at ($ (rightcol) $)
   {\includegraphics[width=\rightcolwidth\tikzunit]%
     {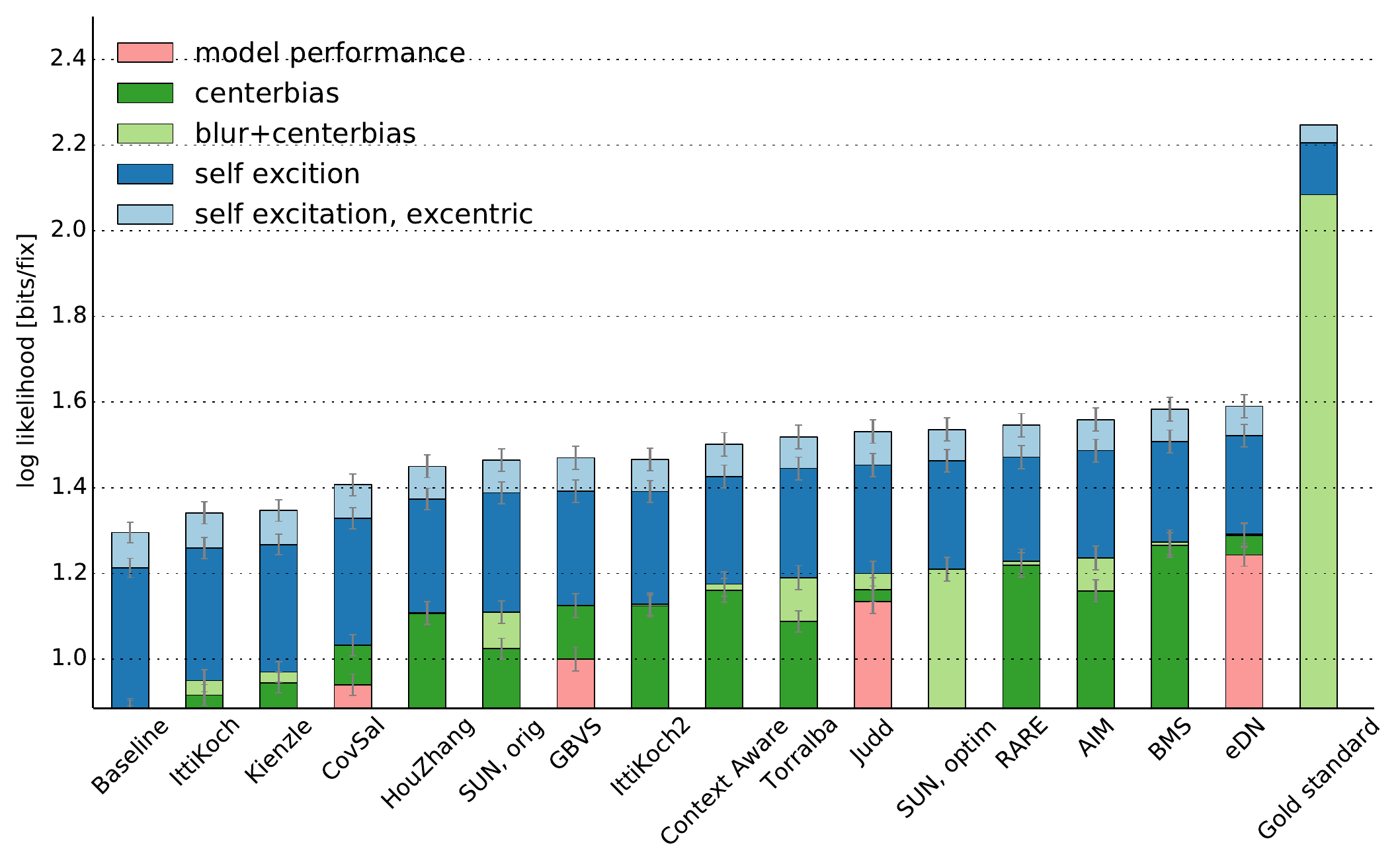}};


\end{tikzpicture}

\let\abstikzunit\undef
\let\scalingfactor\undef
\let\imgwidth\undef

%% file: figures/auc_problems.pgf.tex
\ifdefined\noktikz
  \input{figures/auc_examples}
\else
  \usetikzlibrary{calc}
  \usetikzlibrary{shapes}
  \graphicspath{{/kyb/agmb/mkuemmerer/Documents/Uni/Bethge/Saliency/TPAMI/figures/}}

  \input{/kyb/agmb/mkuemmerer/Documents/Uni/Bethge/Saliency/TPAMI/figures/auc_examples}
\fi

\newcommand\loglik[1]{\textsf{$\mathsf{#1}$\,bits/fix}}
\newcommand\auc[1]{\textsf{$\mathsf{#1}$\,\%}}


\ifdefined\abstikzunit
\else
	\newlength\abstikzunit
	\newlength\tikzunit
\fi

\setlength\abstikzunit{1cm}
\newcommand\scalingfactor{0.6}
\newcommand\imgwidth{5}
\setlength\tikzunit{\scalingfactor\abstikzunit}

\setlength\abstikzunit{1cm}
\setlength\tikzunit{\scalingfactor\abstikzunit}

\begin{tikzpicture}[scale=\scalingfactor,font=\small\sffamily]
  \coordinate (col2) at (5.5, 0);
  \coordinate (row2) at (0, -4);
  \coordinate (label_rel_fig) at (-4.5,3.5);
  \coordinate (ll_rel) at (2.5,-2);

  \node (labela) at (label_rel_fig) {\textsf{a)}};

  \node (modelpredictions) at ($ (col2) + (0, 4.3) $) {Model predictions};
  \node[rotate=90] (fixations) at ($ (row2) + (-5.7, 0) $) {Fixations};

  \node[anchor=north east] (fix_high)  at (0,0)
    {\includegraphics[width=\imgwidth\tikzunit]{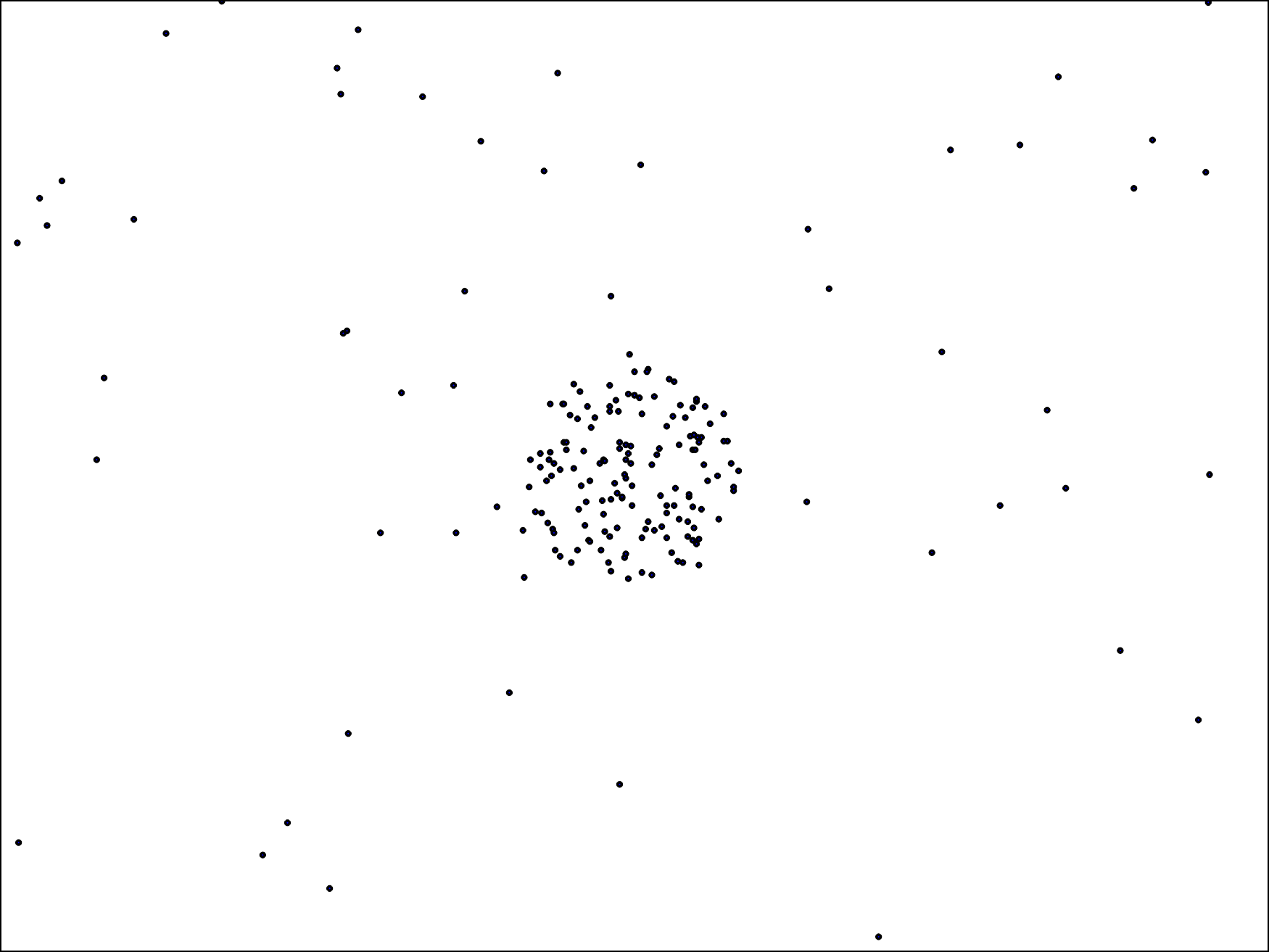}};
  \node[anchor=north east] (fix_low)  at (row2)
    {\includegraphics[width=\imgwidth\tikzunit]{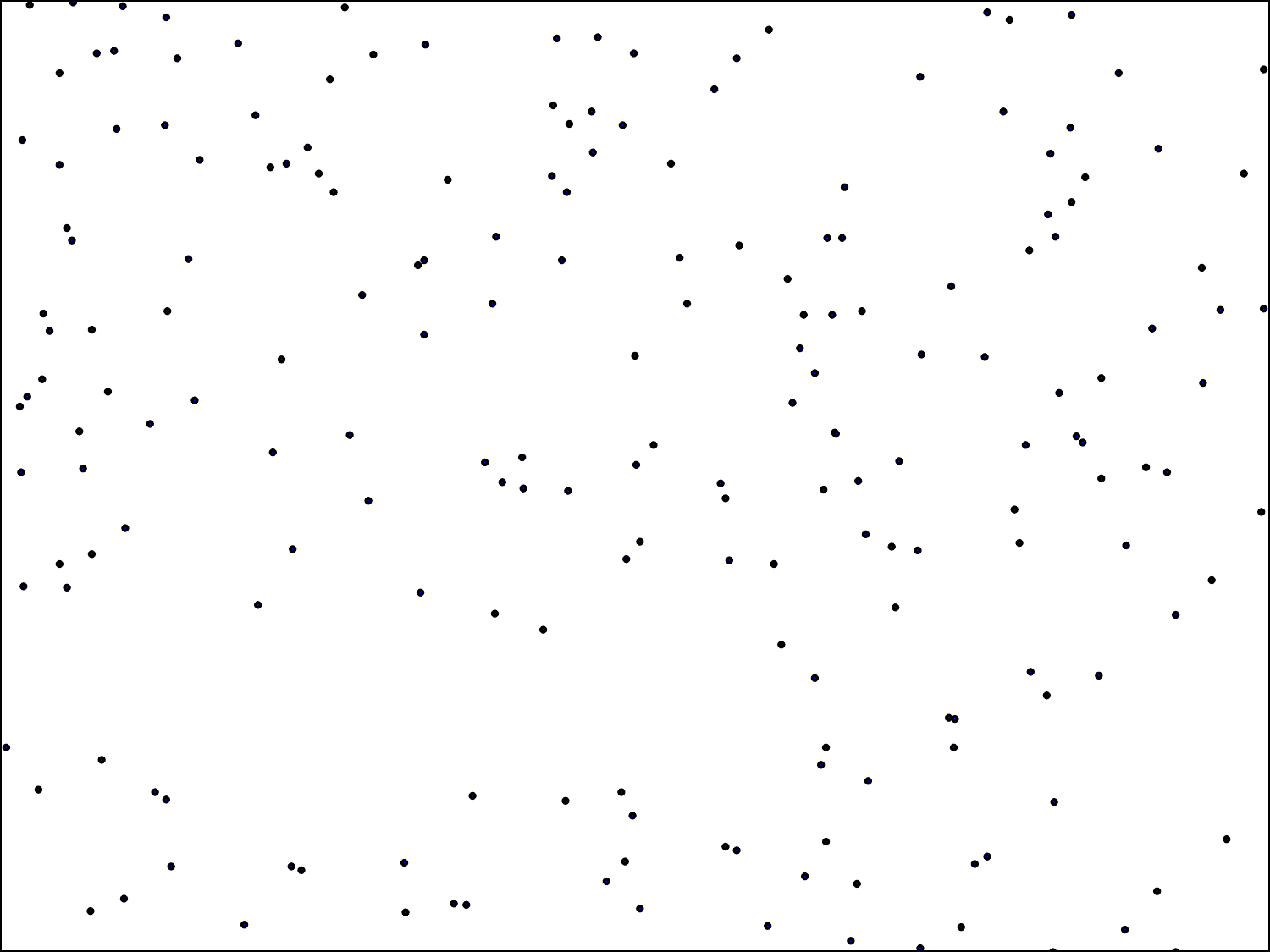}};

  \node[anchor=south west] (fix_high)  at (0,0)
    {\includegraphics[width=\imgwidth\tikzunit]{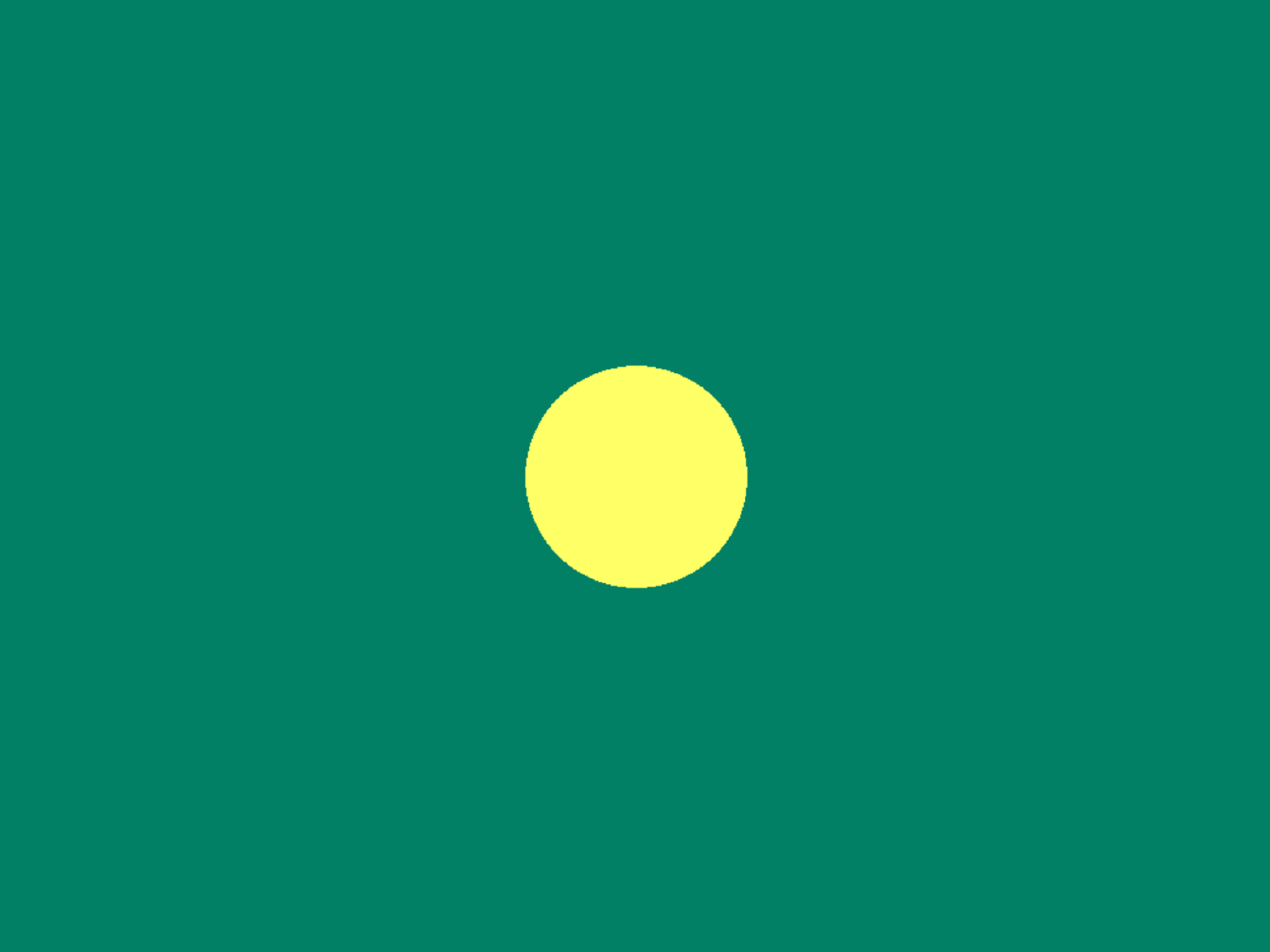}};

  \node[anchor=south west] (fix_high)  at (col2)
    {\includegraphics[width=\imgwidth\tikzunit]{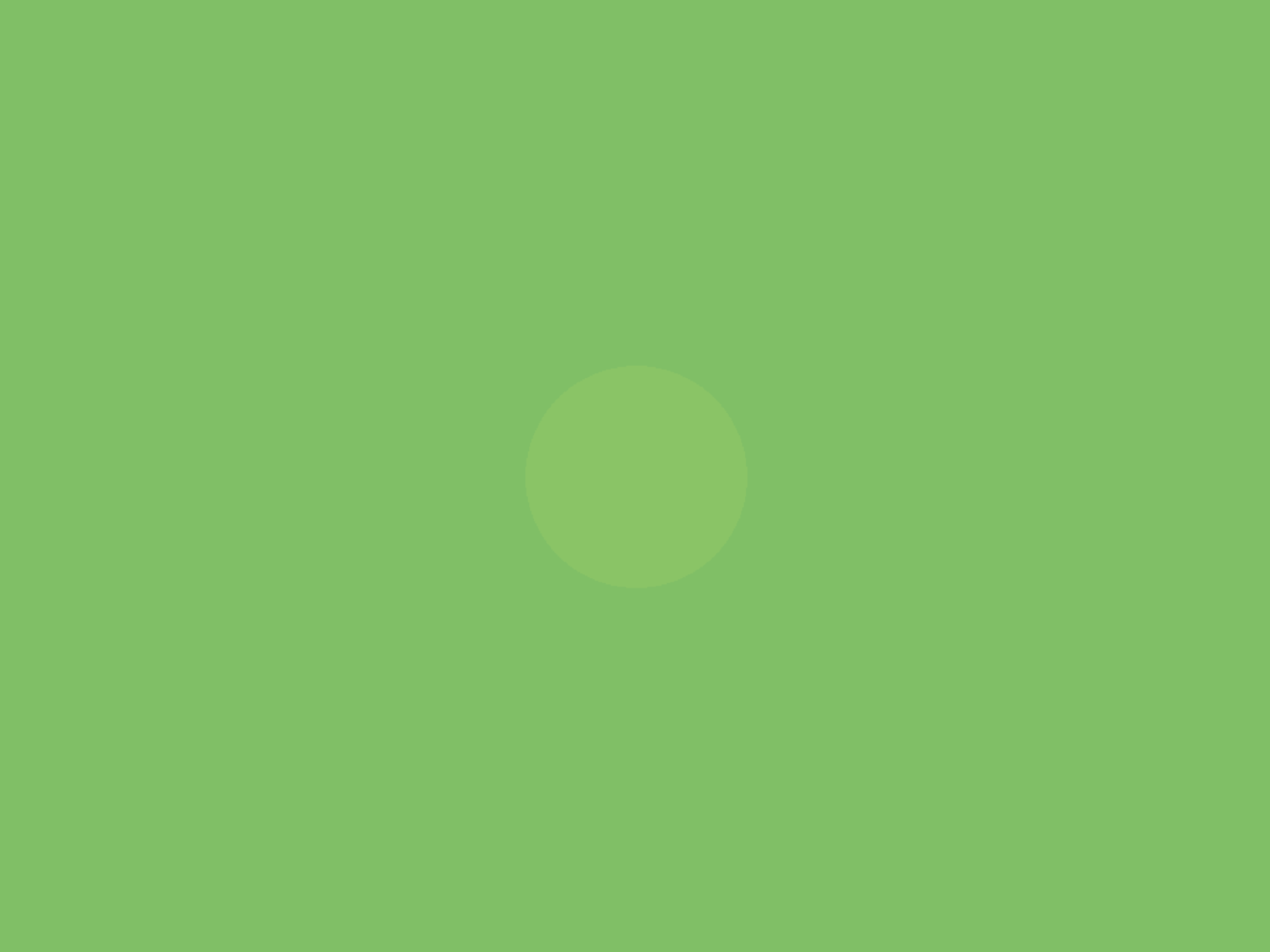}};

  \node[rectangle split, rectangle split parts=2, anchor=center] at (ll_rel) {\loglik\contrastHighFixHighSmapLoglikelihood\nodepart{second}\auc\contrastHighFixHighSmapAUC};

  \node[rectangle split, rectangle split parts=2, anchor=center] at ($ (col2) + (ll_rel) $) {\loglik\contrastHighFixLowSmapLoglikelihood\nodepart{second}\auc\contrastHighFixLowSmapAUC};

  \node[rectangle split, rectangle split parts=2, anchor=center] at ($ (row2) + (ll_rel) $) {\loglik\contrastLowFixHighSmapLoglikelihood\nodepart{second}\auc\contrastLowFixHighSmapAUC};

  \node[rectangle split, rectangle split parts=2, anchor=center] at ($ (row2) + (col2) + (ll_rel) $) {\loglik\contrastLowFixLowSmapLoglikelihood\nodepart{second}\auc\contrastLowFixLowSmapAUC};

  \coordinate (fig2) at (0,-13);
  \coordinate (hline_rel_fig2) at (0,4.5);

  \draw ($ (fig2) - (col2) + (hline_rel_fig2) $) -- ($ (fig2) +(hline_rel_fig2) + 2*(col2) $);


  \node (labela) at ($ (fig2) + (label_rel_fig) $) {\textsf{b)}};

  \node[anchor=north east] (fix)  at ($(fig2) + (0,0) $)
    {\includegraphics[width=\imgwidth\tikzunit]{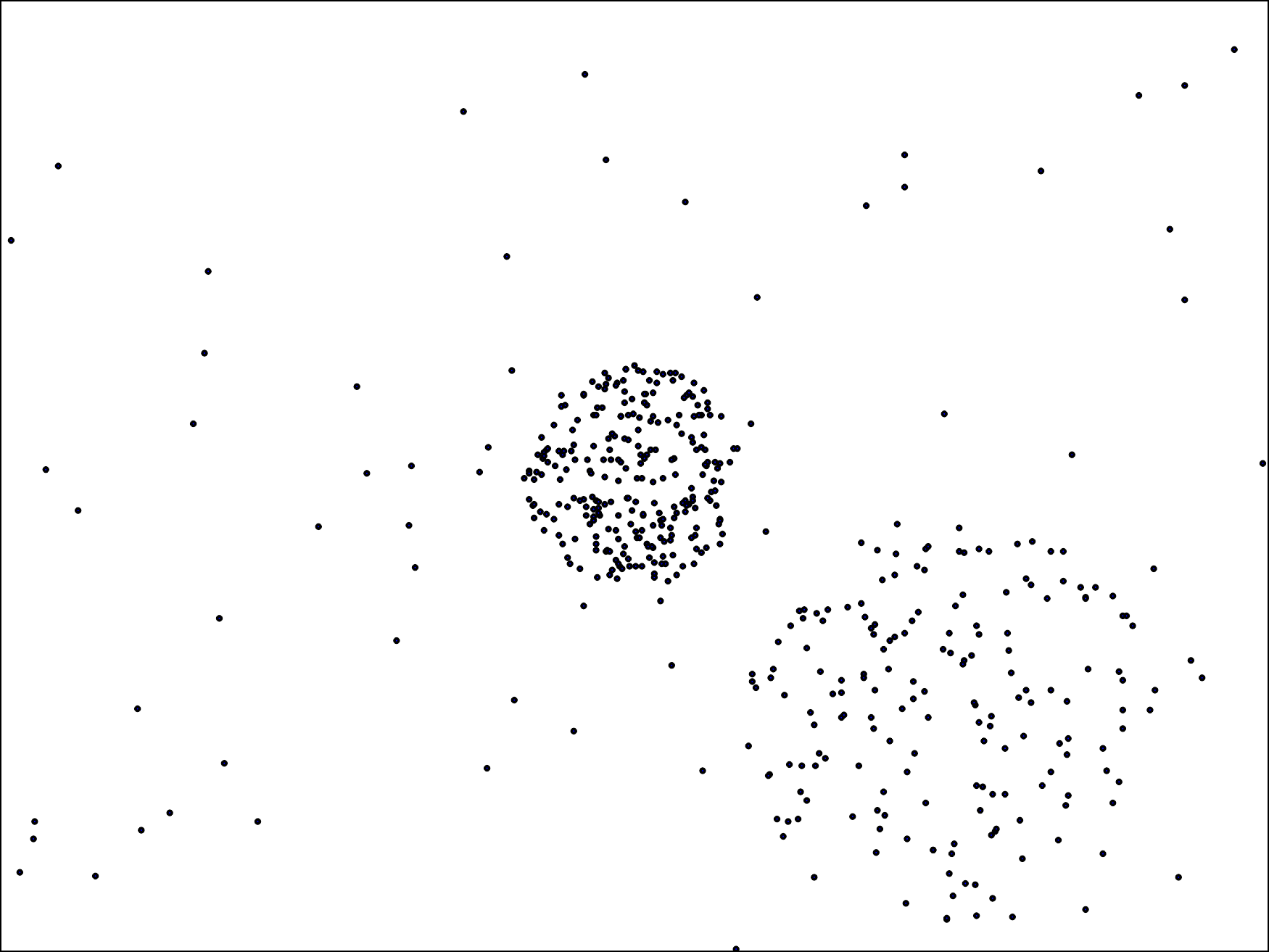}};

  \node[anchor=south west] (sal_good)  at ($ (fig2)  $)
    {\includegraphics[width=\imgwidth\tikzunit]{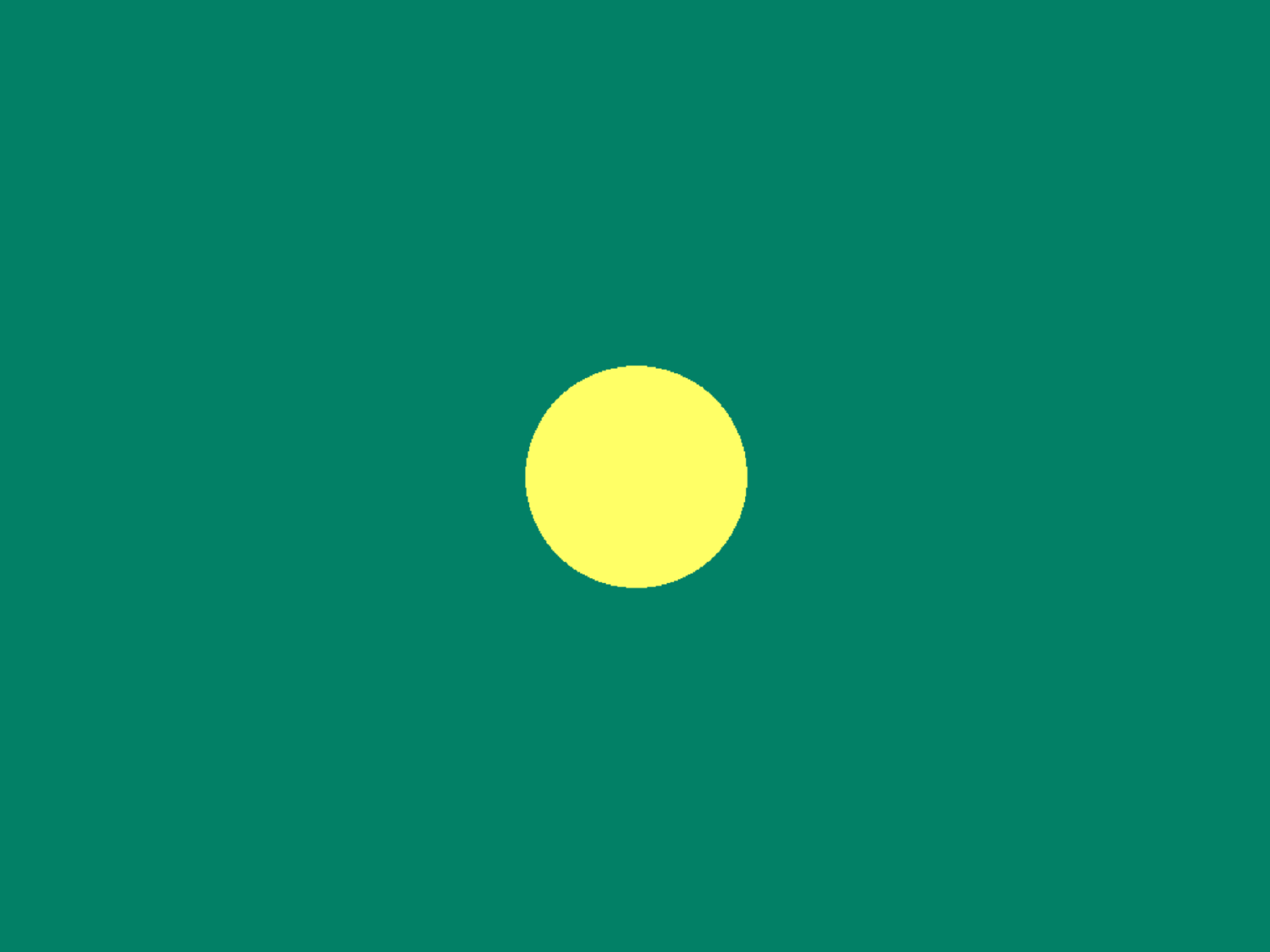}};

  \node[anchor=south west] (sal_bad)  at ($ (fig2) + (col2) $)
    {\includegraphics[width=\imgwidth\tikzunit]{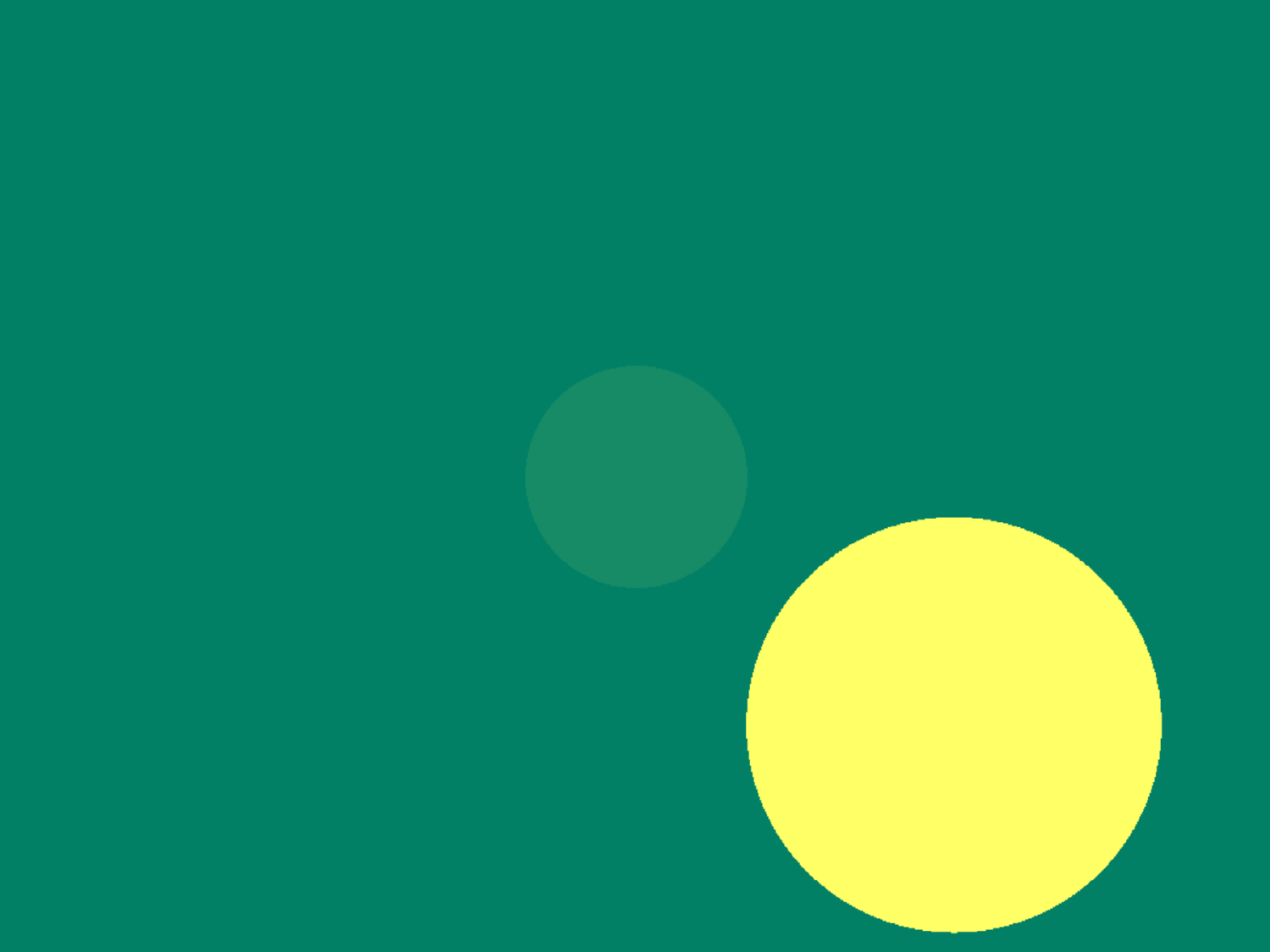}};

  \node[rectangle split, rectangle split parts=2, anchor=center] at ($ (fig2)  + (ll_rel) $) {\loglik\orderingGoodLoglikelihood\nodepart{second}\auc\orderingGoodAUC};

  \node[rectangle split, rectangle split parts=2, anchor=center] at ($ (fig2) + (col2) + (ll_rel) $) {\loglik\orderingBadLoglikelihood\nodepart{second}\auc\orderingBadAUC};


  \coordinate (colorbar_label_correction) at (0, 0.2);

  \node[anchor=south west] (colorbar)  at ($ (fig2) +(row2) +(col2)$)
    {\includegraphics[width=\imgwidth\tikzunit]{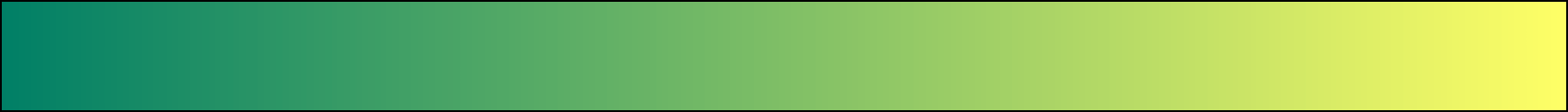}};

  \node[anchor=north west] at ($ (colorbar.south west) + (colorbar_label_correction) $)
    {\textsf{\tiny{low density}}};

  \node[anchor=north east] at ($ (colorbar.south east) + (colorbar_label_correction) $)
    {\textsf{\tiny{high density}}};

\end{tikzpicture}

\let\abstikzunit\undef
\let\scalingfactor\undef
\let\imgwidth\undef
\let\tikzunit\undef

%% file: figures/kl_divergence.pgf.tex
\ifdefined\noktikz
\else
  \usetikzlibrary{calc}
  \usetikzlibrary{shapes}
  \graphicspath{{/kyb/agmb/mkuemmerer/Documents/Uni/Bethge/Saliency/TPAMI/figures/}}

\fi

\ifdefined\abstikzunit
\else
    \newlength\abstikzunit
    \newlength\tikzunit
\fi

\setlength\abstikzunit{1cm}
\newcommand\scalingfactor{1.3}
\newcommand\imgwidth{5.3}
\newcommand\smallimgwidth{2.5}
\setlength\tikzunit{\scalingfactor\abstikzunit}

\setlength\abstikzunit{1cm}
\setlength\tikzunit{\scalingfactor\abstikzunit}

\begin{tikzpicture}[scale=\scalingfactor]
  \coordinate (col2) at (5.6, 0);
  \coordinate (row2) at (0, -4.6);

  \coordinate (smallcol2) at (2.9, 0);
  \coordinate (smallrow2) at (0, -2.1);

  \coordinate (label_rel_subfig) at (-0.45, -0.0);
  \coordinate (label_rel_subfig2) at (-0.1, -0.0);


  \node[anchor=north west] at (0,0)
    {\includegraphics[width=\smallimgwidth\tikzunit]{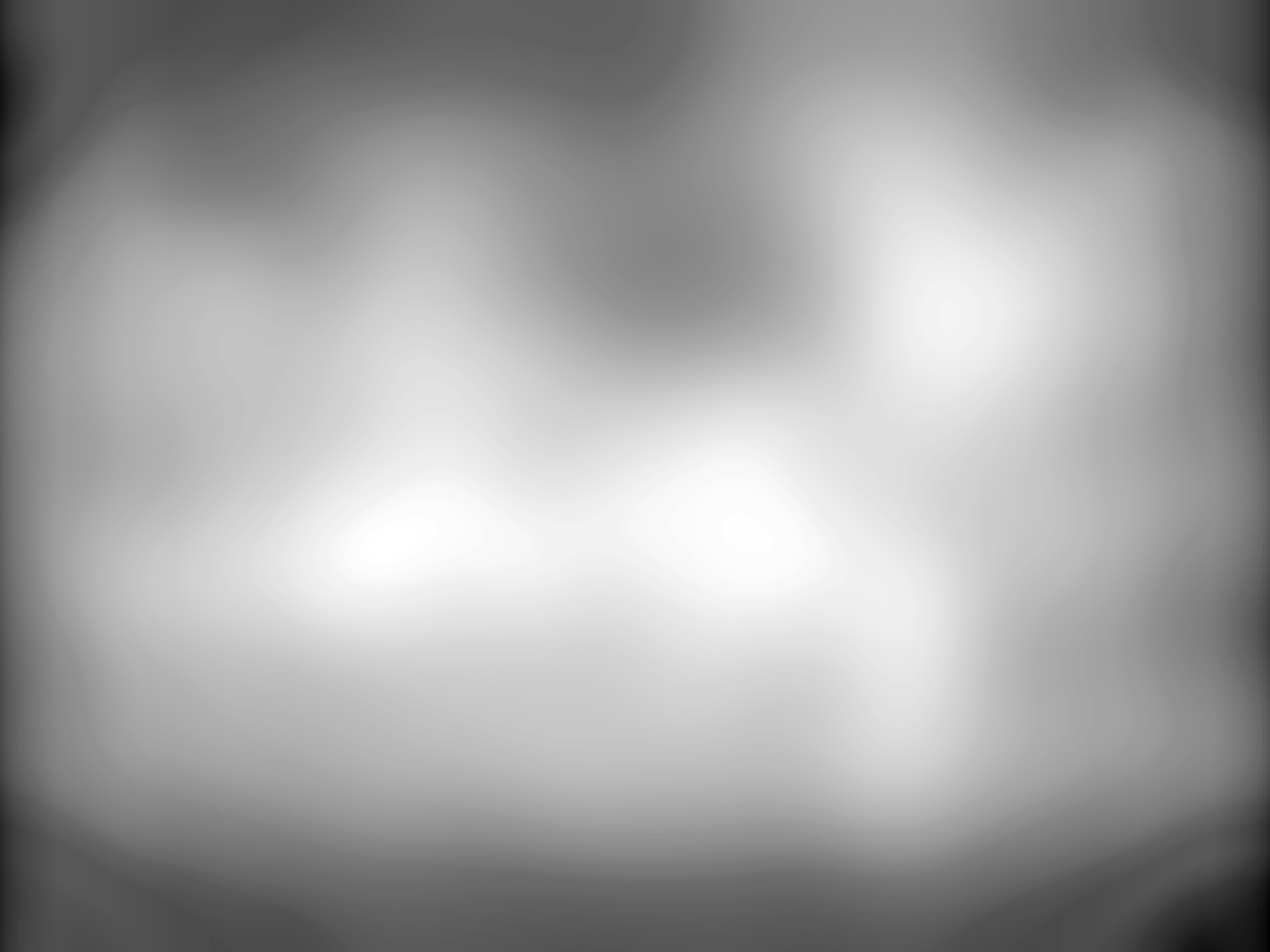}};

  \node[anchor=north west] at (smallcol2)
    {\includegraphics[width=\smallimgwidth\tikzunit]{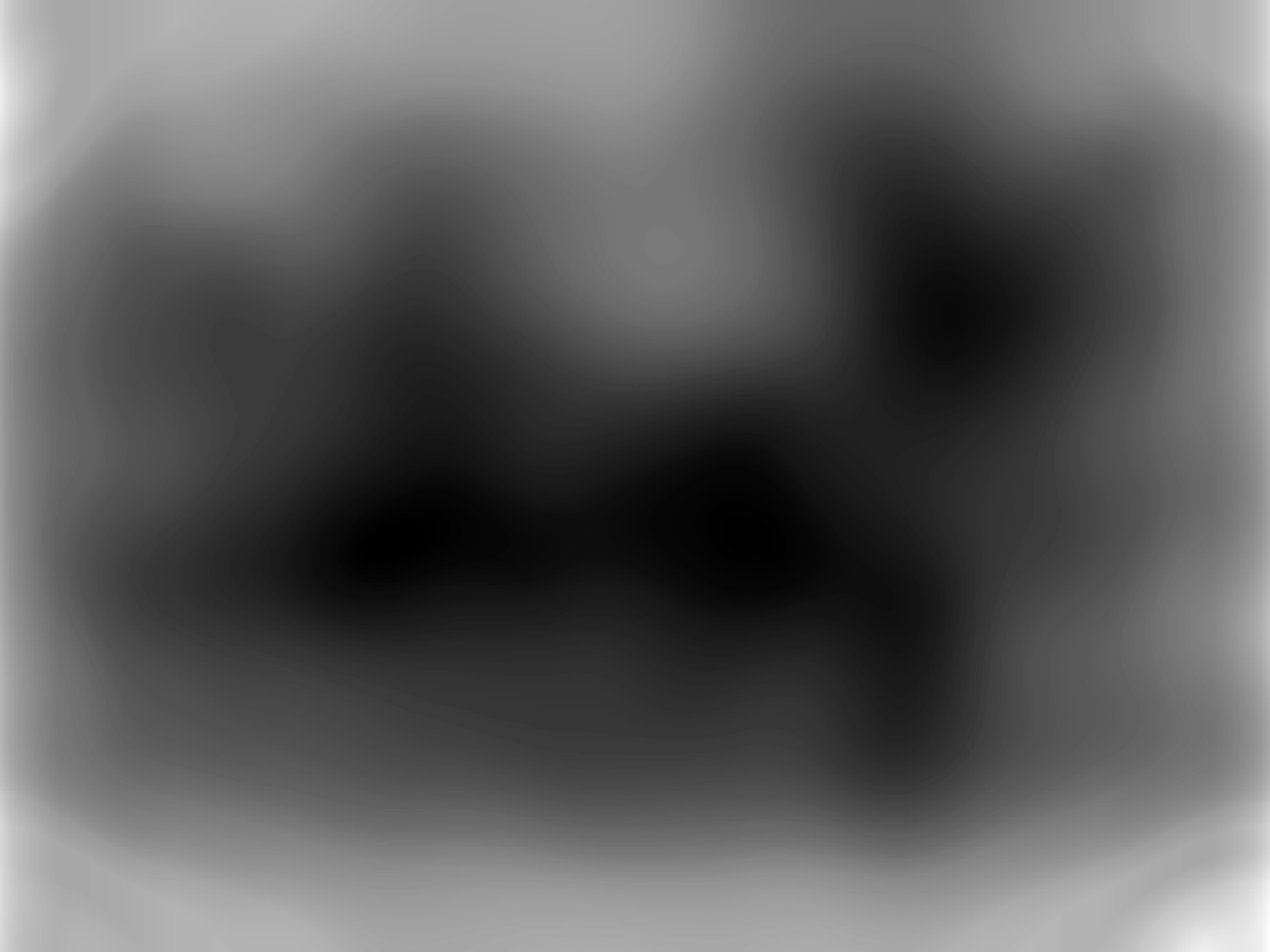}};

  \node[anchor=north west] at (smallrow2)
    {\includegraphics[width=\smallimgwidth\tikzunit]{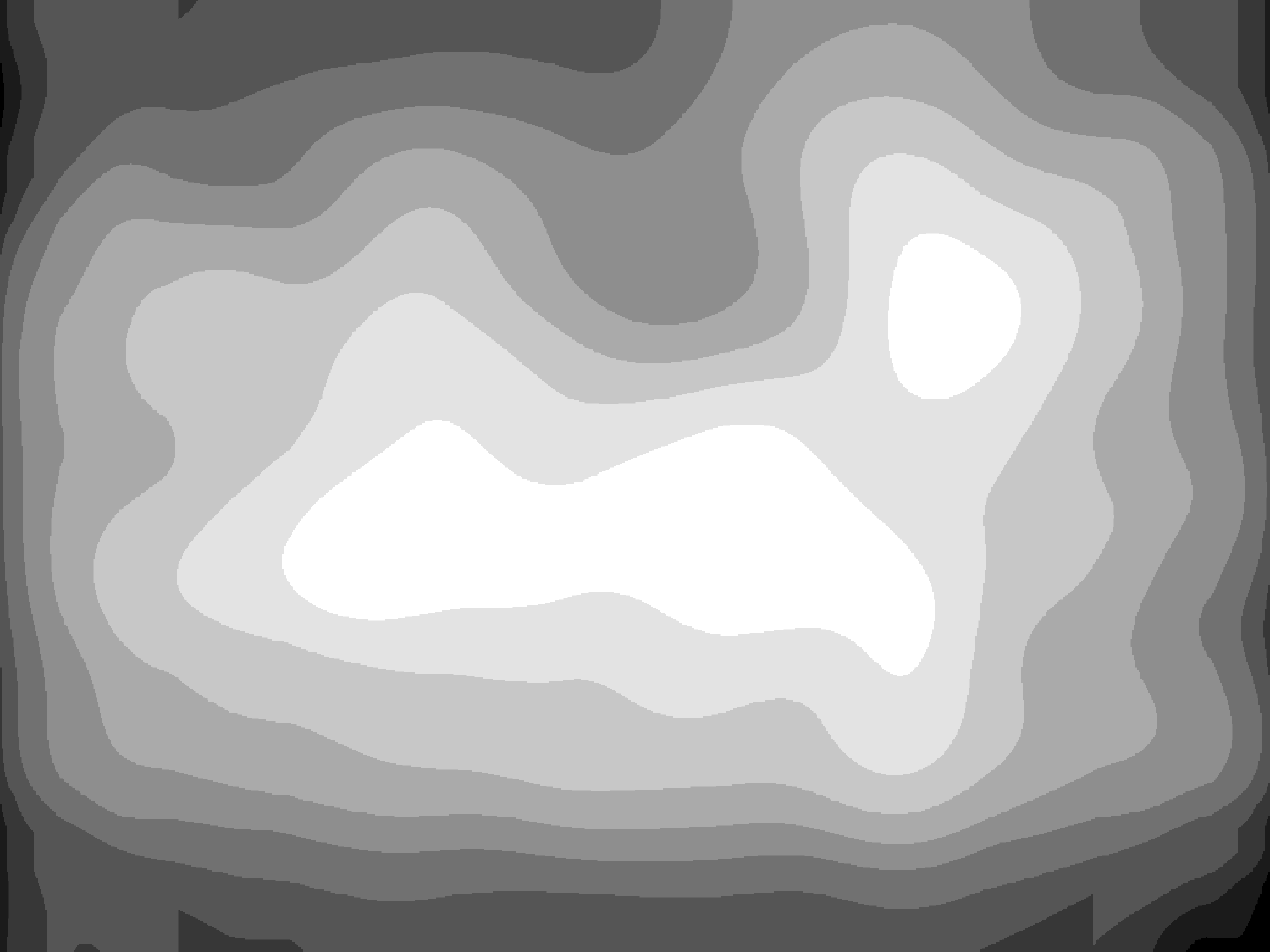}};

  \node[anchor=north west] at ($ (smallrow2) + (smallcol2) $)
    {\includegraphics[width=\smallimgwidth\tikzunit]{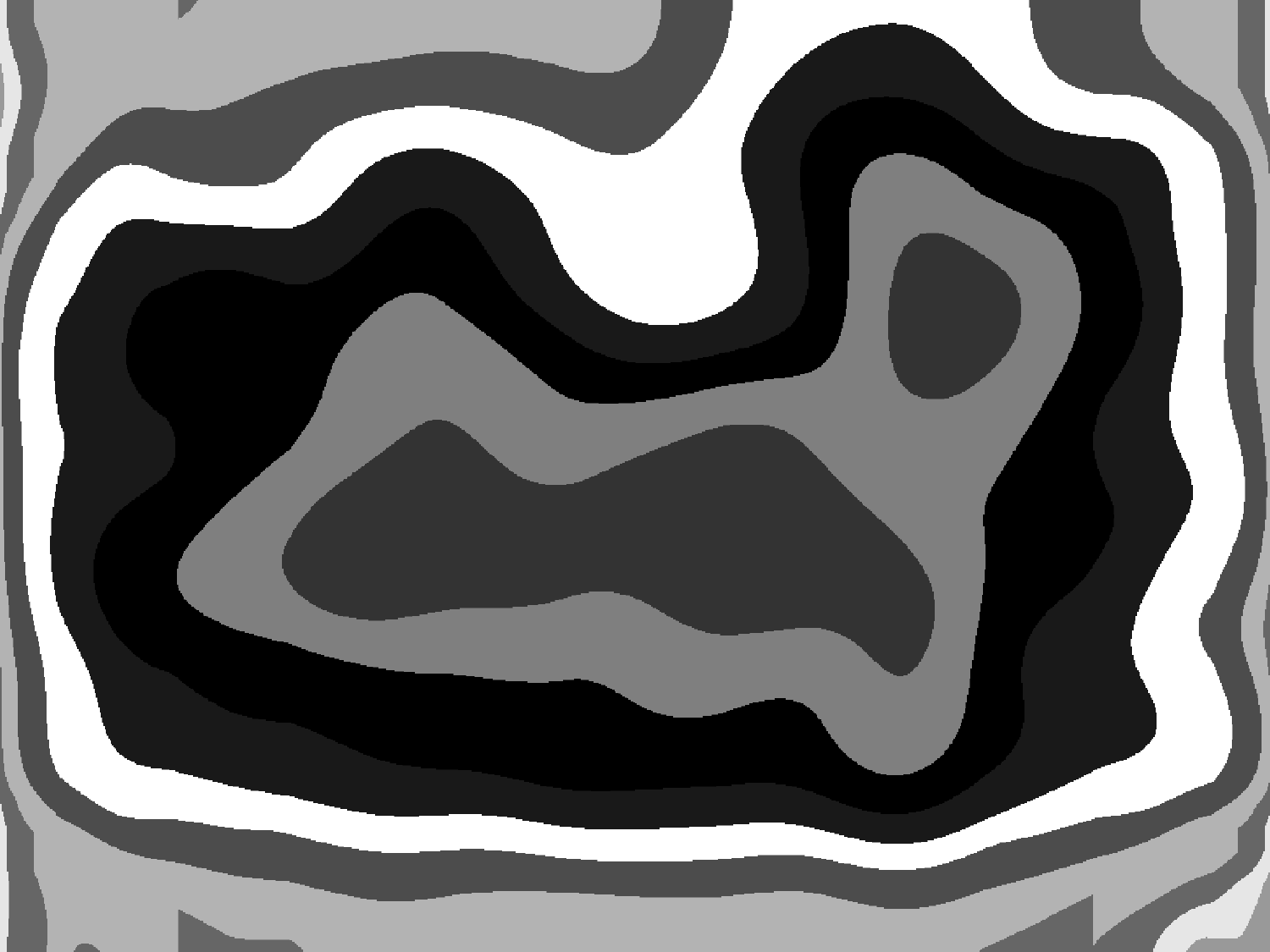}};

%
%
%
%

\end{tikzpicture}

\let\abstikzunit\undef
\let\scalingfactor\undef
\let\imgwidth\undef
\let\tikzunit\undef

%% file: kl_divergence_table.tex
\newcommand{\mycite}[1]{\cite{#1}}

\begin{tabular}{p{1cm}p{3cm}p{8cm}}
Paper & KL-Divergence & estimate of true distribution \\ \hline

\mycite{itti2005bayesian}
    & fixation-based
    & \\

\mycite{itti2005principled}
    & fixation-based
    & \\

\mycite{itti2005quantifying}
    & fixation-based
    & \\

\mycite{baldi2005attention}
    & fixation-based
    & \\

\mycite{Zhang2008} 
    & fixation-based
    & \\ 

\mycite{Bruce2009} 
    & fixation-based
    & \\

\mycite{baldi2010bits}
    & fixation-based
    & \\

\mycite{wang2010measuring}
    & fixation-based
    & \\

\mycite{Borji2013}
    & fixation based
    & \\
\mycite{Borji2013b}
    & fixation-based
    & \\
\mycite{Borji2013d}
    & fixation-based
    & \\
%


\mycite{rajashekar2004point}
    & image-based
    & Gaussian kernel, width of fovea\\

\mycite{Tatler2005} 
    & image-based
    & 2d-histograms, bins of $2^\circ \times 2^\circ$ and $10^{-5}$ added as prior\\

\mycite{lemeur2007predicting}
    & image-based
    & precision of the eye tracking.\\

\mycite{Wilming2011} 
    & image-based
    & Gaussian with $2^\circ$, motivated by fovea + eye tracker.\\

\mycite{lemeur2013methods}
    & image-based, fixation based
    & Gaussian kernel density estimate, kernel size $1^\circ$ of visual angle.\\

\mycite{Riche2013}
    & image-based
    & not stated\\

\mycite{Engbert2014} 
    & image-based
    & ``kernel-density estimates with bandwidth parameters chosen according to Scott's rule''\\



\end{tabular}

%% file: figures/other_measures.pgf.tex
\ifdefined\noktikz
\else
  \usetikzlibrary{calc}
  \usetikzlibrary{shapes}
  \graphicspath{{/kyb/agmb/mkuemmerer/Documents/Uni/Bethge/Saliency/TPAMI/figures/}}

\fi

\ifdefined\abstikzunit
\else
    \newlength\abstikzunit
    \newlength\tikzunit
\fi

\setlength\abstikzunit{1cm}
\newcommand\scalingfactor{1}
\newcommand\imgwidth{8.5}
\newcommand\smallimgwidth{2.5}
\setlength\tikzunit{\scalingfactor\abstikzunit}

\setlength\abstikzunit{1cm}
\setlength\tikzunit{\scalingfactor\abstikzunit}

\begin{tikzpicture}[scale=\scalingfactor]
  \coordinate (col2) at (8, 0);
  \coordinate (row2) at (0, -4.6);

  \coordinate (smallcol2) at (2.9, 0);
  \coordinate (smallrow2) at (0, -2.1);

  \coordinate (label_rel_subfig) at (-0.0, -0.4);
  \coordinate (label_rel_subfig2) at (label_rel_subfig);


  \node[anchor=north west] at (0,0)
    {\includegraphics[width=\imgwidth\tikzunit]{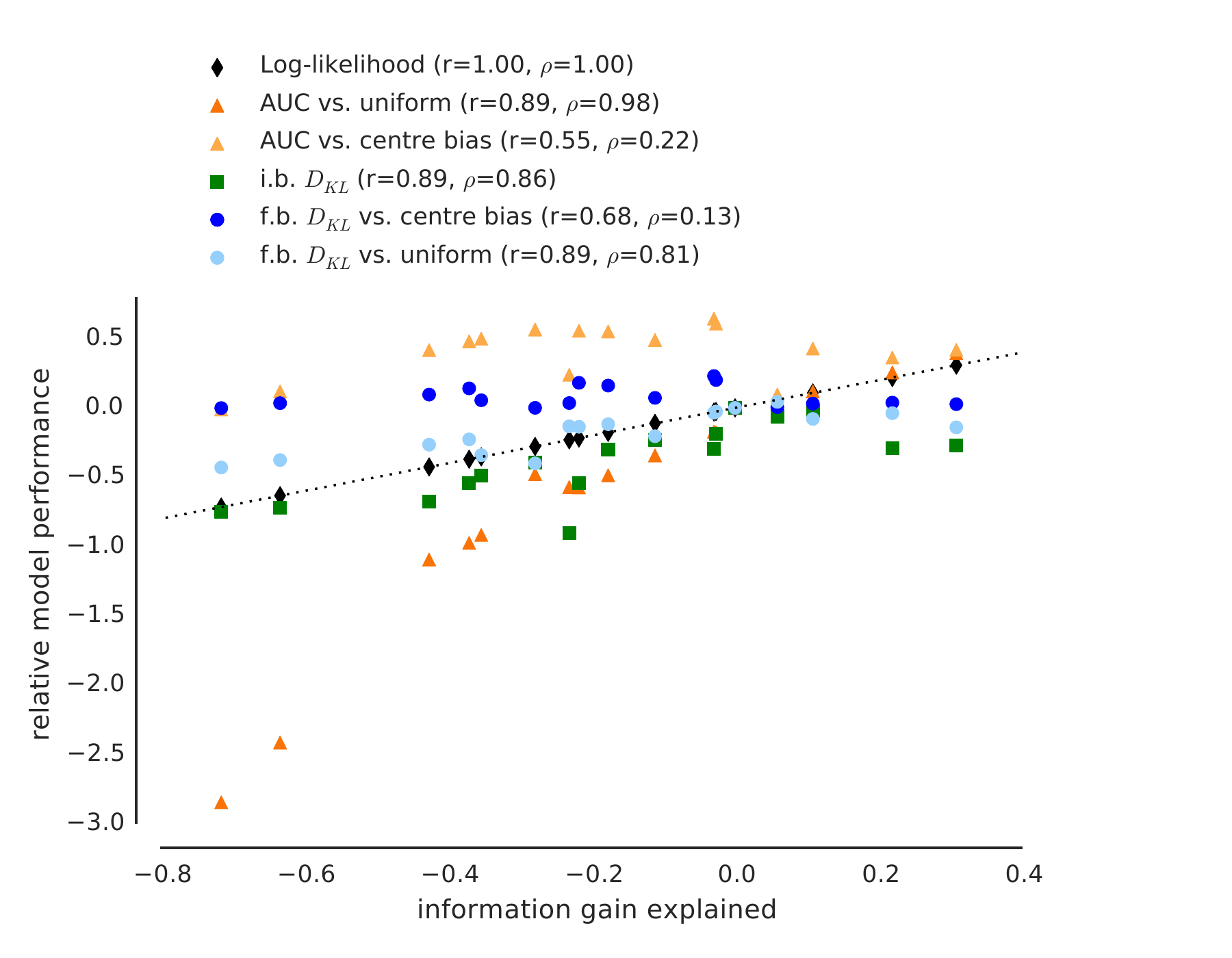}};

  \node[anchor=north west] at (label_rel_subfig) {\textsf{a)}};

  \node[anchor=north west] at (col2)
    {\includegraphics[width=\imgwidth\tikzunit]{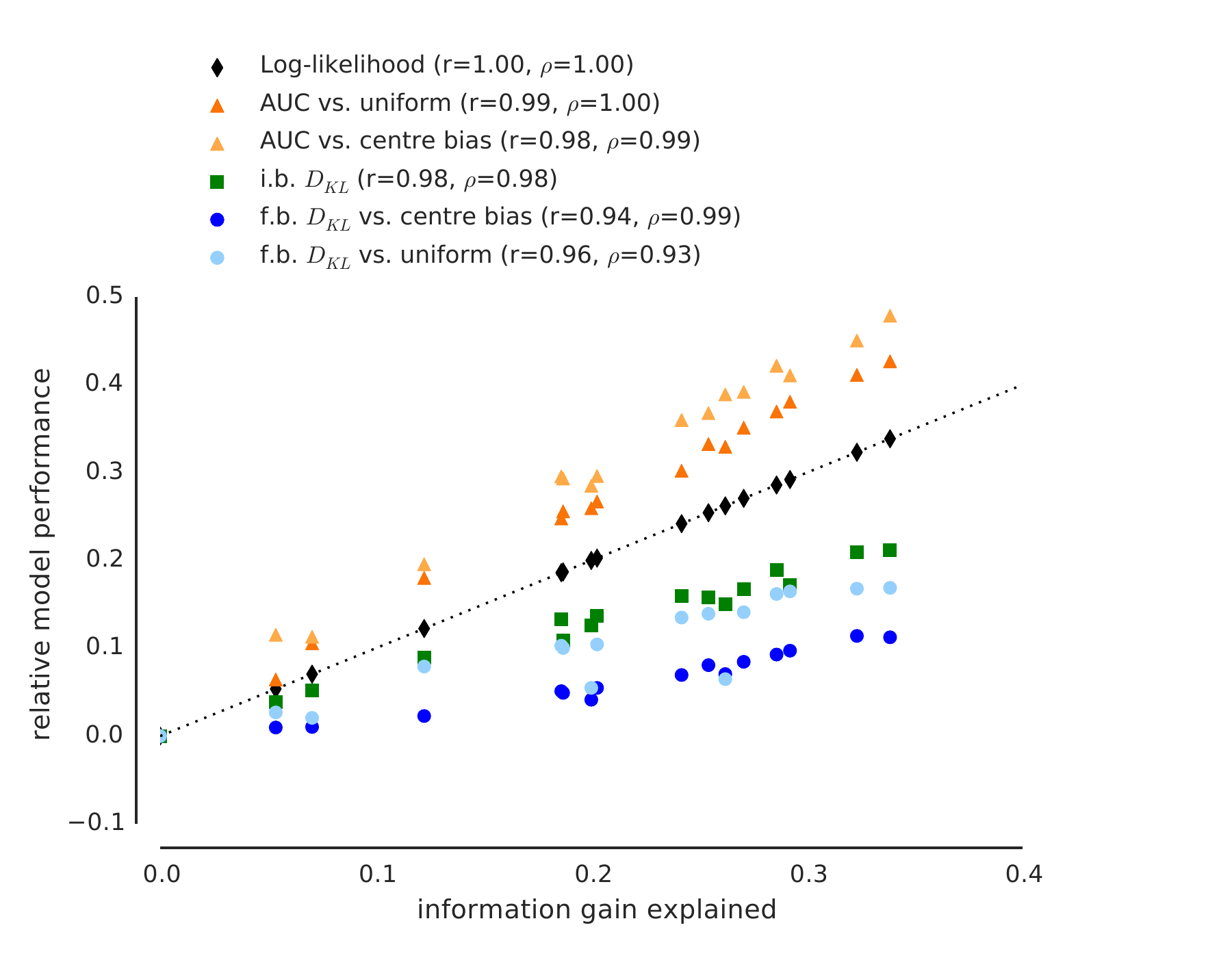}};
  \node[anchor=north west] at ($ (col2) + (label_rel_subfig2) $) {\textsf{b)}};

\end{tikzpicture}

\let\abstikzunit\undef
\let\scalingfactor\undef
\let\imgwidth\undef
\let\tikzunit\undef

%% file: figures/temporal_effects.pgf.tex
\ifdefined\noktikz
\else
  \usetikzlibrary{calc}
  \usetikzlibrary{shapes}
  \graphicspath{{/kyb/agmb/mkuemmerer/Documents/Uni/Bethge/Saliency/TPAMI/figures/}{/home/matthias/Documents/Uni/Bethge/Saliency/TPAMI/figures/}}

\fi

\newcommand\loglik[1]{\textsf{$\mathsf{#1}$\,bit/fix}}
\newcommand\auc[1]{\textsf{$\mathsf{#1}$\,\%}}

\ifdefined\abstikzunit
\else
        \newlength\abstikzunit
        \newlength\tikzunit
\fi
        \setlength\abstikzunit{1cm}
        \newcommand\scalingfactor{0.4}
        \newcommand\imgwidth{5}
        \setlength\tikzunit{\scalingfactor\abstikzunit}

\begin{tikzpicture}[scale=\scalingfactor,font=\tiny\sffamily,every node/.style={inner sep=0,outer sep=0}]
  \coordinate (center) at (5.45, 0);
  \coordinate (col2) at (5.9, 0);
  \coordinate (row2) at (0, -4.6);
  \coordinate (row3) at ($ 2*(row2) + (0, 0) $);
  \coordinate (horiz_text_rel_row) at  (center);
  \coordinate (vertical_text_rel_col) at  (-0.1, -2);

  \node[anchor=south,rotate=90] at (vertical_text_rel_col) {spatial density};  

  \node[anchor=north] (orig)  at (center)
    {\includegraphics[width=\imgwidth\tikzunit]{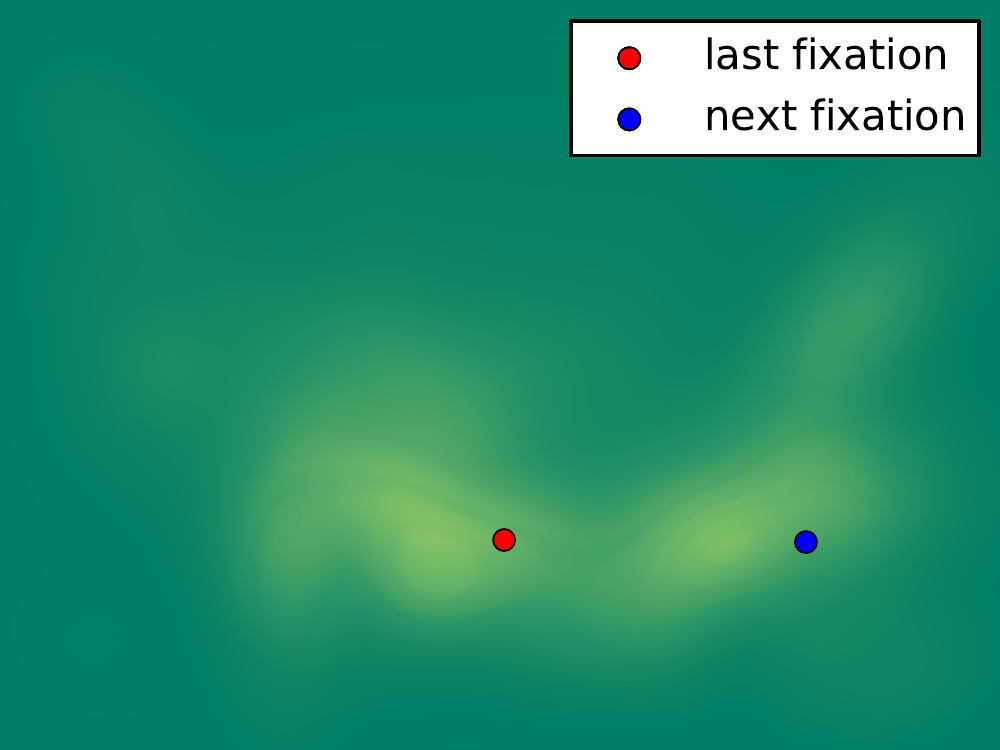}};

  \node at ($  (2.5, 0.4) $) {inhibition of return};  
  \node at ($ (col2) + (2.5, 0.4) $) {self excitation};  

  \node[anchor=south,rotate=90] at ($ (row2) + (vertical_text_rel_col)$) {correcting factor};

  \node[anchor=north west] (inhib_factor)  at (row2)
    {\includegraphics[width=\imgwidth\tikzunit]{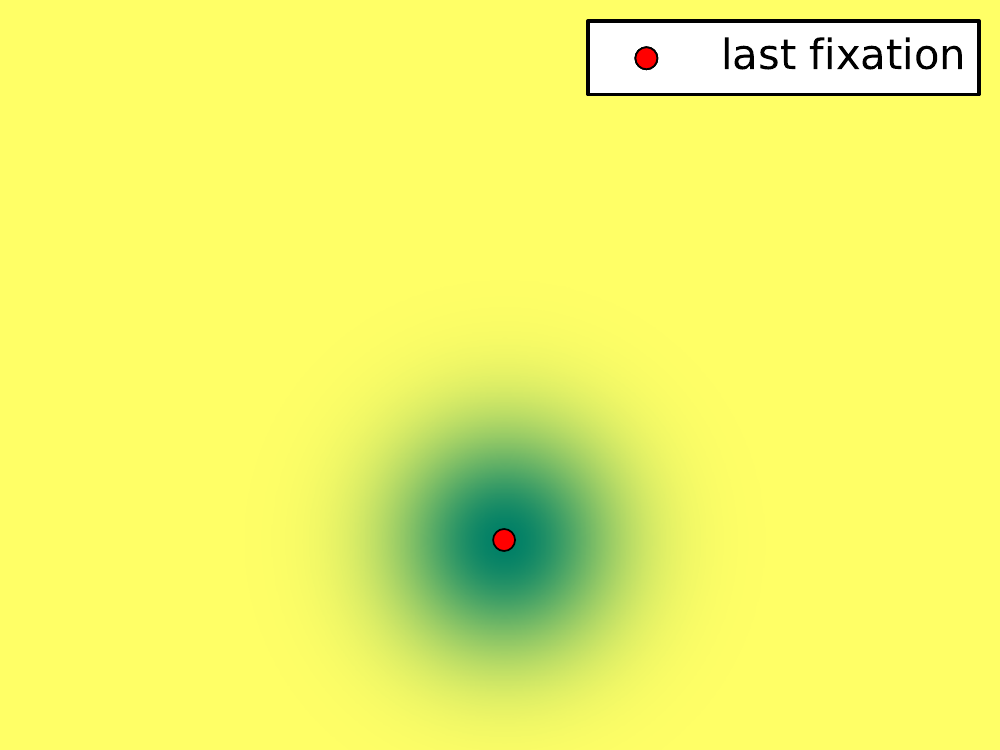}};

  \node[anchor=north west] (excitation_factor)  at ($ (row2) + (col2) $)
    {\includegraphics[width=\imgwidth\tikzunit]{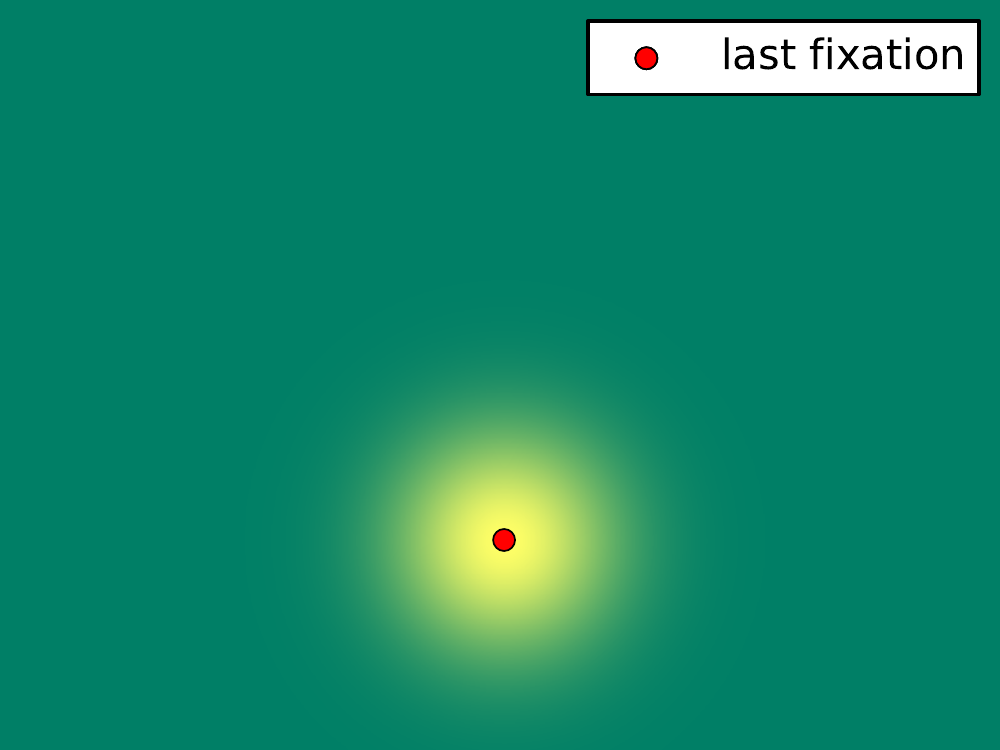}};

  \node[anchor=south,rotate=90] at ($ (row3) + (vertical_text_rel_col) $) {final density};  

  \node[anchor=north west] (inhib_after)  at (row3)
    {\includegraphics[width=\imgwidth\tikzunit]{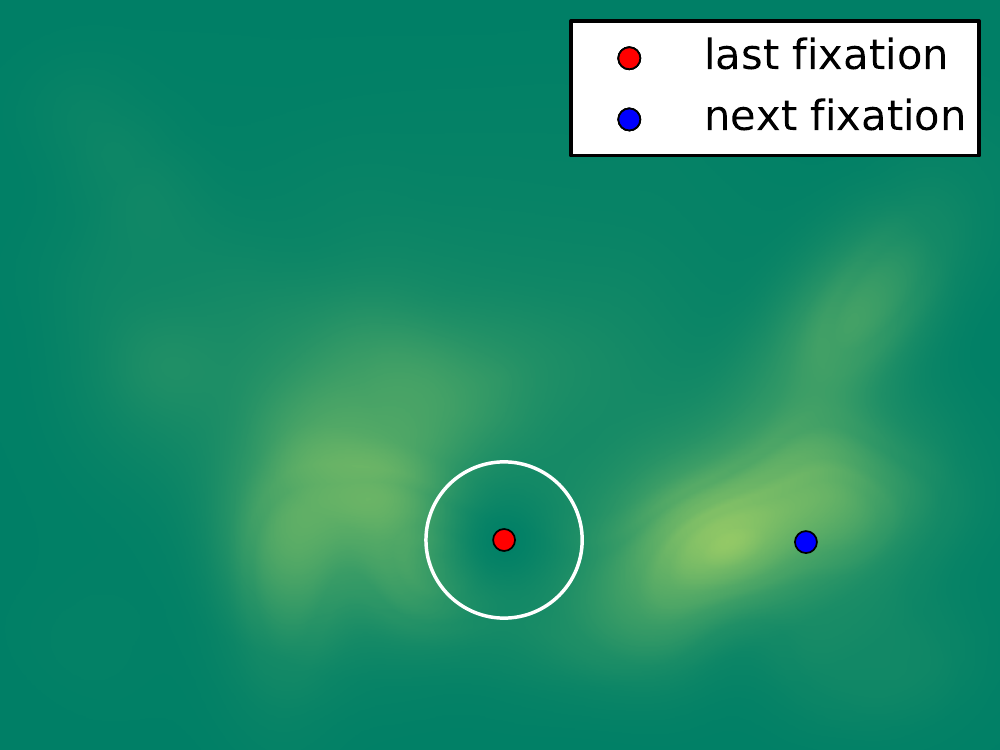}};

  \node[anchor=north west] (excitation_after)  at ($ (row3) + (col2) $)
    {\includegraphics[width=\imgwidth\tikzunit]{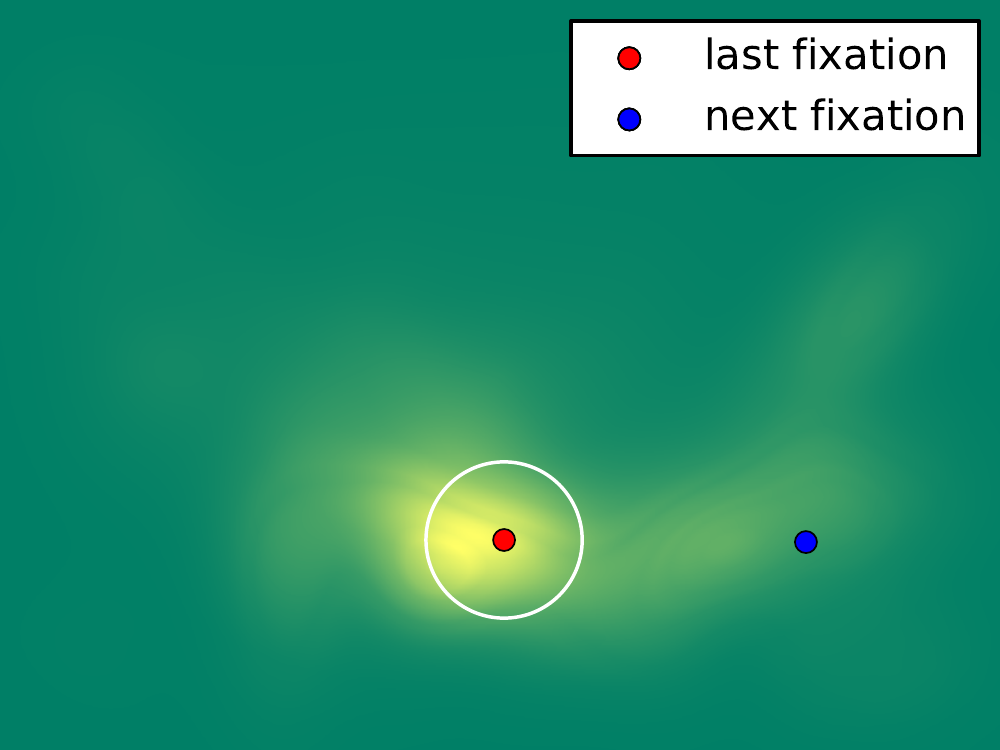}};

  \draw[->] (orig.south) -- (inhib_factor.north);
  \draw[->] (orig.south) -- (excitation_factor.north);

  \draw[->] (inhib_factor.south) -- (inhib_after.north);
  \draw[->] (excitation_factor.south) -- (excitation_after.north);

\end{tikzpicture}

\let\abstikzunit\undef
\let\scalingfactor\undef
\let\imgwidth\undef
\let\tikzunit\undef